\documentclass{article}
\pdfoutput=1

\usepackage{amssymb,dsfont,amsmath,amsthm,mathtools}
\usepackage[utf8]{inputenc}
\usepackage{mystyle, notations}
\usepackage{hyperref}
\usepackage{algorithmic, algorithm}
\usepackage{graphicx}

\newcommand{\rev}[1]{#1} 

\newcommand{\myparagraph}[1]{\paragraph{#1}}

\usepackage{rotating}
\usepackage[usestackEOL]{stackengine}

\begin{document}

\title{Unbalanced Optimal Transport, \\ from Theory to Numerics}

\author{Thibault Séjourné\footnote{DMA, ENS, PSL University}\:, {Gabriel Peyré$^{*}$\footnote{CNRS}}\:, François-Xavier Vialard\footnote{LIGM, UPEM}}

\maketitle


\begin{abstract}
Optimal Transport (OT) has recently emerged as a central tool in data sciences to compare in a geometrically faithful way point clouds and more generally probability distributions. 
The wide adoption of OT into existing data analysis and machine learning pipelines is however plagued by several shortcomings. This includes its lack of robustness to outliers, its high computational costs, the need for a large number of samples in high dimension and the difficulty to handle data in distinct spaces.
In this review, we detail several recently proposed approaches to mitigate these issues. 
We insist in particular on unbalanced OT, which compares arbitrary positive measures, not restricted to probability distributions (i.e. their total mass can vary). This generalization of OT makes it robust to outliers and missing data. The second workhorse of modern computational OT is entropic regularization, which leads to scalable algorithms while lowering the sample complexity in high dimension. The last point presented in this review is the Gromov-Wasserstein (GW) distance, which extends OT to cope with distributions belonging to different metric spaces. The main motivation for this review is to explain how unbalanced OT, entropic regularization and GW can work hand-in-hand to turn OT into efficient geometric loss functions for data sciences. \\
\textit{Keywords:} Unbalanced optimal transport, data sciences, optimization, Sinkhorn's algorithm, Gromov Wasserstein.
\end{abstract}

\section{Introduction}
\label{sec:intro}

\subsection{Distributions and positive measures in data sciences}

\begin{figure*}[h]
	\centering
	\begin{tabular}{c@{}c@{}c@{}c@{}c}
		{\includegraphics[width=0.25\linewidth]{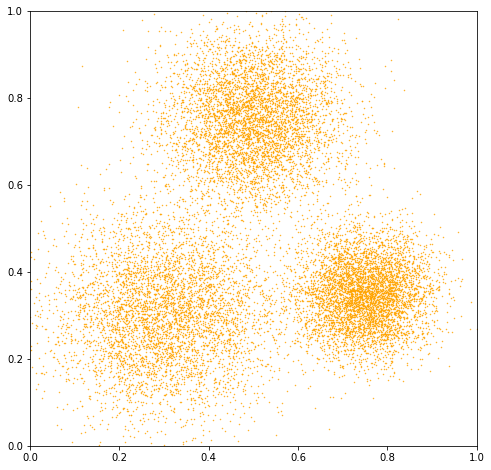}} & $\quad\quad$ &
		{\includegraphics[width=0.25\linewidth]{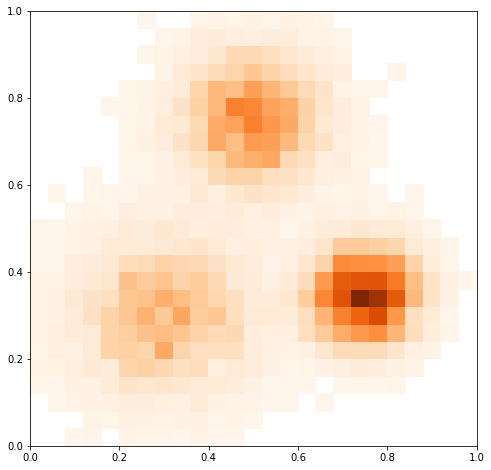}} & $\quad\quad$ &
		{\includegraphics[width=0.25\linewidth]{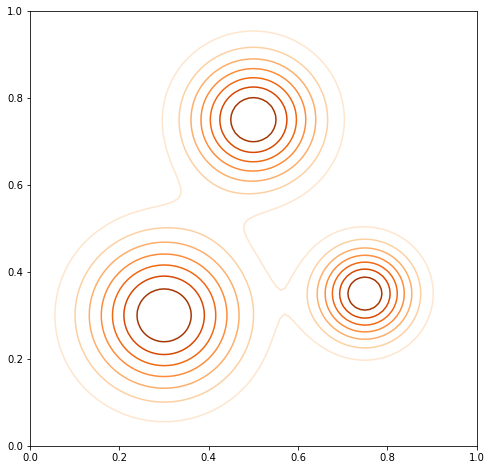}}
	\end{tabular}
	\caption{\textit{The same distribution defined on $\RR^2$ approximated as
			(left) a points cloud,
			(center) a 2D histogram,
			(right) a Gaussian mixture density.
		}
	}
	\label{fig:exmpl-measures}
	\vspace*{1.0em}
\end{figure*}

\myparagraph{Data representation using positive measures.}
Probability distributions, and more generally positive measures, are central in fields as diverse as physics (to model the state density in quantum mechanics~\cite{messiah2014quantum}), in chemistry (to model the electron density of a molecule~\cite{hansen2013assessment}), in biology (to count occurrences of a gene expression in cells~\cite{salzberg1998computational}) or economics (to represent wealth distribution~\cite{galichon2016optimal}).
They are also of utmost importance to represent and manipulate data in Machine Learning (ML)~\cite{james2013introduction,bishop2006pattern}.
In most cases, measures are represented and approximated using either discrete distributions (such as point clouds and histograms) or continuous models (such as parameterized densities).
See Figure~\ref{fig:exmpl-measures} for an example of a distribution discretized using these representations.

\myparagraph{Pointwise comparison of measures.}
Once the data and the sought-after models are represented as measures, many ML or imaging problems require to assess their similarity.
This comparison is performed using a loss function, such as a distance between measures, so that the smaller the loss is, the closer the two measures are.
We usually refer to a \emph{divergence} when the loss is positive and definite, so that the loss is zero if and only if the two measures are identical.
Having a loss satisfying the triangle inequality is a desirable feature, but is not mandatory.

To compare discrete distributions on a fixed grid (i.e. histograms), one can directly rely on usual norms such as the  $\ell_2$ and the $\ell_1$ norms.
The $\ell_1$ norm is often preferable, because it can be extended to general measures (i.e. not necessarily discrete) to define the Total Variation (TV) norm. A related construction is the relative entropy (also called the Kullback-Leibler/KL divergence), which is fundamental in statistics and ML~\cite{kullback1997information} because of its connection to maximum likelihood estimation~\cite{white1982maximum,gine2021mathematical} and information geometry~\cite{amari2000methods}. Section~\ref{sec:phidiv} details these two discrepancies (KL and TV) within the more general framework of $\phi$-divergences. 

These $\phi$-divergences have the advantage of simplicity, since they only perform pointwise comparisons of the measures at each position. This however totally ignores the metric of the underlying space. 
It makes these divergences incapable of comparing discrete distributions with continuous ones (for instance the KL divergence is always $+\infty$ in these cases). 
This issue motivates the use of other losses which leverage the underlying metric.
Examples of such losses are Maximum Mean Discrepancies~\cite{gretton2007kernel} and Optimal Transport distances~\cite{villani2003}, which are detailed in Sections~\ref{sec:mmd} and~\ref{sec:ot}.
This review focuses on Optimal Transport and its extensions, with an emphasis on their applicability for data sciences.

\myparagraph{Leveraging geometry.}
Leveraging some underlying distance is crucial to define loss functions taking into account the relative positions of the points in the distributions. 
This helps to capture the salient features of the problem, since geometrically close samples might share similar properties (such as in domain adaptation~\cite{courty2014domain}, see Section~\ref{sec:ot}).
The choice or the design of this distance is thus of primary importance.
In some cases, some natural metric is induced by the problem under study, such as for instance Euclidean or geodesic distance when studying some physical process~\cite{dukler2019wasserstein,cotar2013density}. 
In some other favorable cases, expert knowledge can be leveraged to design a metric, which can be for instance an Euclidean distance on some hand-crafted features. A representative example is in single cell genomics, where the cell can be embedded in a gene's space over which OT can be computed~\cite{schiebinger2017reconstruction}.
In more complicated setups, such as for images or texts, the metric should be learned in parallel to the resolution of the problem. This ``metric learning'' setup is still an active area of research~\cite{kulis2013metric,mikolov2013efficient}, and raises many challenging computational and theoretical questions.

\myparagraph{Toward robust loss functions between different spaces.}
A key difficulty in these applications to data science is to cope with very noisy datasets, which might contain outliers and missing parts. 
For instance, in applications to single cell genomics~\cite{schiebinger2017reconstruction}, the data is sensitive to changes of experimental conditions and cells exhibit complex mass variation patterns due to birth/death phenomena.   
OT faces difficulties in these cases, and is very sensitive to local variations of mass. 
This sensitivity to noise is illustrated in Section~\ref{sec:uot}, Figure~\ref{fig:ot_uot_matching}, see also~\cite{fatras2021unbalanced, mukherjee2021outlier}.
At the heart of this review is the description of a robust extension of OT, which is often referred to as \emph{unbalanced} OT~\cite{liero2015optimal} (abbreviated UOT), because of its ability to cope with ``unbalanced'' measures, i.e. with different masses. 
But there is more than simply copying with global variation in the total mass of the measure.
Most importantly, UOT is more robust to local mass variations, which includes outliers (which get discarded before operating the transport) and missing parts.

Another limitation of OT is its inability to compare measures defined on two distinct spaces.
This is for instance the case when comparing different graphs~\cite{petric2019got,vayer2020fused} or in genomics when the measurements are performed using distinct modalities~\cite{demetci2021unsupervised}.
Section~\ref{sec:gw} details a variant of OT called Gromov-Wasserstein (GW) which circumvents this issue, at the expense of solving a much more difficult (in particular non-convex) optimization problem.

\myparagraph{Notations.}
In what follows we consider a metric space $\X$ endowed with a distance $d_\X$.
We assume for the sake of simplicity that $\X$ is compact, which makes sense in most applications (extension to non-compact spaces requires some control over the moments of the measures).
We denote as $\Cc(\X)$ the space of continuous functions endowed with the supremum norm $\norm{\f}_\infty\eqdef \max_{x\in\X}|\f(x)|$ for $\f\in\Cc(\X)$.
We denote as $\int_\X f \dd \al = \int_\X\f(x)\dd\al(x)$ the pairing between functions $f \in \Cc(\X)$ and measures in the dual space $\al \in \Mm(\X)$ (so that $\al$ can be identified as a linear form).
A positive measure $\al \in \Mmpo(\X)$ is such that $\int_\X f \dd \al \geq  0$ for all positive functions $f\geq 0$.
A probability measure, denoted $\al\in\Mmpo(\X)$, satisfies the additional property $\al(\X) = \int_\X\dd\al(x)=1$.
For a generic measure $\al\in\Mmp(\X)$, the support is defined as the smallest closed set $\spt(\al)\subset\X$ such that $\al(\X\setminus\spt(\al))=0$.

The strong topology on $\Mm(\X)$ is the dual of the $\norm{\cdot}_\infty$ topology on $\Cc(\X)$ and is metrized by the total variation norm (see Section~\ref{sec:phidiv}).
The weak$^*$ topology, often referred to as the convergence in law in the case of probability distributions, is -- as its name suggests -- much weaker and is in some sense a geometric notion of convergence. 
A sequence $(\al_n)_n$ of measures converges weak$^*$ to $\al$, denoted  $\al_n\rightharpoonup\al$ if and only if for any test function $\f\in\Cc(\X)$, one has $\int_\X\f\dd\al_n\rightarrow\int_\X\f\dd\al$.
In other terms, it is the pointwise convergence on $\Cc(\X)$ of a sequence of linear forms $(\al_n)_n\in\Mm(\X)$.
As detailed in Section~\ref{sec:losses}, OT distances and kernel norms metrize this weak$^*$ topology. 

We consider in what follows ``cost functions'' $\C:\X^2\rightarrow\RR$, which are assumed to be symmetric ($\C(x,y)=\C(y,x)$), positive ($\C(x,y)\geq 0$), definite ($\C(x,y)=0\Rightarrow\, x=y$), and continuous (i.e. $(x,y)\mapsto\C(x,y)$ is continuous). A typical example is $\C(x,y)=d_\X(x,y)^p$ for some exponent $p>0$. We use such cost functions in the definition of OT (see Section~\ref{sec:ot}). It represents in the Monge problem the cost of transporting a unit of mass from $x$ to $y$.

\section{\texorpdfstring{$\phi$}{}-divergences, MMD and Optimal Transport}
\label{sec:losses}

We detail in this section several discrepancies between positive measures or probabilities, namely $\phi$-divergences, kernel norms (Maximum Mean Discrepancies) and Optimal Transport.
We discuss their theoretical and numerical strengths, as well as their shortcomings.
Unbalanced OT in Section~\ref{sec:uot} consists in blending $\phi$-divergences and Optimal Transport altogether, while kernel norms are connected to the entropic regularization of OT in Section~\ref{sec:entropy}.

\subsection{\texorpdfstring{$\phi$}{}-divergences}
\label{sec:phidiv}

One of the simplest family of discrepancies between positive measures are Csisz\'ar divergences~\cite{csiszar1967information} (also called $\phi$-divergences).
They consist in integrating pointwise comparisons of two positive measures, and are defined below.

\begin{definition}[$\phi$-divergence]\label{def:entropy}
Define an entropy function as a function $\phi:(0,\infty)\rightarrow[0,+\infty]$ which is convex, positive, lower-semi-continuous function and such that $\phi(1)=0$.
Define its recession constant $\phi^\prime_\infty\triangleq \lim_{p\rightarrow\infty} \phi(p) / p$.
For any $(\al,\be)\in\Mmp(\X)$, write the Radon-Nikodym-Lebesgue decomposition as $\al = \tfrac{\dd\al}{\dd\be}\be + \al^\bot$.
The $\phi$-divergence is defined as
\begin{align*}
\D_\phi(\al|\be)\triangleq \int_\X \phi\big(\tfrac{\dd\al}{\dd\be}(x)\big)\dd\be(x) + \phi^\prime_\infty\int_\X \dd\al^\bot(x).
\end{align*}
\end{definition}

Two popular instances of $\phi$-divergences are the Kullback-Leibler divergence where $\phi(p)=p\log p - p + 1$ (and $\phi^\prime_\infty=+\infty$) or the Total Variation divergence $\phi(p)=|p-1|$ (and $\phi^\prime_\infty=1$).
See also Section~\ref{sec:exmpl-phidiv} for other examples.
The former reads
\begin{align}\label{EqKLDef}
	\KL(\al|\be)\triangleq \int_\X \log\big(\tfrac{\dd\al}{\dd\be}(x)\big)\dd\al(x) - \int_\X\dd\be + \int_\X\dd\al,
\end{align}
when $\al^\bot=0$ (which corresponds to $\al$ as being absolutely continuous with respect to $\be$, and is noted $\al\ll\be$), and $\KL(\al|\be) = +\infty$ otherwise.
The total variation is a norm and reads
\begin{align*}
\TV(\al|\be)\triangleq \int_\X |\tfrac{\dd\al}{\dd\be}(x) - 1|\dd\be(x) + \int_\X\dd\al^\bot = |\al-\be|(\X).
\end{align*}
Total Variation admits another formulation as the dual norm of $\norm{\cdot}_\infty$ (it is an example of an integral probability metric~\cite{muller1997integral}).
It reads 
$$
	\TV(\al|\be) = \sup_{\norm{\f}_\infty\leq 1} \int_\X\f\dd(\al-\be), 
$$
so that TV induces the dual topology induced by the strong topology on $\Cc(\X)$.
Thus $\TV$, and more generally Csisz\'ar divergences with $\phi^{-1}(\{0\})=\{1\}$ (see~\cite[Corollary 2.9]{liero2015optimal}), induce a stronger topology than the weak$^*$ convergence, i.e. $\D_\phi(\al_n|\al)\rightarrow 0$ implies  $\al_n\rightharpoonup\al$, but the converse does not hold.
For instance take $x_n\rightarrow y$ but $x_n \neq y$, one has $\de_{x_n}\rightharpoonup\de_y$, but $\KL(\de_{x_n}|\de_y)=+\infty$ and $\TV(\de_{x_n}|\de_y)=2$.
Csisz\'ar divergences integrate pointwise comparison of measures $\tfrac{\dd\al}{\dd\be}(x)$ at any $x\in\X$, but ignore the proximity between positions $(x,y)$, hence their incapacity to metrize the weak$^*$ topology.

A chief advantage of these divergences is that they can be computed in $O(N)$ operations for discrete measures $\al=\sum_i^N \al_i\de_{x_i}$ and $\be=\sum_i^N \be_i\de_{x_i}$ sharing the same support $(x_i)_i$, since one has
\begin{align*}
\D_\phi(\al|\be)= \sum_{\be_i>0} \phi\big(\tfrac{\al_i}{\be_i}\big)\be_i + \phi^\prime_\infty\sum_{\be_i=0} \al_i.
\end{align*}

\myparagraph{Applications.}
As mentioned previously, the KL divergence is at the heart of several statistical methods such as  logistic regression~\cite{james2013introduction} and maximum-likelihood (ML) estimation~\cite{white1982maximum} \rev{to fit parametric models of measures}.
Given an empirical measure $\al_N=\tfrac{1}{N}\sum_{i=1}^N \de_{x_i}$, the maximum-likelihood method aims at finding a parameterized probability distribution $\beta_\thh$ approximating $\al_N$. The key hypothesis is that $\beta_\thh$ has a density $\xi_\thh$ with respect to some fixed reference measure $\mu$, i.e. $\dd \beta_\thh = \xi_\thh \dd \mu$ on $\X$. 
The model is estimated on the data support as $\hat\be_\thh = \sum_{i=1}^N \xi_\thh(x_i)\de_{x_i}$.
Maximizing the likelihood is equivalent to minimizing $\thh\mapsto\KL(\al_N|\hat\be_\thh)$ with respect to the parameter $\thh$.
Logistic regression or Gaussian mixture models~\cite{reynolds2009gaussian} correspond to the above minimization problem with a particular choice of the density $\xi_\thh$.
These minimization problems are often challenging because of the non-convexity w.r.t $\thh$.
A standard algorithm is the expectation-maximization (EM) algorithm~\cite{moon1996expectation}, which is sensitive to the initialization and can be trapped in local minima.

The KL divergence is also at the heart of \emph{information geometry}~\cite{amari2000methods}, \rev{which is an alternative way to define distances between distributions using Riemannian geometry}.
It consists in defining a canonical Riemannian manifold structure on a parametric family $(\beta_\theta)_\theta$ of distributions.
In the case of distributions with parameters $\thh\in\RR^d$, if all $\beta_\theta$ have a density $\xi_\theta$ with respect to a fixed reference measure, \rev{then the Riemannian metric is the second order Taylor expansion at $0$ of the function $\delta\thh\mapsto\KL(\be_\thh|\be_{\thh+\de\thh})$ w.r.t. $\delta\thh$}.
\rev{The second order term is called the Fisher information matrix and reads}
\begin{align*}
	\g(\thh) \triangleq \int_\X 
	\big[
	    \nabla\log \xi_\theta(x)\nabla\log \xi_\theta(x)^\top
	\big] 
	\dd\beta_\theta(x) \in\RR^{d\times d}.
\end{align*}
\rev{This matrix defines a quadratic form which in turn induces a geodesic distance. It can be used to perform Riemannian optimization} (to solve parameter estimation problems within this family) using for instance the ``natural'' gradient, which is the gradient with respect to this metric. 
In general, this metric is difficult to compute, but in some simple cases, closed forms have been derived. For instance, for a Gaussian distribution in $\RR$, with covariance $\si^2$ and mean $m \in \RR$, so that $\theta=(m,\sigma) \in \RR^2$, the Fisher matrix is $\g(\theta) = \tfrac{1}{\sigma^2}\diag(\tfrac{1}{2},1)$. 
This Riemannian metric is known as the \emph{hyperbolic Poincaré model}~\cite{cannon1997hyperbolic}.

\myparagraph{Challenges.} 
ML estimation enjoys many desirable features, such as providing optimal parameter selection processes in some cases. However, the optimal ML estimator might fail to exist (such as for instance when estimating Gaussian mixtures~\cite{lin2004degenerate}), and the model might even fail to have a density with respect to a fixed reference measure, or equivalently, the KL divergence between the data and the model is $+\infty$.
A typical example is when training generative learning models~\cite{goodfellow2014generative}, which correspond to models of the form $\beta_{\theta} = (\h_\theta)_\sharp \beta_0$, where $(\h_\theta)_\sharp$ denotes the push-forward by some parameterized map $\h_\theta$, and $\beta_0$ is a low-dimensional distribution. In this case, the measures $\be_\theta$ are singular, typically supported on low-dimensional manifolds.
For these applications, the KL loss should be replaced by weaker losses, which is an important motivation to study geometrically aware losses as we detail next. 

\subsection{Maximum mean discrepancies}
\label{sec:mmd}

Euclidean norms over a space of measures are defined by integrating so-called kernel functions between the input distributions. They define weaker discrepancies which take into account the underlying distance over $\X$.

\myparagraph{Reproducing kernels.}
A symmetric function $k:\X\times\X\rightarrow\RR$ is called a \emph{positive definite} (p.d.) kernel on $\X$ if for any $n\in\NN$, for any $(x_1,\ldots,x_n)\in\X^n$ and $(c_1,\ldots,c_n)\in\RR^n$ one has $\sum_{i,j=1}^n c_i c_j k(x_i, x_j)\geq 0$.
It is said to be \emph{conditionally positive} if the above inequality only holds for scalars $(c_1,\ldots,c_n)\in\RR^n$ such that $\sum_{i=1}^n c_i = 0$.
In practice the points $(x_1,\ldots,x_N)$ correspond to the dataset's samples, and computing the matrix $K\triangleq (k(x_i,x_j))_{ij}$ is the building block at the basis of all numerical kernel methods.
The kernel is called \emph{reproducing} if there exists a Hilbert space $\Hh$ of functions from $\X$ to $\RR$ (with inner product $\dotp{\cdot}{\cdot}_\Hh$) such that: (i) for any $x\in\X$, $k(\cdot,x)\in\Hh$ and (ii) for any $x\in\X$, for any $\f\in\Hh$, one has $\dotp{\f}{k(\cdot,x)}_\Hh=\f(x)$.
The Moore-Aronsazjn Theorem~\cite{aronszajn1950theory} states that for any p.d. kernel there exists a unique Hilbert space $\Hh$ of functions from $\X$ to $\RR$ such that $k$ is a \emph{reproducing kernel}.
For any reproducing kernel we denote $\psi(x)=k(x,\cdot)$, so that $\psi : \X \rightarrow \Hh$ defines a so-called feature map satisfying $k(x,y) = \dotp{\psi(x)}{\psi(y)}_\Hh$.

\myparagraph{Kernel norms for measures.}
The dual norm associated with $\Hh$ induces the so-called MMD norm on the space of Radon measures since they belong by hypothesis to the dual space of 
$\Hh$~\cite{micchelli2006universal, gretton2012kernel}.

\begin{definition}[MMD norms]\label{def:mmd}
The MMD norm is defined for any $\al\in\Mmp(\X)$ as 
$$
    \norm{\al}_{\Hh^*}\triangleq \sup_{\norm{\f}_{\Hh}\leq 1} \int_\X\f\dd\al.
$$
This supremum is reached with a function $\f$ which is proportional to $k\star\al\triangleq\int_{\X} k(\cdot,y)\dd\al(y)$ with $\norm{\f}_{\Hh}=1$, so that 
\begin{equation}\label{eq:mmd-cont}
        \norm{\al}^2_{\Hh^*}=\int_{\X^2} k(x,y)\dd\al(x)\dd\al(y).
\end{equation}
\end{definition}

This norm endows $\Mmp(\X)$ with a Hilbert structure. Thanks to the previous formula, the inner product on $\Mmp(\X)$ reads 
$$
    \dotp{\al}{\be}_{\Hh^*} = \int_{\X^2} k(x,y)\dd\al(x)\dd\be(y).
$$
When the pseudo-norm $\norm{\al}_{\Hh^*}$ is a norm \rev{over the space of measures $\Mm(\X)$, the kernel is said to be definite}. When it is a norm only when restricted to  measures of total mass $0$,  it is said to be  conditionally definite.
\rev{This means that if the kernel is definite (such as the Gaussian kernel), then the associated MMD norm is ``unbalanced'' and can  be used to compare measures with different total mass.}
It is known to metrize the weak* topology when the kernel is \emph{universal}, i.e. when the set of functions $\{x\mapsto k(x,y), y\in\X\}$ is dense in $\Cc(\X)$~\cite{micchelli2006universal, sriperumbudur2010relation}.
On $\X=\RR^d$, typical examples of universal kernels are the Gaussian and Laplacian kernels, of the form $k(x,y) = e^{-\norm{x-y}^p/s}$ for respectively $p=2$ and $p=1$, and $s>0$.   
In contrast, the energy distance kernel $k(x,y) = -\norm{x-y}_2$ is universal but only conditionally positive, so it can only be used to compare probability distributions \rev{(it is a ``balanced'' norm)}.

\myparagraph{Applications.}
Many machine learning methods such as classification via support vector machines or regression~\cite{scholkopf2002learning}, statistical tests~\cite{gretton2007kernel, gretton2012kernel}, or generative learning~\cite{li2017mmd, bellemare2017cramer} can be adapted to the setting of reproducing kernel Hilbert spaces.
Through the design of adapted kernels, this enables the application of these methods to non-Euclidean data such as strings, sequences and graphs~\cite{scholkopf2004kernel}.
A typical application is the modeling and classification of proteins~\cite{borgwardt2005protein} using graph kernels~\cite{bishop2006pattern}.
The MMDs leverage these existing kernel constructions to compare collections of points (point clouds) or more generally distributions representing data~\cite{gretton2007kernel,gretton2012kernel}. 
In practice, one usually considers discrete empirical distributions,  $(\al_N,\be_N)$, often assumed to be drawn from some unknown measures $(\al,\be)$.
This is at the heart of two-sample tests to check whether $\al=\beta$ by checking if $\norm{\al_N - \be_N}_{\Hh^*}$ is small enough. 
Another application of kernels is to perform density estimation~\cite{terrell1992variable} by smoothing a discrete measure $\al_N$ to obtain $(k\star\al_N)(x) \triangleq\int k(x,y)\dd\al_N(y)$. 
When $\al_N=\sum_i \al_i\de_{x_i}$, this smoothing reads $(k\star\al_N)(x) = \sum_i \al_i k(x,x_i)$.

\myparagraph{Numerical computation.}
The computation of the MMD norm of some measure $\al_N = \sum_i \al_i \de_{x_i}$ with $N$ samples requires $N^2$ operations, since 
\begin{equation}\label{eq:mmd-discr}
    \norm{\al_N}_{\Hh^*}^2 = \sum_{i,j} k(x_i,x_k) \al_i \al_j = \dotp{K\al}{\al}_{\RR^N}
\end{equation}
where $K \triangleq (k(x_i,x_j))_{i,j} \in \RR^{N \times N}$. This might be prohibitive for large datasets, but this complexity can be mitigated using GPU computations~\cite{feydy2020fast} and low rank Nystr\"{o}m methods~\cite{williams2000using, rudi2017falkon, meanti2020kernel}. 

An important feature of these MMD norms, compared to OT distances (see Section~\ref{sec:sample-comp}) is that they suffer less from the curse of dimensionality~\cite{sriperumbudur2012empirical, rudi2017generalization}. The intuition is that the discrete formula~\eqref{eq:mmd-discr} is similar to a Monte-Carlo integration technique to approximate~\eqref{eq:mmd-cont}. If the $(x_i)_i$ are drawn independently from $\al$, this explains why the discretization error is of the order $1/\sqrt{N}$, and the decay with $N$ is independent of the dimension.  

\begin{figure}[h]
	\centering
	\includegraphics[width=0.4\linewidth]{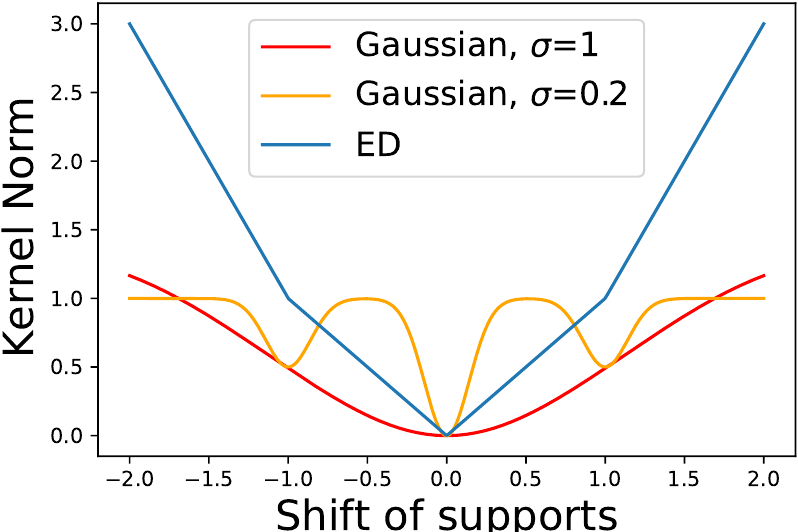}
	\caption{\textit{Computation of $\norm{\al - \al_\tau}_{\Mmp(\X)}^2$ where $\X=\RR$, $\al=\tfrac{1}{2}\de_{-0.5} + \tfrac{1}{2}\de_{0.5}$ and $\al_\tau=\tfrac{1}{2}\de_{\tau-0.5} + \tfrac{1}{2}\de_{\tau+0.5}$. The plot displays the kernel norm value as a function of $\tau\in[-2,2]$. We consider the Gaussian kernel (with $\sigma\in\{0.2,\, 1\}$) and the energy distance (ED) kernel.}}
	\label{fig:kernel_shift}
\end{figure}

\myparagraph{Challenges.}
In practice, the choice of the kernel function has a significant impact on the performance in applications.
For instance, the Gaussian kernel $k(x,y)=e^{-\norm{x-y}^2 / 2\si^2}$ depends on a bandwidth parameter $\sigma$.
As illustrated in Figure~\ref{fig:kernel_shift}, a poor tuning of $\sigma$ yields a kernel norm which is highly non-convex with respect to shifts of the measures' supports.
By comparison, the energy distance kernel $k(x,y)=-\norm{x-y}_2$ is parameter free, and yields a convex loss for this example, but can only be used between probability distributions (and not arbitrary positive measures).

\begin{figure}[h]
	\begin{subfigure}{.32\textwidth}
		\centering
		\includegraphics[width=.95\linewidth]{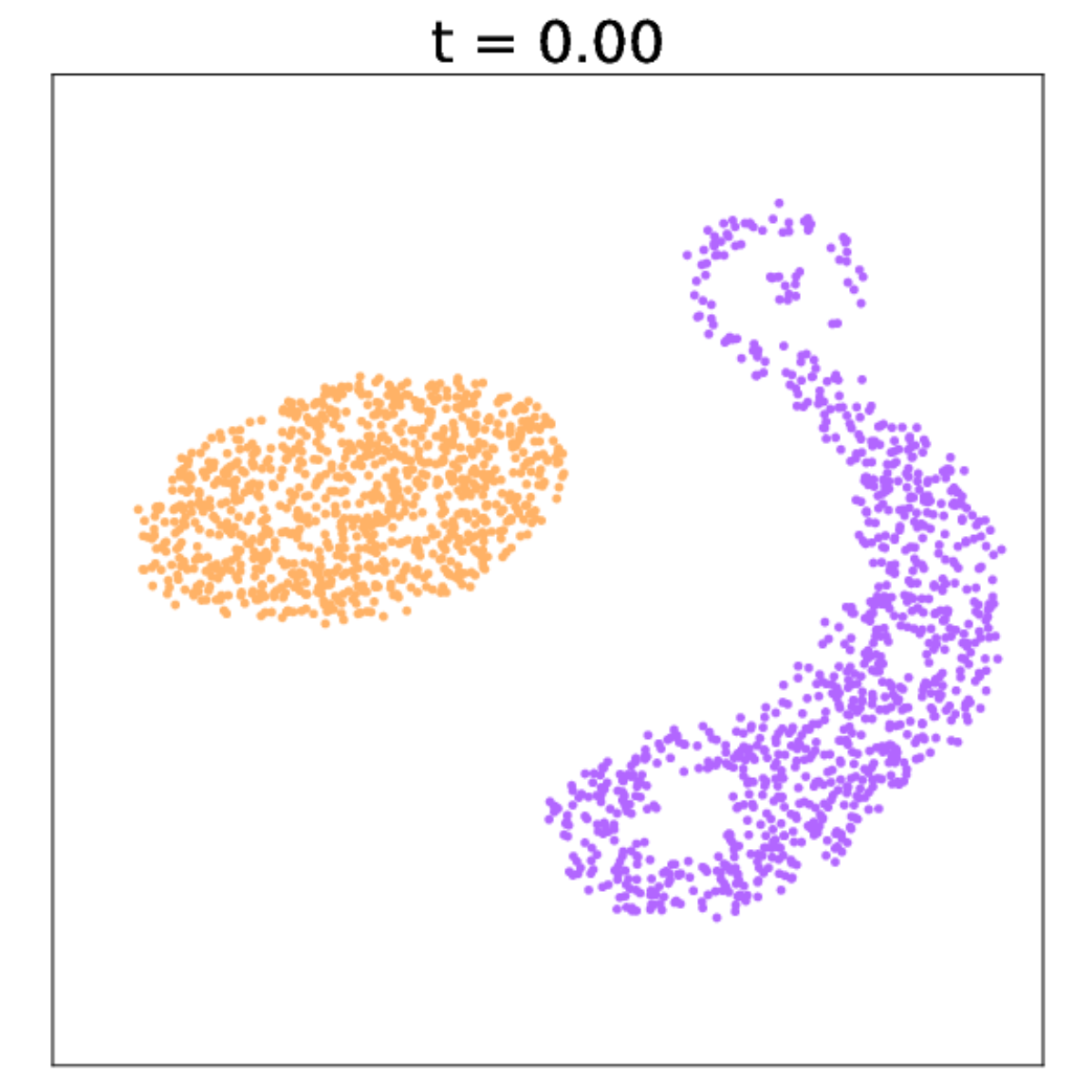}
		\caption{$\al(x_0)$ (orange) and $\be$ (purple)}
		\label{fig:flowa}
	\end{subfigure}
	\hfill
	\begin{subfigure}{.32\textwidth}
		\centering
		\includegraphics[width=.95\linewidth]{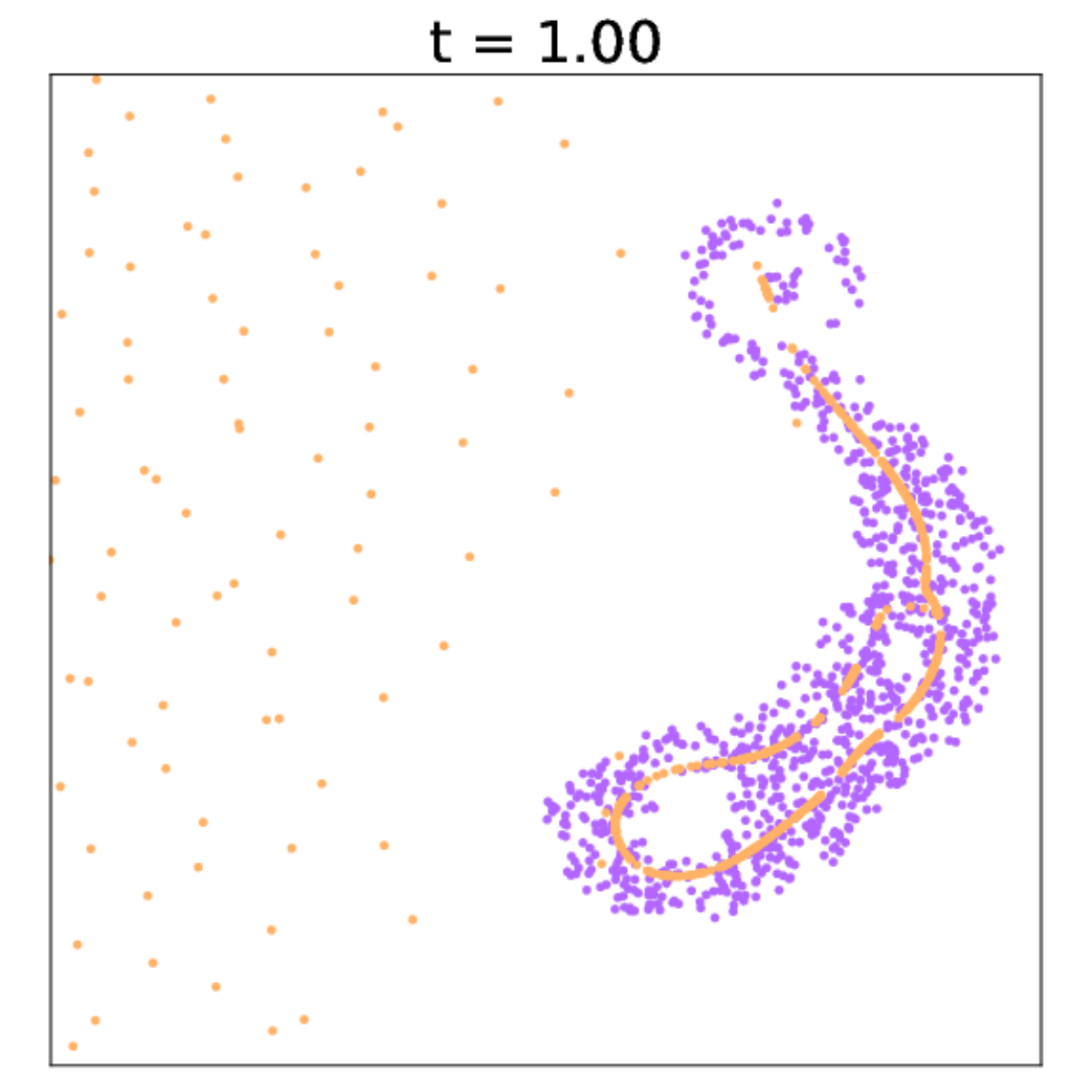}
		\caption{Gaussian kernel}
		\label{fig:flowb}
	\end{subfigure}
	\hfill
	\begin{subfigure}{.32\textwidth}
		\centering
		\includegraphics[width=.95\linewidth]{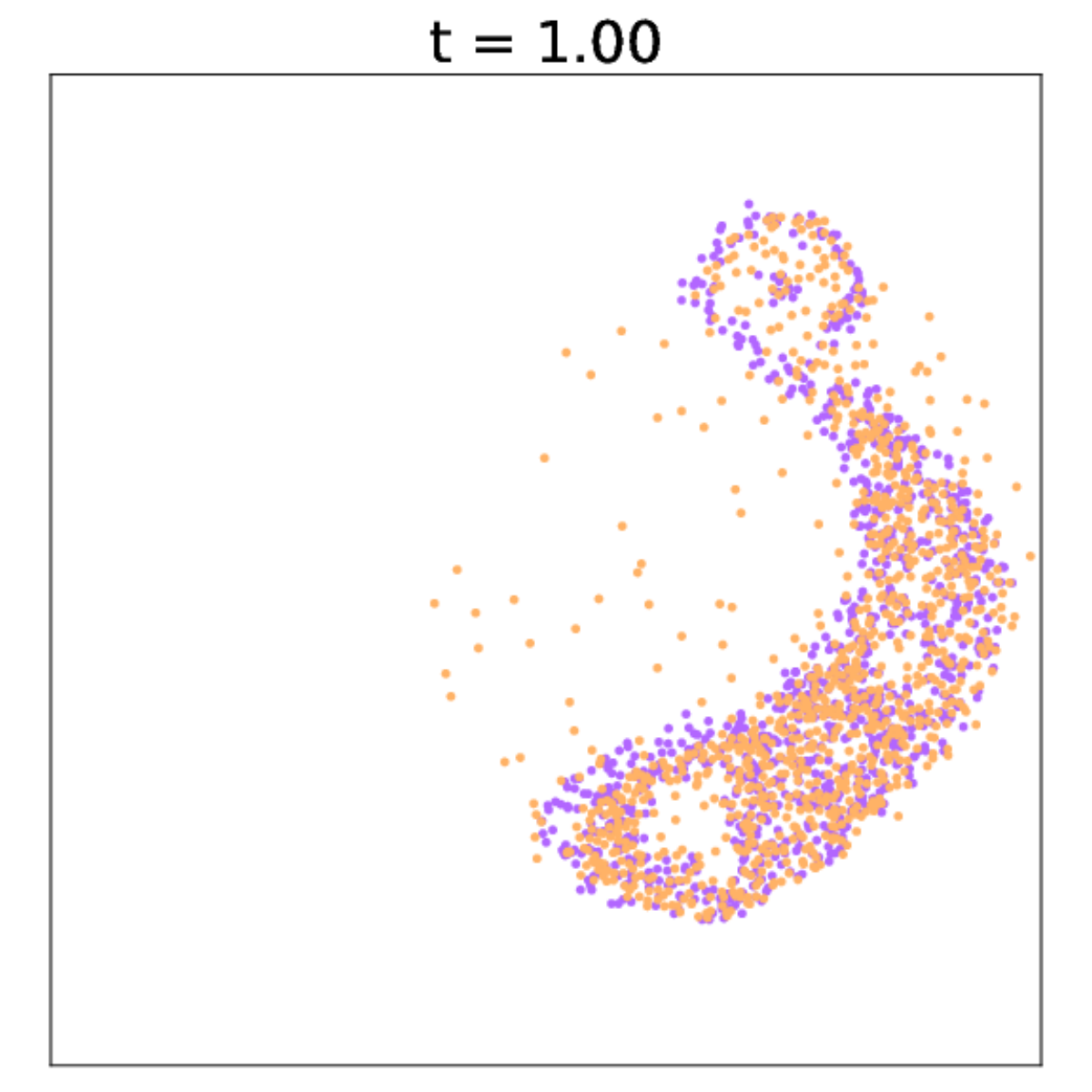}
		\caption{ED kernel}
		\label{fig:flowc}
	\end{subfigure}
	\caption{\textit{
		Display of initialization~\ref{fig:flowa} and outputs of the gradient flow for the Gaussian kernel~\ref{fig:flowb} and the energy distance (ED) kernel~\ref{fig:flowc}.
		}
	}
	\label{fig:grad_flow_kernel}
\end{figure}

This non-convexity can be detrimental when using these kernel loss functions to fit models by distance minimization, as in Figure~\ref{fig:grad_flow_kernel}. 
We consider two discrete distributions $\al_x = \tfrac{1}{N}\sum_{i=1}^N \de_{x_i}$ and $\be_y = \tfrac{1}{N}\sum_{i=1}^N \de_{y_i}$.
We optimize $\al_x$, which is directly parameterized by the positions $x=(x_1,\ldots,x_N)$, by doing a gradient descent of the energy $x \mapsto \Ff(x) \triangleq \norm{\al - \be}_{\Hh^*}^2$. 
This type of optimization is motivated by applications as diverse as the training of neural networks with a single hidden layer \cite{chizat2018global, rotskoff2019global} (where the $x_i$ represent the hidden neurons), domain adaptation~\cite{courty2014domain} and optical flow estimation~\cite{menze2015object, liu2019flownet3d}.
We display in Figure~\ref{fig:grad_flow_kernel} the initialization and the output of the gradient flow after performing gradient descent on $\Ff(x)$.
One sees that for the Gaussian kernel (with small $\sigma$), the output $\al$ does not match $\be$, with even particles moving away from the target $\be$.
For this task the Gaussian kernel is thus not efficient.
In contrast, the energy distance kernel is able to retrieve the support of $\be$, but the convergence is slow and particles are lagging behind. 
This lag is problematic in training as it requires more gradient descent iterations (and thus it takes more time) to obtain the model's convergence. Our observations are only empirical and this kernel is not known to have better (Wasserstein) geodesic convexity than Gaussian kernels.
Leveraging OT, we show in Section~\ref{sec:ot} it is possible to avoid this issue (see Figure~\ref{fig:flow-reg}), \rev{thus making it a powerful tool for model fitting tasks}.

Besides fitting distributions, 
\rev{another important use case of OT is to compute} ``averages'' (or barycenters) of several input distributions.
A barycenter of $K$ measures $(\al_1,\ldots,\al_K)$ for some discrepancy $\Ll$ is defined as the minimizer 
$$
    \be^\star=\arg\min_\be \sum_{k=1}^K \om_k \Ll(\be,\al_k)
    \qwhereq 
    \om_k\geq 0, \; 
    \sum_{k=1}^K \om_k =1.
$$
Since kernel norms induce a Hilbertian geometry on $\Mmp(\X)$, when $\Ll(\alpha,\beta) = \norm{\alpha-\beta}^2_{\Hh^*}$, the barycenter is simply a linear mixture $\be^\star = \sum_{k=1}^K \om_k \alpha_k$.
In particular, the interpolation between $(\de_x,\de_y)$ with $\om = (t,1-t)$ for $t\in[0,1]$ is $\be^\star = t\de_x + (1-t)\de_y$.
However, in some applications such as template matching for shapes~\cite{bazeille2019local}, one might be interested by an interpolation acting on the support instead of the weights, i.e. which reads $\be^\star = \de_{tx + (1-t)y}$.
Doing so requires the use of a non-Hilbertian distance.
As we explain in the next section, such interpolation is made possible using Optimal Transport distances (see Figure~\ref{fig:interp-gaussian} later in Section~\ref{sec:uot}).

\subsection{Balanced Optimal Transport}
\label{sec:ot}

\rev{We provide in this section a brief review on the main properties of Optimal Transport which are useful to the understanding of its unbalanced extension. 
We refer to the monographs~\cite{peyre2019computational, santambrogio2015optimal} for the reader seeking additional numerical illustrations or a deeper analysis of its properties.}
\rev{We recall that for the sake of simplicity, we assume the domain $\X$ to be compact.}

\myparagraph{Monge's transport.}
Optimal Transport (OT) is a problem which traces back to the work of Gaspard Monge~\cite{monge1781memoire}.
Monge's problem seeks a matching between two measures $(\al,\be)$ to dig trenches modeled as distributions of sand.
Such a map $T:\X\rightarrow\X$ is constrained to send the mass of $\al$ onto $\be$.
This constraint reads $T_\sharp\al = \be$, where $T_\sharp$ is the push-forward operator defined by the relations $\int_\X \f\circ T \dd\al = \int_\X\f\dd\be$ for any test function $\f\in\Cc(\X)$.
In particular, this ensures that $T_\sharp \de_x = \de_{T(x)}$.
Among all these admissible maps, the optimal transport minimizes the overall cost of transportation, where one assumes some price $\C(x,y)$ is associated with a matching $y=T(x)$. A typical example is $\C(x,y)=\norm{x-y}_2^p$ in $\X=\RR^d$ for some $p>0$ (Monge original problem corresponding to $p=1$, while the case $p=2$ enjoys several additional mathematical properties as detailed next). 

\begin{definition}[OT, Monge formulation]\label{def:monge}
The optimal transport Monge map, when it exists, solves 
\begin{align*}
	\inf_{T_\sharp\al=\be} \int_\X \C(x,T(x)) \dd\al(x).
\end{align*}
\end{definition}

OT maps find numerous applications, for instance in domain adaptation~\cite{courty2014domain} to extend a trained prediction model from one dataset to another one, or in genomics~\cite{demetci2020gromov} to align different pools of cells over some genes’ spaces.

An important question is to check whether such an optimal map exists. If $\al$ is discrete but not $\be$, then no Monge map exists, since $T_\sharp \al$ is necessarily discrete, so that the mass conservation constraint set $T_\sharp \al=\be$ is empty. Brenier's theorem~\cite{brenier1991polar} ensures the existence and uniqueness of a Monge map in $\RR^d$, for the cost $\C(x,y)=\norm{x-y}^2$, if $\al$ has a density with respect to Lebesgue's measure. Furthermore the Monge map is the unique gradient of a convex function satisfying the mass conservation constraint, i.e. $T=\nabla u$ with $u$ a convex function.

\myparagraph{Kantorovich's transport.}
To deal with arbitrary measures, it is thus necessary to relax the mass conservation equation and to allow for some form of ``mass splitting''. This relaxation was introduced by Kantorovich~\cite{kantorovich42transfer}, first for applications to economic planification.
In this formulation, the map $T$ is replaced by a measure $\pi\in\Mmp(\X\times\X)$,  called a \emph{transport plan} .
Instead of moving all the mass from a position $x$ to $T(x)$, the measure $\pi$ is allowed to move the mass from $x$ to several destinations.
This problem is in some sense a relaxation of Monge's problem (see~\cite[Section 1.5]{santambrogio2015optimal}).

\begin{definition}[OT, Kantorovitch formulation]\label{def:ot}
An Optimal Transport plan between two \rev{positive} measures $(\al,\be)\in\Mmp(\X)$ solves
\begin{align*}
\OT(\al,\be) \triangleq \min_{\pi\in\Uu(\al,\be)} \int_{\X^2} \C(x,y)\dd\pi(x,y),
\end{align*}
where the constraint set $\Uu(\al,\be)$ reads
\begin{align*}
	\Uu(\al,\be)\triangleq \bigg\{ \pi\in\Mmp(\X\times\X),\,(p_\X)_\sharp\pi=\al,\, (p_\Y)_\sharp\pi=\be  \bigg\},
\end{align*}
and $p_\X$ and $p_\Y : \X\times\X\rightarrow\X$ are canonical projections defined as
\begin{align*}
	p_\X:  (x,y)\mapsto x, 
    \qquad
    p_\Y:  (x,y)\mapsto y.
\end{align*}
\end{definition}

The coupling encodes the amount of mass transported between positions $x$ and $y$, and in this case one pays a price $\C(x,y)$ multiplied by the transported mass.
The constraint $(p_\X)_\sharp\pi=\al$ formally means that one has $\int_{\X^2} \f(x)\dd\pi(x,y)=\int_\X\f(x)\dd\al(x)$ for any test function $\f\in\Cc(\X)$.
Intuitively, it means that the sum of the mass transported from a source position $x$ is equal to the mass of $\al$ at $x$.
It imposes a transportation performed from a source measure $\al$ to a target $\be$, and such that mass is conserved because $\int\dd\pi = \int\dd\al=\int\dd\be$.
\rev{Note that the problem is infeasible when $\int\dd\al\neq\int\dd\be$, because $\Uu(\al,\be)=\emptyset$, thus one has $\OT(\al,\be)=+\infty$.}

In sharp contrast to Monge's problem an optimal solution to Kantorovitch's problem always exists, but might fail to be unique. 
In particular, the feasible set is always non empty, because $\al\otimes\be\in\Uu(\al,\be)\neq\emptyset$, (where $\dd(\al\otimes\be)(x,y)\triangleq \dd\al(x)\dd\be(y)$).
In addition to providing access to an optimal coupling $\pi$ between the measure, the cost $\OT(\al,\be)$ of the optimization problem itself is of primary importance. Indeed, if $\C(x,y)=d_\X(x,y)^p$ for some $p \geq 1$, then $\OT^{1/p}$ defines the so-called  $p$-Wasserstein distance~\cite{villani2003}. This Wasserstein distance metrizes, when the underlying space is compact, the weak$^*$ convergence, i.e. $\al_n\rightharpoonup\al$ if and only if $\OT(\al_n,\al)\rightarrow 0$. A distinctive feature of this distance, among all possible distances metrizing the weak$^*$ topology (such as MMD norms), is that it exactly ``lifts'' the ground distance in the sense that $\OT^{1/p}(\de_x,\de_y) = d_\X(x,y)$.

\myparagraph{Discrete setting.}
In many applications in data sciences, the distributions are discrete empirical measures, $\al=\sum_{i=1}^N \al_i\de_{x_i}$ and $\be=\sum_{j=1}^M \be_j\de_{y_j}$.
In this case the $\OT$ problem becomes a finite dimensional linear program 
\begin{align*}
	\OT( \al, \be ) \triangleq \inf_{\pi\in\Uu(\al,\be)}\quad
	\sum_{i=1}^N\sum_{j=1}^M \C(x_i,y_j)\pi_{i,j},
\end{align*}
where the marginal constraints on $\pi$ read
\begin{align*}
\Uu(\al,\be)\triangleq \bigg\{ \pi\in(\RR_+)^{N\times M},\,\,\sum_j\pi_{ij}=\al_i,\,\, \sum_i \pi_{ij}=\be_j  \bigg\}.
\end{align*}
In this case, $\Uu(\al,\be)$ defines a compact polytope.
A remarkable setting is when $N=M$ (same number of samples) and $\al_i=\be_j=\tfrac{1}{N}$ (uniform weights), in which case $\Uu(\al,\be)$ defines the Birkhoff polytope, whose extremal points are permutation matrices.
Since OT is a linear program, we thus have the existence of an optimal $\pi^\star$ which is a permutation minimizing the transportation of samples $(x_i)_i$ to match samples $(y_j)_j$. 
This result is true independently of the cost and it does not extend to the continuous setting without assumptions. Indeed, Brenier's theorem requires a non-degeneracy (so-called twist) condition on the cost to hold.
Yet, it highlights the fact that the OT couplings generalize the notion of optimal matching, and are thus quite different from ``closest point'' matchings routinely used in many applications. 
For general input measures (i.e. non uniform weights, or when $N \neq M$), the optimal $\pi^\star$ is a sparse assignment matrix with less than $M+N$ non zero entries (see~\cite[Exercise 41]{santambrogio2015optimal}).

\myparagraph{Dynamic formulation.}
On the Euclidean space and more generally on Riemannian manifolds, when the cost $c$ is the distance squared denoted $d^2$, a dynamic formulation of the Kantorovich's transport was first introduced in \cite{benamou2000computational}. This so-called Benamou-Brenier formulation  consists in the minimization of the kinetic energy of the probability measure under the constraint of the continuity equation. Let us denote $\mu(t,x)$ a time-dependent density on the domain $X$ assumed to be a closed convex domain in $\RR^d$ or a closed Riemannian manifold. This density follows the continuity equation which reads, for $t \in [0,1]$ and $x \in X$
\begin{equation}\label{EqContinuityEquation}
    \partial_t \mu(t,x) + \nabla \cdot (\mu(t,x) v(t,x)) = 0\,,
\end{equation}
where $\partial_t$ denotes the partial derivative w.r.t. time, $\nabla \cdot$ is the divergence w.r.t. a chosen reference volume form and $v:[0,1] \times X \to TX$ is a time-dependent vector field, taking values in the tangent space of $X$ denoted by $TX$.
Now, the optimal transport distance between $\mu(0,\cdot)$ and $\mu(1,\cdot)$ for the cost $d^2$ can be written as the minimization over $\mu(t,x)$ (with fixed endpoints in time) of the kinetic energy
\begin{equation}
    \int |v(t,x)|^2 d\mu(t,x)\,,
\end{equation}
under the constraint \eqref{EqContinuityEquation}.
Note that this formulation of the problem is a priori more complex since it introduces an additional time variable. Even worse, the initial Kantorovich problem is convex whereas this formulation is not convex.
One of the key contribution of  Benamou and Brenier is to transform it back to a convex problem via  
\begin{equation}\label{EqContinuityEquationnew}
\partial_t \mu + \nabla \cdot \omega = 0\,,
\end{equation}
with $\omega = \mu v$, which is now a measure. This change of variable  transforms the kinetic energy which is non-convex in $(\mu, v)$ into a one-homogeneous convex Lagrangian $\frac 12 \frac{\|\omega\|^2}{\mu}$. As put forward in \cite{benamou2000computational} this formulation can now be solved numerically with convex optimization algorithms.

 Note that it is not always the case that the Kantorovich problem can be re-written in such a dynamic form.
Indeed, the cost often cannot  be naturally reformulated as a Lagrangian on a space of curves in contrast to the Riemannian distance squared. However, the dynamic formulation presents several advantages such as being more flexible to incorporate physical constraints or to extend the Wasserstein metric to the space of positive Radon measures. As we see in the Section \ref{sec:uot-dynamic}, deriving metric properties of the unbalanced optimal transport problem follows easily by construction.

\myparagraph{Algorithms.}
In general, OT maps and couplings cannot be computed in closed form. 
An exception to this is when $(\al,\be)$ are Gaussians, in which case the optimal plan is an affine map and the Wasserstein$-2$ distance admits a closed form, which is the Bures distance between the covariance matrices~\cite{bures1969extension}.

Several algorithms solve exactly or approximately the OT problem between two discrete measures.
Network flow simplexes~\cite{orlin1997polynomial} solve OT for any inputs $(\al,\be)$ in $O(N^3\log N)$ time.
When measures have the same cardinality $N$ and uniform weights $1/N$, the Hungarian algorithm~\cite{kuhn1955hungarian} and the auction algorithm~\cite{bertsekas1990auction, bertsekas1992auction} compute the permutation matrix in $O(N^3)$ time. 
These methods are efficient for medium scale problems, \rev{where typically $N\lesssim 10^4$}. 

Under additional assumptions, specific (and thus more efficient) solvers are available.
For univariate data (i.e. $(\al,\be)\in\Mmp(\RR)$), if the cost reads $\C(x,y)=\h(x-y)$ with $\h:\RR\rightarrow\RR_+$ convex on $\RR_+$~\cite{santambrogio2015optimal}, optimal maps are increasing functions. In the discrete case, solving OT is thus achieved by sorting the data, which requires $O(N \log(N))$ operations. 

In the setting of semi-discrete OT where $\al$ has a density and $\be$ is discrete, and for the cost $\C(x,y)=\norm{x-y}^2$, the Monge map can be solved by leveraging the structure of this map, which is constant on so-called ``Laguerre cells'' (which generalizes Voronoi cells by including weights).
Fortunately, these cells can be computed in linear time in dimension 2 and 3, which can itself be integrated into a quasi-Newton solver to find the weights of the cells~\cite{kitagawa2019convergence}.
The Wasserstein-$1$ distance on a graph is equivalent to a min-cost-flow problem~\cite{santambrogio2015optimal} which is solvable in $O(N^2\log N)$ time by Network simplex algorithms.
Solving the dynamic formulation is also a way to address medium to large scale OT problems using proximal splitting methods, see \cite{papadakis2014optimal}.

It is also possible to build approximations or alternative distances on top of OT computations. An option is to integrate OT distance along 1-D projection, which corresponds to the so-called sliced-Wasserstein distance~\cite{bonneel2015sliced}.
Combining 1-D OT maps along the axes defines the Knothe map~\cite{santambrogio2015optimal}, which is a limit of OT maps for degenerate cost functions, and is related to the subspace detours variation of OT~\cite{muzellec2019subspace}. 
One can also approximate OT between mixtures of Gaussians by leveraging the closed form expression of OT between Gaussians~\cite{delon2020wasserstein}, which finds applications in texture synthesis~\cite{leclaire2022optimal}. 

\myparagraph{Applications.}
Some models can be directly re-casted as OT optimization problems. This is typically the case in physical models, such as for instance finding the optimal shape of a light reflector problem~\cite{glimm2010rigorous,benamou2020entropic} or reconstructing the early universe from a collection of stars viewed as point masses~\cite{frisch2002reconstruction, levy2021fast}. 

In data sciences, OT distances can be used to perform various learning tasks over histograms (often called ``bag of features''), such as image retrieval~\cite{rubner2000earth, rabin2008circular}.
OT distances are also at the heart of statistical tests~\cite{ramdas2017wasserstein, sommerfeld2018inference, hundrieser2022unifying}, to check equality of two distributions from their samples, and to check the independence by comparing a joint distribution to the product of the marginals.
More advanced uses of OT consist in using it as a loss function for imaging and learning.
In supervised learning, the features and the output of the learned function are histograms compared using the OT loss.  This can for instance be used for supervised vision tasks~\cite{frogner2015learning}. This is also used for graph predictions~\cite{vayer2020fused, petric2019got}, where the features of all nodes on a graph are aggregated into a distribution in $\RR^d$, and the OT distance serves to train an embedding of these node distributions modeled via a graph neural network~\cite{wu2020comprehensive}.
In unsupervised learning, for density fitting, one optimizes some parametric model $\be_\thh$ to fit empirical observations viewed as a discrete measure $\al_N$. This is achieved by minimizing $\thh\mapsto\OT(\al_N,\be_\thh)$.  A typical instantiation of this problem is to train a generative model~\cite{arjovsky2017wasserstein}, in which case the model is of the form $\be_\thh = (\h_\thh)_{\sharp} \beta_0$ where $\h_\thh$ is a deep network and $\beta_0$ a fixed reference measure in a low dimensional latent space.
For imaging sciences, such as in shape matching (which is similar in spirit to generative model training), the transport map cannot be directly used for registration (because it lacks regularity). 
Though the OT loss can be used to help the trained diffeomorphic models to avoid local minima of the registration energy ~\cite{feydy2017optimal}.
Other applications in imaging include inverse problems resolution, such as seismic tomography~\cite{metivier2016optimal} where the OT loss tends to reduce the presence of local minimizers when resolving the location of reflecting layers in the ground.

The OT map itself also finds applications, for instance in joint domain adaption or transfer learning~\cite{courty2016optimal,courty2017joint}, where the goal is to learn in an unsupervised setting the labels of a dataset based on another similar dataset which is labeled.
In single-cell biology~\cite{schiebinger2017reconstruction}, the optimal transport plan interpolates between two observations of a cell population at two successive time-steps.
In natural language processing~\cite{grave2019unsupervised}, the input measure of words represents a language embedded in Euclidean spaces using e.g. word2vec~\cite{mikolov2013efficient}, and assignments are related to translation processes.

\myparagraph{Challenges. } The definition of OT imposes a perfect conservation of mass, so that it can only compare probability measures. While it is tempting to use ad-hoc normalizations to cope with this constraint, using directly this formulation of OT makes it non-robust to noise and local variation of mass. We explain in the following Section~\ref{sec:uot} a simple workaround leading to an extended version of the classical OT theory. 
We then address in Section~\ref{sec:entropy} the two other major limitations of OT: its high computational cost and its high sample complexity.

\section{Unbalanced Optimal Transport}
\label{sec:uot}

\begin{figure}[h]
	\begin{subfigure}{.45\textwidth}
		\centering
		\includegraphics[width=\linewidth]{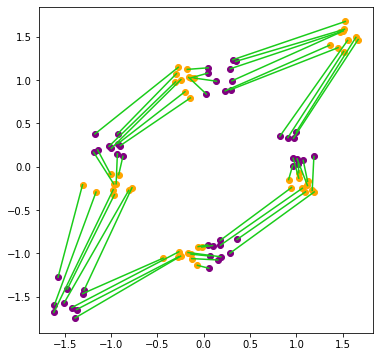}
		\caption{OT matching}
	\end{subfigure}
	\hfill
	\begin{subfigure}{.45\textwidth}
		\centering
		\includegraphics[width=\linewidth]{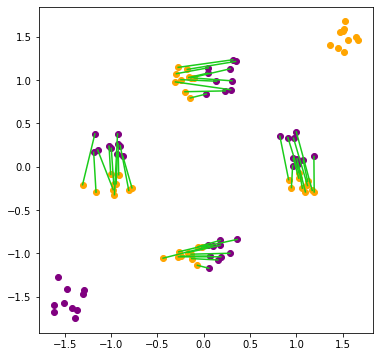}
		\caption{Unbalanced OT matching}
	\end{subfigure}
	\caption{
	\textit{Plot of the OT and unbalanced OT matching (in green) between two point clouds $(\al,\be)$ equipped with uniform distributions (in red and blue).
	Four modes are close to each other, while a fifth mode of each distribution is distant from the rest, and should thus be treated as outlier samples. For OT the matching is forced to make a correspondence between those samples, while for UOT they are indeed discarded.
	The UOT matching is performed using partial OT~\cite{figalli2010optimal}, i.e. Definition~\ref{def:uot} with $\D_\phi=\rho\TV$ and $\rho=0.25$, such that no transport happens when $\C\geq2\rho$.}
	}
	\label{fig:ot_uot_matching}
\end{figure}

The main idea to lift this mass conservation restriction is to replace the hard constraint encoded in $\Uu(\al,\be)$ by a soft penalization, leveraging Csiszàr divergence $\D_\phi$ from Section~\ref{sec:phidiv}.
As illustrated by Figure~\ref{fig:ot_uot_matching}, this constraint forces OT to transport all samples of input distributions, which is an undesirable feature in the presence of outliers, i.e.  irrelevant samples.
By comparison, relaxing such constraint as is performed by unbalanced OT (defined below) allows to discard such outliers.
We review the existing (often equivalent) ways to achieve this, leading to static, dynamic and conic formulations.

\subsection{Static formulation}
\label{sec:uot-static}

As mentioned above, Optimal Transport only has a finite value when the two inputs have the same mass, i.e. $\int_\X \dd\al = \int_\X \dd\be$.
It is imposed by the constraint set $\Uu(\al,\be)$ which implies $\int_{\X^2}\dd\pi = \int_\X \dd\al = \int_\X \dd\be$.
Thus if this property does not hold, $\Uu(\al,\be)=\emptyset$ and $\OT(\al,\be)=+\infty$.

In the last few years, several extensions of OT comparing arbitrary positive measures have been developed, based on the different equivalent formulations of OT.
The first extensions in the literature focused on the Benamou-Brenier formulation~\cite{benamou2000computational}, and relax a mass conservation constraint encoded by a continuity PDE~\cite{liero2015optimal, liero2016optimal, chizat2018interpolating, kondratyev2016fitness}.
This review focuses mostly on the one proposed by~\cite{liero2015optimal}, which consists in replacing the hard constraints $(p_\X)_\sharp\pi=\al$ and $(p_\Y)_\sharp\pi=\be$ by $\phi$-divergences $\D_{\phi_1}((p_\X)_\sharp\pi | \al)$ and $\D_{\phi_2}((p_\Y)_\sharp\pi | \be)$ (presented in Section~\ref{sec:phidiv}).
It was also proposed in~\cite{frogner2015learning} for the particular setting $\D_\phi=\KL$.
We mainly focus on it because it is convenient from a computational perspective, as we detail in Section~\ref{sec:sinkhorn}.

\begin{definition}[Static unbalanced OT]\label{def:uot}
The unbalanced OT program optimizes over transport plans $\pi\in\Mmp(\X\times\X)$, and reads
\begin{align*}
\UOT(\al,\be) \triangleq \min_{\pi\geq 0} \int_{\X^2} \C(x,y)\dd\pi(x,y) + \D_{\phi_1}(\pi_1|\al) + \D_{\phi_2}(\pi_2|\be),
\end{align*}
where $(\pi_1,\pi_2)\eqdef((p_\X)_\sharp\pi, (p_\Y)_\sharp\pi)$ are the plan's marginals.
\end{definition}

This problem is coined \hyphen{unbalanced} OT in reference to~\cite{benamou2003numerical} who first proposed to relax the marginal constraints for computational purposes.
\rev{In general $\pi_1\neq\al$ and $\pi_2\neq\be$. In the UOT problem, the marginals $(\pi_1,\pi_2)$ are interpreted as the transported mass, while the ratio of density between e.g. $\pi_1$ and $\al$ corresponds to the mass which is destroyed and created, i.e. ``teleported'' instead of being transported.}

\begin{figure}[h]
	\begin{subfigure}{.31\textwidth}
		\centering
		\includegraphics[width=\linewidth]{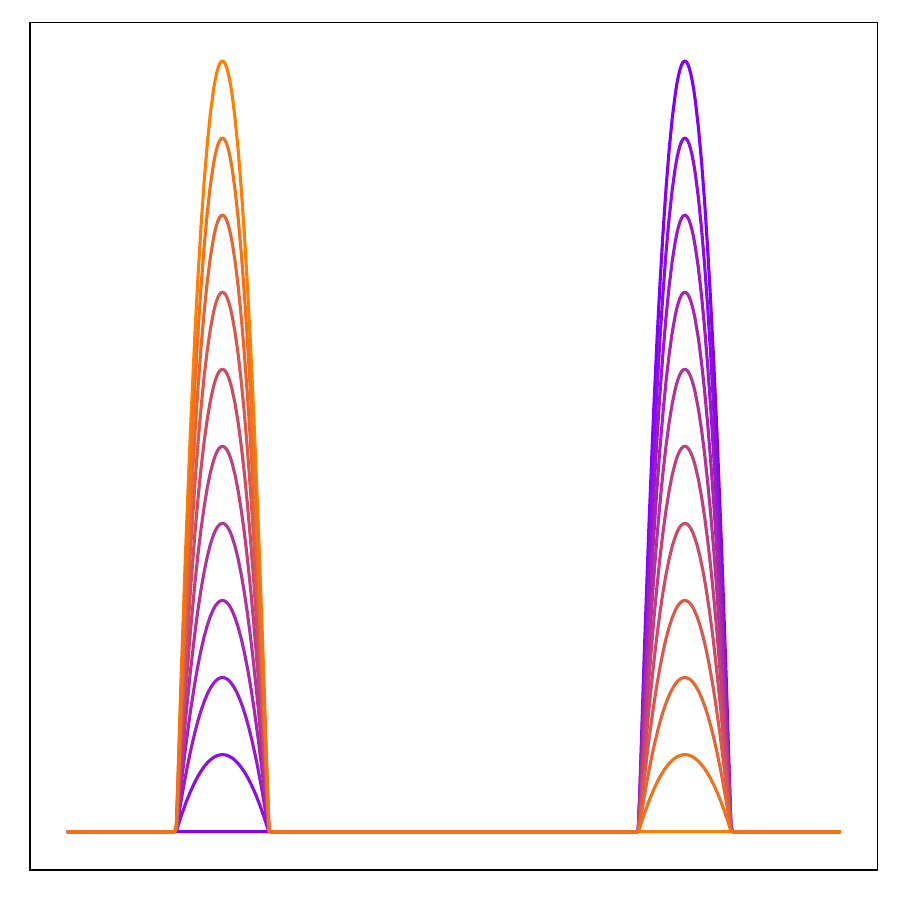}
		\caption{$\Ll = \D_\phi=\KL$}
	\end{subfigure}%
	\hfill
	\begin{subfigure}{.31\textwidth}
		\centering
		\includegraphics[width=\linewidth]{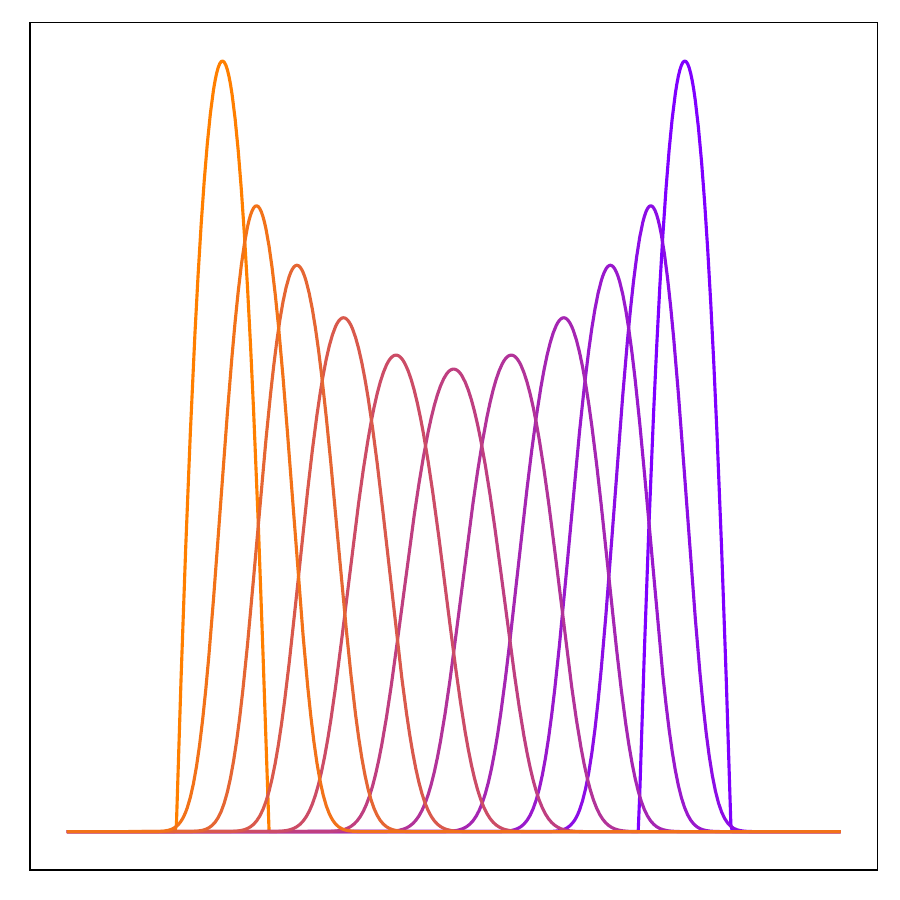}
		\caption{$\Ll = \UOT$ (same $\C$ and $\D_\phi$)}
	\end{subfigure}%
	\hfill
	\begin{subfigure}{.31\textwidth}
		\centering
		\includegraphics[width=\linewidth]{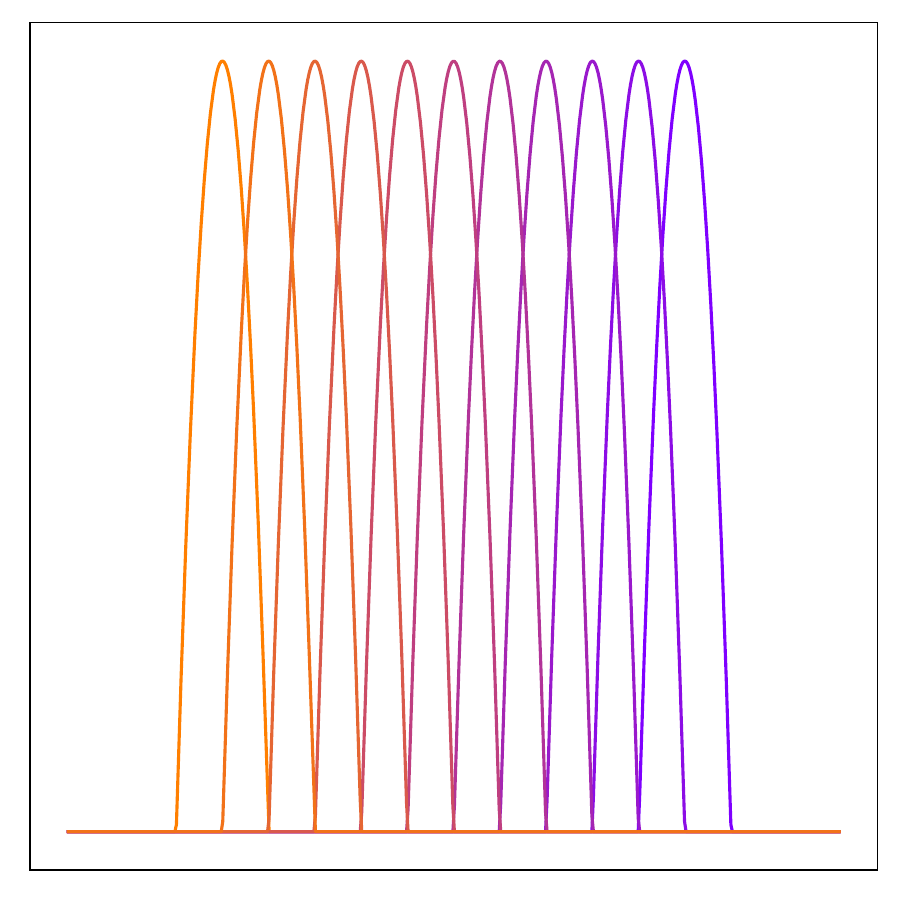}
		\caption{$\Ll = \OT^2$ ($\C=\norm{\cdot-\cdot}_2^2$)}
	\end{subfigure}
	\caption{\textit{
		Display of the barycentric interpolation between two (compactly supported) parabolic distributions. It is defined as $\be^\star=\arg\min_{\be\in\Mmpo(\X)} t\Ll(\be,\al_0) + (1-t)\Ll(\be,\al_1)$ for $t\in[0,1]$.}
	}
	\label{fig:interp-gaussian}
\end{figure}

Note that we retrieve $\OT$ as a particular instance of this problem with $\phi = \iota_{\{1\}}$ (i.e. $\phi(1)=0$ and $+\infty$ otherwise), so that $\D_\phi=\iota_{(=)}$ satisfies $\iota_{(=)}(\pi_1|\al)=0$ if $\pi_1=\al$ and $+\infty$ otherwise.
Unbalanced OT is interpretable as a combination of the comparison of masses performed by Csisz\'ar divergences with the comparison of supports operated by OT.
Figure~\ref{fig:interp-gaussian} illustrates this informal intuition in the setting of barycentric interpolation between two compactly supported parabolic densities in $\X=\RR$.

It is possible to tune the strength of the mass conservation, i.e. if we prefer to have $\pi_1\simeq\al$ or allow $\pi_1$ to differ significantly from $\al$.
To do so one can add a parameter $\rho >0$ and use $\D_{\rho\phi}=\rho\D_\phi$.
One retrieves balanced OT when $\rho \rightarrow\infty$ (provided $\phi^{-1}(\{0 \})=\{ 1 \}$).

\myparagraph{Influence of the unbalancedness parameter $\rho$.}
\rev{
Tuning the parameter $\rho$ might be challenging in ML tasks.
It corresponds to a characteristic radius of transportation (when $\D_\phi=\rho\TV$ there is no transportation if $\C_{i,j}>2\rho$), thus its value should be adapted to the cost matrix $\C$.
In ML tasks where the cost $\C$ is learned or corresponds to a deep feature space, the lack of interpretability complicates the choice of $\rho$, and thus imposes the use of grid-search to tune it.
}

\begin{figure*}[h]
	\centering
	\includegraphics[width=\linewidth]{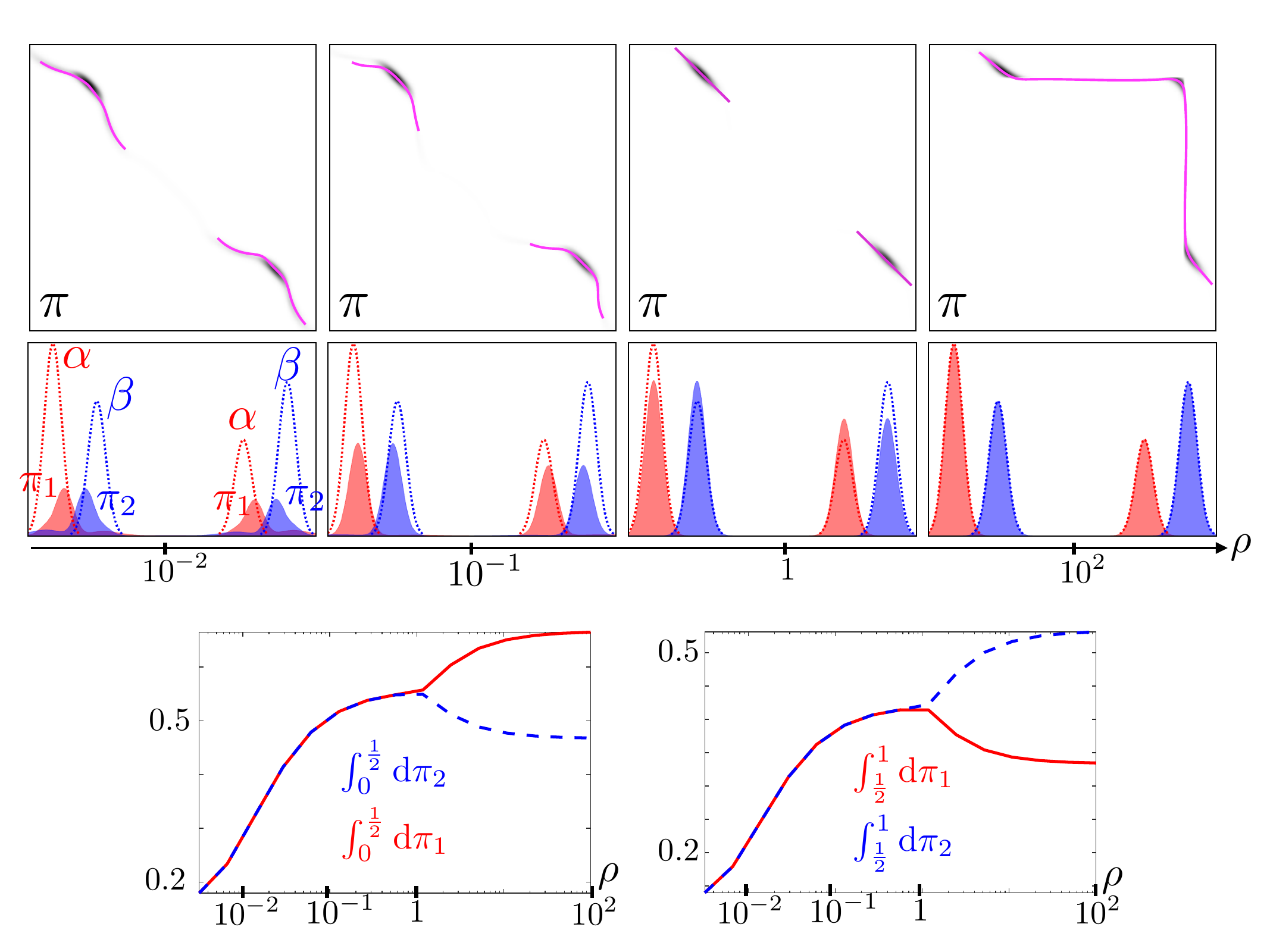}
	\caption{\rev{Influence of the relaxation parameter $\rho$ when using $\rho \KL$ relaxation of the marginals' constraints.}}
	\label{fig-influ-param}
\end{figure*}

\rev{Figure~\ref{fig-influ-param} shows the influence of the relaxation parameter $\rho$ on the optimal plan $\pi$ and on its  marginals $(\pi_1,\pi_2)$. 
The computation are done using Sinkhorn's algorithm detailed in Section~\ref{sec:entropy}, which induces a small diffusion (blurring) of the transport plan $\pi$ (the overlaid purple curve shows an approximation of the transport map extracted from this approximated plan).
The goal is to highlight the trade-off induced by this parameter selection process, which is crucial for the successful deployment of unbalanced OT in practice. The two inputs $(\al,\be)$ (displayed in dashed lines) are two 1-D mixtures of two Gaussians, but the modes' amplitudes are not equal. The purpose of this toy example is to illustrate the impact on balanced OT of the mass discrepancy of nearby modes, which forces the optimal transport map to be irregular and to split the mass of each mode. This effect can be seen on the top row, where for $\rho>1$, mass conservation forces the plan $\pi$ to be non-regular. On the contrary, for $\rho<1$, the plan is regular and is locally a translation of the modes, as one should expect.

The middle row shows the evolution of the relaxed marginals $(\pi_1,\pi_2)$ as $\rho$ increases. The bottom row gives a quantitative assessment of the evolution: it displays the evolution of the mass of each mode (i.e. on $[0,1/2]$ and $[1/2,1]$).  This highlights that as soon as $\rho<1$, neighboring modes have exactly the same mass. This allows the transport plan $\pi$ to match almost exactly the modes using a regular map (almost a translation), which is the desired effect in practice. For $\rho<0.1$, an undesirable effect kicks in: the mass of the marginals becomes too small, and they start to shift from their original location, creating a bias in the transport plan. One can conclude from this numerical analysis that a wide range of $\rho \in [0.1,1]$ leads to a precise estimate of the transportation between the modes of the mixtures, each one being relaxed to modes with constant mass consistent with the original inputs.

}

\myparagraph{Applications.}
Unbalanced OT often improves over the results obtained using balanced OT in most of its application to data sciences.
\rev{All these examples are also instructive to understand how the parameter $\rho$ should be tuned to reach good performances on large scale datasets. }
An example in image processing is for video registration using the UOT plan~\cite{lee2019parallel}.
In shape registration for medical imaging~\cite{feydy2019fast}, biology~\cite{schiebinger2017reconstruction, wu2022metric} and in LiDAR imaging~\cite{bevsic2022unsupervised, cattaneo2022lcdnet}, UOT is used to provide assignments robust to acquisition noise.
UOT can also be used for the theoretical analysis of some statistical and machine learning methods. 
It is possible to express the gradient descent training of the neurons of a two layers perceptron as a UOT gradient flow, and prove its convergence to global optimality~\cite{chizat2018global, rotskoff2019global}.  
One can also state stability guarantees on the reconstruction of sparse measures in super-resolution problems using UOT~\cite{poon2021geometry}. 


\subsection{Dynamic formulation}
\label{sec:uot-dynamic}

\newcommand{\growth}{\zeta}

In this section we present a dynamic formulation of unbalanced optimal transport, based on \cite{chizat2018unbalanced}. This time dependent formulation consists in an optimal control problem on the space of nonnegative Radon measures.
\par
This model is a generalization of the Benamou-Brenier formula presented in the previous section. In order to account for mass change, one considers a continuity equation with a source term. With the notations of the previous section, we introduce
\begin{equation}
    \partial_t \mu(t,x) + \nabla \cdot (\mu(t,x) v(t,x)) = \growth(t,x) \mu(t,x)\,,
\end{equation}
where $\growth: [0,1] \times X \to \RR$ can be understood as a growth rate.
One considers a Lagrangian of the type
\begin{equation}\label{EqLagrangianL}
   \int_0^1 \int_X  L(x,v(t,x),\growth(t,x)) \dd\mu(t,x)\,.
\end{equation}
Recall that the case of the Benamou-Brenier formula is when $\alpha$ is constrained to be $0$ and $L(x,v) \triangleq \frac 12 \| v \|^2$ for the corresponding norm on $TX$.
Similarly, this problem has a convex reformulation by introducing
\begin{equation}\label{EqContinuityEquation-UOT}
\partial_t \mu + \nabla \cdot \omega = \xi\,,
\end{equation}
with $\omega \triangleq \mu v$ and $\xi \triangleq \growth \mu$, which are now measures. 
A quite general setting for dynamic formulations of unbalanced optimal transport is the following minimization problem
\begin{equation}\label{EqDefDynamic}
 C_D(\alpha,\beta) = \min_{\mu,\omega,\xi}   \int_0^1 \int_X f\left(x,\frac \mu \lambda,\frac \omega \lambda,\frac \xi \lambda\right) \dd\lambda(t,x)\,
\end{equation}
under the continuity equation constraint \eqref{EqContinuityEquation-UOT} and the time-boundary constraints $\mu(t=0,\cdot) = \alpha$ and $\mu(t=1,\cdot) = \beta$. Here, we consider
 a function $f$ such that $f(x,\cdot,\cdot,\cdot)$ is nonnegative, convex, positively $1-$homogeneous\footnote{We refer the reader to \cite{chizat2018unbalanced} for further details on the function $f$.}. In comparison with the Lagrangian in Equation \eqref{EqLagrangianL}, we incorporate $\mu$ in the Lagrangian and we impose its $1-$homogeneity. In addition, $\lambda(t,x)$ stands for a measure in time and space dominating all other measures $\mu,\omega$ and $\xi$ and due to the hypothesis on $f$, Definition \eqref{EqDefDynamic} does not depend on the choice of $\lambda$. 
 \par
 From the continuity equation with a source term, it is clear that simultaneous displacement of mass and change of mass (destruction or creation) is made possible. This is what happens once the optimization has been performed, although the behaviour depends on the chosen Lagrangian.
One important example of Lagrangian is, with a little abuse of notations on the variables, 
$f_{\operatorname{WFR}}(\mu,\omega,\xi) = \frac 1{2\mu} (\| \omega \|^2 + \delta^2 \xi^2)$ for a positive parameter $\delta$ which leads to the Wasserstein-Fisher-Rao metric, also called Hellinger-Kantorovich. 
This particular example gives rise to a static formulation with the Kullback-Leibler as the divergence and $c_{\operatorname{WFR}}(x,y) = -\log(\cos^2(\min(\frac {1}{2\delta}\| x - y\|,\frac \pi 2) ))$ as the transportation cost in Definition \ref{def:uot}.
Another example is $f(\mu,\omega,\xi) = \frac {1}{p} \frac{\| \omega\|^p}{\mu^{p-1}} + \delta | \xi|$ for a real parameter $\delta >0$ and it is directly related to partial optimal transport \cite{figalli2010optimal}. 
In fact, if $(v,g) \mapsto f(x,1,v,g)$ is $p-$positively homogeneous for some $p>1$ and symmetric w.r.t. the origin, then $C_D^{1/p}$ is a metric on the space of nonnegative Radon measures.
\par
The main result of this section is that Formulation \eqref{EqDefDynamic} is equivalent to a static formulation for a cost $c$ that is the convexification of the following cost defined on the cone $X \times \RR_{+}$ (see next Section~\ref{sec:uot-conic} for more information on the cone)
\begin{equation}
    \tilde c((x_0,m_0),(x_1,m_1)) = \inf_{x(t),m(t)} \int_0^1 f(x(t),m(t),m(t)x'(t),m'(t)) dt\,,
\end{equation}
where the optimization variable is an absolutely continuous path $(x(t),m(t))$ between $(x_0,m_0)$ and $(x_1,m_1)$.
The convexification of $\tilde c$ is $ c((x_0,m_0),(x_1,m_1))\eqdef \min_{m_0^i,m_1^i} \tilde c((x_0,m_0^1),(x_1,m_1^1)) + \tilde c((x_0,m_0^2),(x_1,m_1^2))$ where $m_0^1 + m_0^2 = m_0$ and $m_1^1 + m_1^2 = m_1$.   
We now introduce a static formulation called semi-couplings introduced in \cite{chizat2018global} and used in \cite{bauer2021square} to derive an explicit algorithm.
Semi-couplings are couples of nonnegative Radon measures $\gamma,\pi$ on $X \times X$ such that $\gamma_1 = \alpha$ and $\pi_2 = \beta$.
The static problem is the minimization over the set of semi-couplings of 
\begin{equation}\label{eq:semi_coupling_integral}
  C_K(\gamma,\pi) =  \inf_{\gamma,\pi}\int_{X \times X}  c\left((x,\frac{\gamma}{\lambda}(x,y)),(y,\frac{\pi}{\lambda}(x,y))\right) d\lambda(x,y)\,,
\end{equation}
for $\lambda$ any measure dominating $\gamma,\pi$. 
Again, if for some $p\geq 1$ the cost $c$ is such that $c^{1/p}$ is a distance on the cone then $C_K^{1/p}$ is a distance on the space of positive Radon measures on $X$.
\begin{theorem}{(Informal, from \cite{chizat2018unbalanced})}
    The dynamic  and the semi-coupling formulations give the same value, i.e. $C_D(\alpha,\beta) = C_K(\alpha,\beta)$.
\end{theorem}
We refer to \cite{chizat2018unbalanced} for the dual problem of dynamic and static formulations. Using the dual formulation of the static problem one proves that the semi-coupling formulation has an equivalent UOT formulation. Let us detail the Wasserstein-Fisher-Rao metric case for which $f_{\operatorname{WFR}}$ is defined above. Denoting $C_{\operatorname{WFR}}$ the corresponding dynamic cost,
$
    C_{\operatorname{WFR}}(\alpha,\beta) = 2\delta^2 \UOT(\al,\be)
$
in which $\UOT$ is used with the cost $c_{\operatorname{WFR}}$ and the Kullback-Leibler divergences defined in Formula \eqref{EqKLDef}.
To conclude, we underline that the dynamic cost often gives a length space on the space of positive Radon measures; the most important example being the Wasserstein-Fisher-Rao metric presented above which is an equivalent of the Wasserstein $L^2$ metric to the space of positive Radon measure.

\subsection{Conic formulation}
\label{sec:uot-conic}

This section focuses on another formulation of unbalanced OT, which is proved in~\cite{liero2015optimal} to be equivalent to the UOT formulation (Definition~\ref{def:uot}).
This second formulation is useful to prove metric properties of UOT.
It was adapted to the setting of Gromov-Wasserstein distances to extend it for the unbalanced setting (see Section~\ref{sec:ugw} or~\cite{sejourne2021unbalanced}).

We detail below the principle of the conic formulation, as well as its derivation to provide understanding on its link with UOT.
Then we present its metric properties, and review explicit settings existing in the literature.

\begin{figure}
	\centering
  \includegraphics[width=0.5\linewidth]{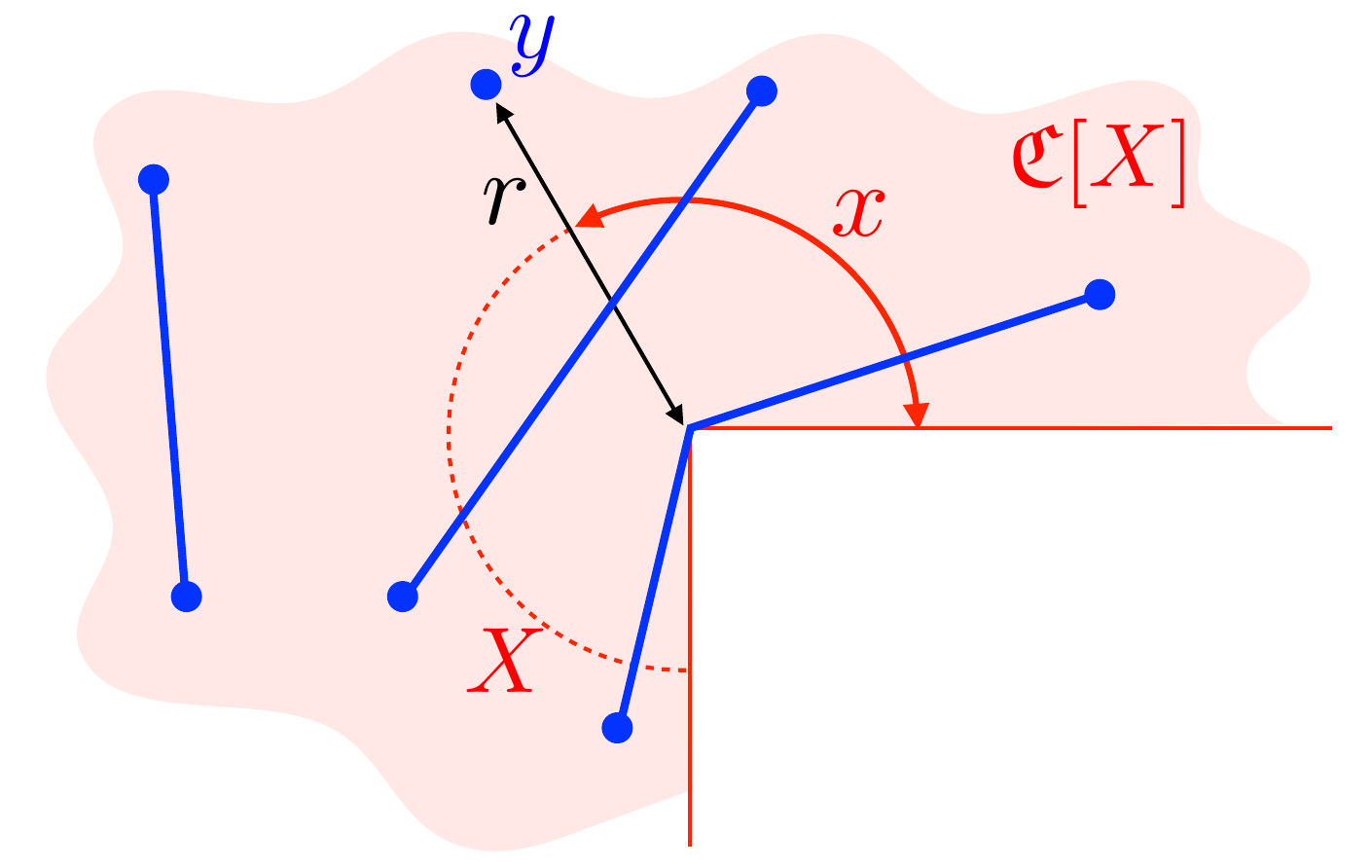}
  \caption{
  \rev{Example of cone $\Co[\X]$, where $\X$ is a piece of circle whose angles $x$ lie in $[0, \tfrac{3\pi}{2}]$, endowed with the ground (geodesic) distance $d_\X(x_1,x_2)=|x_1-x_2|$ for $(x_1,x_2)\in\X$.
  The cone $\Co[\X]$ is represented in red and embedded in $\RR^2$ via $(y_1,y_2)=(r\cos x, r\sin x)$.
  Three geodesics in $\Co[\X]$ are represented in blue.
  Note the bottom geodesic has a shortest path passing via the cone's apex.}
  }
  \label{fig:example-cone}
\end{figure}

\myparagraph{Conic lifting.}
Informally, this lifting corresponds to map a weighted Dirac mass $r\de_x\in\Mmp(\X)$ to $\de_{(x,r)}$, where $(x,r)\in\X\times\RR_+$ and $r$ is the particle's mass.
This lifts the space $\X$ into $\X\times\RR_+$, more precisely it is lifted onto the cone $\Co[\X]\triangleq(X\times\RR_+)/\equiv$, where the equivalence relation on $\X\times\RR_+$ reads $[x,r]\equiv[y,s]\Leftrightarrow (r=s=0)\,\,\text{or}\,(x=y \,\,\text{and}\,\, r=s)$.
Note that having $r=s=0$ corresponds to comparing two particles with no mass, thus their positions do not matter and are considered to be equal w.r.t. $\equiv$.
Geometrically, such samples are located at the apex of the cone, and the notation $[x,r]\in\Co[\X]$ emphasizes the quotient at the apex (compared to $(x,r)\in\X\times\RR_+$).
See Figure~\ref{fig:example-cone} for an illustrated example of a cone set.

\myparagraph{Lifting positive measures on a cone.}
Given a positive measure $\al\in\Mmp(\X)$ one can lift it as a measure defined over $\Co[\X]$ (i.e. in $\Mmp(\Co[\X])$).
One admissible lift of $\al$ is $\margc_\sharp(\al\otimes\de_1)$, where $\margc(x,r)\triangleq[x,r]$ is the canonical injection from $\X\times\RR_+$ to $\Co[\X]$.
In the case of a Dirac $r\de_x$ this lift is $r\de_{[x,1]}$.
However, more elaborate lifts are possible, and play a key role in the analysis of metric properties.
In particular, one can show that it is always possible to lift a positive measure of $\Mmp(\X)$ to a probability distribution in $\Mmpo(\Co[\X])$, in order to re-cast unbalanced OT problems over $\Mmp(\X)$ as balanced OT problems over $\Mmpo(\Co[\X])$, with respect to a new cost $d_\Co$ \rev{(see~\cite[Corollary 7.7]{liero2015optimal})}.

\myparagraph{Deriving the conic formulation - Homogeneous formulation.}
\rev{We detail in this paragraph how one can reformulate the UOT problem (Definition~\ref{def:uot}) to connect it with its equivalent conic formulation.
This derivation involves an intermediate formulation which we call \emph{homogeneous}.
To the best of our knowledge, this formulation has not yet been considered for applications in the literature.
However, it is a key intermediate problem to clarify the equivalence between UOT and the conic formulation.}

The derivation consists in decomposing Csisz\'ar entropies $\D_\phi$ and refactoring all terms involving $\pi$ as a single integral.
To do so we define the \emph{reverse entropy} as $\psi(r)\triangleq r \phi(\tfrac{1}{r})$ for $r>0$, $\psi(0)=\phi^\prime_\infty$ and $\psi^\prime_\infty=\phi(0)$.
The reverse entropy is such that $\D_\phi(\pi_1|\al) = \D_\psi(\al|\pi_1)$.
Before rewriting the $\UOT$ formulation, we define the following Lebesgue decompositions of $\al$ w.r.t $\pi_1$ and $\be$ w.r.t. $\pi_2$, which read 
\begin{equation}
\al = \xi_1 \pi_1 + \al^\bot \qandq \be = \xi_2 \pi_2 + \be^\bot,  \label{eq:leb-dens-1}
\end{equation}
such that e.g.
\begin{align*}
    \D_\phi(\pi_1|\al) = \D_\psi(\al|\pi_1) = \int_\X \psi\big(\xi_1(x)\big)\dd\pi_1(x) + \psi^\prime_\infty\int_\X \dd\al^\bot(x).
\end{align*}
\rev{Write $\Pp_0$ the primal functional of Definition~\ref{def:uot}. Its reformulation reads}
\begin{align*}
	\Pp_0(\pi) &= \int_{\X^2} \C(x,y)\dd\pi(x,y) + \D_{\phi_1}(\pi_1|\al) + \D_{\phi_2}(\pi_2|\be)\\
	&= \int_{\X^2}[ \C(x,y) + \psi(\xi_1(x)) + \psi(\xi_2(y))]\dd\pi(x,y)\\ 
	&\quad+ \psi^\prime_\infty\int\dd\al^\bot + \psi^\prime_\infty\int\dd\be^\bot\\
	&= \int L_{\C(x,y)}(\xi_1(x), \xi_2(y))\dd\pi(x,y)\\
	&\quad+\psi^\prime_\infty(|\al^\bot| + |\be^\bot|),
\end{align*}
where $L_{c}(r,s) \eqdef c + \psi(r) + \psi(s)$, $|\al^\bot|\eqdef\al^\bot(\X)$ and $|\be^\bot|\eqdef\be^\bot(\X)$.
The above formulation is helpful to explicit the terms of pure mass creation/destruction $(|\al^\bot| + |\be^\bot|)$, and reinterpret the integral under $\pi$ as a transport term with a new cost $L_{\C}$ accounting for partial transport and destruction/creation of mass.

The conic formulation is motivated by a pre-optimization of the plan's mass, i.e. optimizing over plans $\thh\pi$ where $\thh\in\RR_+$.
The plan $\thh\pi$ has marginals $(\thh\pi_1,\thh\pi_2)$, the Lebesgue densities of Equation~\eqref{eq:leb-dens-1} become $(\tfrac{\xi_1}{\thh},\tfrac{\xi_2}{\thh})$ and $(\al^\bot,\be^\bot)$ are unchanged, which yields
\begin{align*}
	\Pp_0(\thh\pi) = \int \thh L_{\C(x,y)}(\tfrac{\xi_1(x)}{\thh}, \tfrac{\xi_2(y)}{\thh})\dd\pi(x,y) +\psi^\prime_\infty(|\al^\bot| + |\be^\bot|).
\end{align*}
Note the scaling $\thh$ performs a perspective transformation on the function $L_\C$.

The homogeneous formulation is obtained by optimizing over scalings $\thh(x,y)$ in $\RR_+$ the function $L_\C$ (but not the functional $\Pp_0$).
Informally speaking, the mass is rescaled at a local scale (i.e. for any $(x,y)$) instead of globally optimizing the plan's mass (with a constant $\thh\in\RR_+$).
Define the 1-homogeneous function $H_c$ as the perspective transform of $L_c$ 
\begin{align}\label{eq:persp_hc}
	H_c(r, s) \eqdef \inf_{\theta\geq 0} \theta\big( c + \psi(\tfrac{r}{\theta}) + \psi(\tfrac{s}{\theta} ) \big) 
	= \inf_{\theta\geq 0} \theta L_c(\tfrac{r}{\theta}, \tfrac{s}{\theta}).
\end{align}
The function $H_c$ can be computed in closed form for a certain number of common entropies $\phi$.
\rev{We detail explicit settings in Section~\ref{sec:conic-setups}, and we refer to~\cite[Section 5]{liero2015optimal} for a broader overview.}
Eventually one defines another formulation called \emph{homogeneous}. It reads
\begin{equation}
	\begin{aligned}\label{eq:uw-homog}
	\HOT(\al,\be) \eqdef \uinf{\pi \in \Mm(\X^2)} &\int H_{\C(x,y)}(\xi_1(x), \xi_2(y))\dd\pi(x,y)\\
	&+\psi^\prime_\infty(|\al^\bot| + |\be^\bot|),
	\end{aligned}
\end{equation}
The term \hyphen{homogeneous} stems from the $1$-homogeneity of $H_\C$ in $(r,s)$.
By definition one has $L_c\geq H_c$, so that $\UOT\geq\HOT$.
Thanks to the convexity of these problems one actually has $\UOT=\HOT$ (see~\cite[Section 5]{liero2015optimal}).

\myparagraph{Definition of the conic formulation.}
The \emph{conic} formulation consists in rewriting the program $\HOT$ (Equation~\eqref{eq:uw-homog}) over the cone.
Instead of a transport plan $\pi(x,y)\in\Mmp(\X^2)$, we consider a plan $\eta([x,r], [y,s])\in\Mmp(\Co[\X]^2)$ matching points defined over the cone.
The formulation involves a cone cost $d_{\Co[\X]}$ which lifts the cost $\C$ over $\Co[\X]^2$, and is derived from the function $H_c$.
One replaces the densities $(\xi_1,\xi_2)$ by radial coordinates $(r,s)$ in $\HOT$.
This change of variables allows to cancel out the term $(|\al^\bot| + |\be^\bot|)$, and replace it by two \emph{conic} marginal constraints reminiscent of the set $\Uu(\al,\be)$ of balanced OT.
It yields the following definition of the conic formulation.

\begin{definition}\label{def:cot}
The conic formulation is defined as
\begin{align}\label{eq:def-cone-pb}
	\COT(\al,\be) \triangleq \min_{\eta\in\Uu^\Co(\al,\be)}\int_{\Co[\X]^2} d_{\Co[\X]}([x,r],[y,s])^q\dd\eta([x,r],[y,s]),
\end{align}
where the cone cost $d_{\Co[\X]}$ is defined using the function $H_c$ (Equation~\eqref{eq:persp_hc}), and involves exponents $(p,q)\in\RR_+$ (larger than $1$), such that 
\begin{align}\label{eq:def-cone-cost}
	d_{\Co[\X]}([x,r], [y,s])^q \eqdef H_{\C(x,y)}(r^p, s^p),
\end{align}
and where the conic constraints $\Uu^\Co(\al,\be)$ are defined as 
\begin{align}\label{eq:def-conic-set}
	\enscond{ \eta\in\Mm_+(\Co[\X]^2) }{
		\int_{\RR_+} r^p \dd\eta_1([\cdot,r])=\al,
		\int_{\RR_+} s^p \dd\eta_2([\cdot,s])=\be
	}.
\end{align}
\end{definition}
Here $\eta_1,\eta_2 \in\Mmp(\Co[\X])$ denote the marginals of $\eta$ by integrating over one cone coordinate $[x,r]$ or $[y,s]$.
If one takes a test function $\xi\in\Cc(\X)$, the first constraint reads $\int r^p\xi(x)\dd\eta_1([x,r]) = \int\xi(x)\dd\al(x)$.
The additional constraints on $(\eta_1,\eta_2)$ mean that the lift of the mass on the cone must be consistent with the total mass of $(\al,\be)$.
We emphasize that the conic lifts satisfy an invariance to some operation called \emph{dilations}, such that optimal plans can only be unique up to such transformation.

The optimization program defining COT thus consists in minimizing the Wasserstein distance $\OT_{d_{\Co[\X]}}(\eta_1,\eta_2)$ on the cone $(\Co[\X], d_{\Co[\X]})$. 
As we show below, when $d_{\Co[\X]}$ is a distance, COT inherits the metric properties of $\OT_{d_{\Co[\X]}}$.
\begin{proposition}[From~\cite{liero2015optimal} and~\cite{de2019metric}]\label{prop-equiv-cone-formulation}
	One has $\UOT=\COT$, which are symmetric, positive and definite if $d_{\Co[\X]}$ is definite.
	Furthermore, if $(\X,d_\X)$ and $(\Co[\X], d_{\Co[\X]})$ are metric spaces with $\X$ separable, then $\Mm_+(\X)$ endowed with $\COT$ is a metric space.
\end{proposition}
This proposition emphasizes that the metric properties of $d_{\Co[\X]}$ are key to obtain metric properties of COT and UOT. We detail explicit settings where it holds in Section~\ref{sec:conic-setups}.

\myparagraph{Computation of COT.}
It seems a priori harder to solve this conic problem than the original UOT problem, because of the extra radial coordinates $(r,s)\in\RR_+$.
However it is proved in~\cite[Proposition 3.9]{liero2016optimal} that among optimal conic plans $\eta^\star$ of $\COT(\al,\be)$ (recall from the previous paragraph there is not uniqueness), there exists a plan which can be expressed using the optimal plan $\pi^\star$ for $\UOT(\al,\be)$.
This optimal plan has a single Dirac along each radial coordinate, i.e. it is such that $\eta^\star$-a.e., the radiuses are functions $(r(x),s(y))$ of the ground space variables $(x,y)\in\X^2$.
This means that solving the $\COT$ problem reduces to the computation of a UOT problem followed by a conic lifting.

\subsubsection{Examples of cone distances}
\label{sec:conic-setups}
As highlighted by Proposition~\ref{prop-equiv-cone-formulation}, the settings where $d_{\Co[\X]}$ is a distance on $\Co[\X]$ are of particular interest.
\rev{
In this section we write the cost as $\C(x,y)=\Om(d_\X(x,y))$, where $d_\X$ is a distance on $\X$, so that $\Om:\RR_+\rightarrow\RR_+$ plays the role of a renormalization that transfers metric properties from $\X$ to $\Co[\X]$.
}
In general $d_{\Co[\X]}$ is not a distance, but it is always definite on $\Co[\X]$, provided $d_\X$ is definite on $\X$, $\Om^{-1}(\{0\})=\{0\}$ and $\phi^{-1}(\{0\})=\{1\}$, as proved in~\cite[Theorem 7]{de2019metric}.
Given its definition (Equation~\ref{eq:def-cone-cost}), the metric properties depend on the choice of $(\phi,\Om, p, q)$. 
We detail four particular settings where this is the case. 
In each setting we provide $(\D_\phi, \Om, p, q)$ and its associated cone distance $d_{\Co[\X]}$.

\myparagraph{Gaussian Hellinger distance.}
It corresponds to
\begin{align*}
	\D_\phi = \rho\KL, \quad \Om(t) = t^2 \qandq q=p=2,\\
	d_{\Co[\X]}([x,r], [y,s])^2 = \rho\big[ r^2 + s^2 - 2rse^{-d_\X(x,y)^2 / 2\rho} \big],
\end{align*}
in which case it is proved in~\cite{liero2015optimal} that $d_{\Co[\X]}$ is a cone distance.

\myparagraph{Hellinger-Kantorovich / Wasserstein-Fisher-Rao distance.}
It reads
\begin{align*}
	\D_\phi = \KL, \quad \Om(t) = -\log\cos^2(\min(\tfrac{\pi}{2}, t)) \qandq q=p=2,\\
	d_{\Co[\X]}([x,r], [y,s])^2 = r^2 +s^2 - 2rs\cos(\min(\tfrac{\pi}{2}, d_\X(x,y))),
\end{align*}
in which case it is proved in~\cite{burago2001course} that $d_{\Co[\X]}$ is a cone distance.
The weight $\Om(t) = -\log\cos^2(\min(\tfrac{\pi}{2}, t))$ might seem more peculiar but it arises from the dynamical formulation of section \ref{sec:uot-dynamic}. The Wasserstein-Fisher-Rao metric is actually the length space induced by the Gaussian-Hellinger distance (if the ground metric $d_\X$ is itself geodesic), as proved in~\cite{liero2016optimal,chizat2018interpolating}.
This weight introduces a cut-off, because $\Om(d_\X(x,y))=+\infty$ if $d_\X(x,y)>\pi/2$. There is no transport between points too far from each other. The choice of $\pi/2$ is arbitrary, and can be modified by scaling $t \mapsto \Om(t/\rho)$ for some cutoff $\rho>0$. \rev{Note it differs slightly from taking $\D_\phi = \rho\KL$.}

\myparagraph{Power entropy distance.}
The distance induced by $\D_\phi=\KL$ is extended in~\cite{de2019metric} to power entropies $\D_\phi$ with $\phi(t)=\rho\frac{t^k - k(t-1)-1}{k(k-1)}$, where $k \notin \{0,1\}$.
The Kullback-Leibler divergence is retrieved when $k=1$.
When $\Om(t) = t^2$, $p=1$, $q=2$ and $k>1$, they prove that the induced cone cost $d_{\Co[\X]}$ is a metric over $\Co[\X]$ which reads
\begin{align*}
		d_{\Co[\X]}([x,r], [y,s])^2 = \frac{2\rho}{k} \bigg[ \frac{r + s}{2} - \big(\tfrac{r^{1-k} + s^{1-k}}{2}\big)^{\frac{1}{1-k}} \bigg(1 + (1-k) \tfrac{d_\X^2(x,y)}{2\rho}\bigg)_+^{\frac{k}{k-1}}  \bigg].
\end{align*}
We refer to~\cite{de2019metric} for variants of parameterized power-entropies which define cone metrics.
A Sinkhorn algorithm for these entropies is derived in~\cite{sejourne2019sinkhorn}, see also Section~\ref{sec:exmpl-phidiv}.

\myparagraph{Partial optimal transport.}
It corresponds to
\begin{align*}
	\D_\phi = \rho\TV, \quad \Om(t)=t^q \qandq q\geq 1 \qandq p=1,\\
	d_{\Co[\X]}([x,r], [y,s])^q = \rho\big[ r + s - \min(r,s)(2-\tfrac{d_\X(x,y)^q}{\rho})_+ \big],
\end{align*}
in which case it is proved in~\cite{chizat2018unbalanced} that $d_{\Co[\X]}$ is a cone distance.
The case $\D_\phi=\TV$ is equivalent to partial unbalanced OT, which produces discontinuities (because of the non-smoothness of the divergence) between regions of the supports which are being transported and regions where mass is being destroyed/created.
Note that~\cite{liero2015optimal} does not mention that such cone cost $d_{\Co[\X]}$ defines a distance, although it can be proved without a conic lifting that partial OT defines a distance as explained in~\cite{chizat2018unbalanced}.

\subsection{Discussion and synthesis on UOT formulations}

\rev{To conclude this exposition of the various formulations of unbalanced OT, we provide below important remarks on the different formulations, their advantages and shortcomings, as well as their relations.

\myparagraph{Relationships. } The conic formulation is more general  than the static and the dynamic one. Indeed, as formalized by Equations~\eqref{eq:persp_hc} and~\eqref{eq:semi_coupling_integral}, for any static and dynamic formulation, one can associate a cone cost and thus a conic formulation. However, one can imagine costs on the cone which are not derived from such settings, hence the broader generality of conic formulations. 

The static and the dynamic formulations are not completely equivalent. Dynamic formulations induce geodesic distances, but not all static formulations satisfy this property. Reciprocally, not all dynamic problems can a priori be rephrased as a static one for which the cost is explicit, such as for e.g. a Lagrangian with a term $|\xi|^p$ for $p\notin\{1,2\}$.
There is however an overlap between the static and the dynamic formulations. The two well-known settings are the Wasserstein-Fisher-Rao/Hellinger-Kantorovich distance and Partial OT. The former has a dynamic Lagrangian $f_{\operatorname{WFR}}(\mu,\omega,\xi) = \frac 1{2\mu} (\| \omega \|^2 + \delta^2 \xi^2)$, corresponding to the cost $\C(x,y) = -\log(\cos^2(\min(\frac {1}{2\delta}\| x - y\|,\frac \pi 2) ))$ and $\D_\phi=\KL$. The latter has Lagrangian $f(\mu,\omega,\xi) = \frac {1}{p} \frac{\| \omega\|^p}{\mu^{p-1}} + \delta | \xi|$, corresponding to a static setting $\C(x,y)=\| x - y\|^p / p$ and $\D_\phi=\delta\TV$. Recall also that both settings are equivalent to the third conic formulation, see Section~\ref{sec:conic-setups} for formulas.

These two settings also share equivalence with other formulations which are not detailed in this section. The semi-coupling formulation, mentioned Section~\ref{sec:uot-dynamic}, is equivalent to the static formulation.

\myparagraph{Metric properties and extra constraints. } Conic and dynamic formulations are powerful to derive metric properties. 
Dynamic formulations always define a geodesic distance on $\Mmp(\X)$. It is interesting from a modeling point of view, because of its interpretation as a PDE. One can also incorporate physical priors into the model such as incompressibility (see e.g.~\cite{maury2011handling}). Such priors can not be expressed with the static formulation for arbitrary  $(\C,\D_\phi)$. However, the time variable remains computationally intensive to deal with, thus it is desirable to remove it so as to tackle large scale datasets.
The static formulation is hard to study theoretically.
For instance, as detailed in Section~\ref{sec:uot-conic}, when possible, one should consider its associated conic formulation to prove its metric properties on $\Mmp(\X)$.

\myparagraph{Computational complexity. } 
Conic formulations are not computationally friendly because of the extra radial variables $(r,s)\in\RR_+^2$ (see Equation~\eqref{eq:def-cone-pb}).
In sharp contrast, static and dynamic formulations are interesting because they do not need these radial variables. The dynamic formulation replaces them by involving a time variable $t$ (see Equation~\eqref{EqDefDynamic}). 
The static formulation offers the advantage of removing both time and radial variables, thus scaling computationally to larger applications. This is the main reason why this formulation is frequently used in applications, see Section~\ref{sec:entropy} for more details. 
The semi-coupling formulation, mentioned Section~\ref{sec:uot-dynamic}, can be useful in the WFR/KL setting to derive an algorithm sharing similarities with the Sinkhorn algorithm (detailed in the next Section~\ref{sec:entropy})~\cite{bauer2021square}.
We also refer to~\cite{chapel2021unbalanced} for another algorithm solving the static formulation with quadratic marginal penalties.
Similar to~\cite{bauer2021square}, it shares similarities with the Sinkhorn algorithm but involves no entropic regularization.

\myparagraph{Special cases of TV and $W_1$.}
The TV setting is equivalent to variants of Partial OT, that either relaxes the marginal constraint~\cite{piccoli2014generalized, figalli2010optimal} or adds a boundary/extra sample whose role is to add or remove mass from the system~\cite{figalli2010new, gramfort2015fast}. 

In this review, we did not detail a last equivalence of the static formulation when the cost is a metric $\C=d_\X$, which corresponds to $W_1$ OT in the balanced case. In that case one can reformulate it as an integral probability metric~\cite{muller1997integral} over the set of Lipschitz functions. It is also equivalent to a specific dynamic formulation called Beckmann. We refer to~\cite{santambrogio2015optimal} for more details on the balanced OT setting. For the unbalanced extension of the Lipschitz formulation, see~\cite{hanin1992kantorovich, hanin1999extension,schmitzer2019framework}, and we refer to~\cite{schmitzer2019framework} for the generalized Beckmann problem.
}

\section{Entropic Regularization}
\label{sec:entropy}

For most applications in data sciences, it makes sense to trade precision on the resolution of OT with speed and versatility. This can be achieved by solving a regularized OT problem, where some penalty makes the problem strictly convex and simpler to solve. These regularizations also bring stability to the solution with respect to perturbations of the input measures. This is useful to ease downstream tasks (such as when training neural networks using backpropagation) and is also a way to mitigate the curse of dimensionality, as we explain in Section~\ref{sec:sample-comp}. 
Popular regularizations of the plan include squared Euclidean norm~\cite{blondel2018smooth}, negative logarithm for interior point methods~\cite{den2012interior} and the Shannon entropy~\cite{cuturi2013lightspeed}, on which we focus now. 
This entropic regularization of OT can be traced back to the Schr\"odinger problem~\cite{schroedinger1931} as a statistical model for gas. We refer to~\cite{leonard2012schrodinger, leonard2013survey} for a survey on this problem.
It appeared in different fields with various modeling or computational motivations, see~\cite[Remark 4.4]{peyre2019computational} for a historical perspective.
The interest in the ML community was impulsed by~\cite{cuturi2013lightspeed} who emphasized its fast computations on GPUs (which are the computing devices used to train neural networks) and its smoothness amenable for back-propagation when using regularized OT as a training loss.

\subsection{Sinkhorn algorithm}
\label{sec:sinkhorn}

\myparagraph{Regularized UOT formulation.}
We detail this regularization in the context of unbalanced OT.

\begin{definition}\label{def:uot-entropic}
    Given positive measures $(\al,\be)\in\Mmp(\X)^2$, the entropic regularization of UOT reads
\begin{align}\label{eq:def-uot-primal}
\begin{aligned}
&\qquad\quad\OTb(\al,\be) \triangleq \inf_{\pi\in\Mmp(\X^2)} \Pp_\epsilon(\pi),\\
\Pp_\epsilon(\pi) &\triangleq \int\C\dd\pi + \D_{\phi_1}(\pi_1|\al) + \D_{\phi_2}(\pi_2|\be) + \epsilon\KL(\pi|\al\otimes\be),
\end{aligned}
\end{align}
where $\D_\phi$ is a $\phi$-divergence~\cite{csiszar1967information} (see Section~\ref{sec:phidiv}).
In what follows, $\OTb$ denotes both balanced and unbalanced OT, since balanced OT is a specific instance of the above definition.
\end{definition}

Problem introduced in Definition \ref{def:uot-entropic} is thus a Kantorovich problem with an extra entropic penalty $\KL(\pi|\al\otimes\be)$. This entropy is computed with respect to the input marginals to ease the discussion, but other reference measures can be used (such as in~\cite{cuturi2013lightspeed}).
As $\epsilon$ increases, the approximation of the original OT problem degrades, but as detailed in Section~\ref{sec:sinkhorn}, the computational scheme converges faster, and it is more stable in high dimension.
To derive a simple computational scheme, we detail below the dual formulation, which is obtained through Fenchel-Rockafellar duality theorem~\cite{rockafellar1967duality}, and corresponds to the optimization of a pair of continuous functions $(f,g)\in\Cc(\X)^2$.

\begin{proposition}
One has
\begin{align}\label{eq:def-uot-dual}
\begin{aligned}
\OTb(\al,\be) &\triangleq \sup_{(\f,\g)\in\Cc(\X)} \Dd_\epsilon(\f,\g),\\
\Dd_\epsilon(\f,\g) \triangleq &-\int_\X\phi_1^*(-\f(x))\dd\al(x) -\int_\X\phi_2^*(-\g(y))\dd\be(y)\\
&- \epsilon\int_{\X^2}\big(e^{(\f(x)+\g(y) - \C(x,y)) / \epsilon} \,-\, 1\big)\dd\al(x)\dd\be(y),
\end{aligned}
\end{align}
where $(\phi^*_1,\phi^*_2)$ are the Legendre transforms of $(\phi_1,\phi_2)$, i.e. they read $\phi_i^*(q)\triangleq \sup_{p\geq 0}pq - \phi_i(p)$.
\end{proposition}

When $\epsilon\rightarrow 0$ the term $- \epsilon\dotp{\al\otimes\be}{\big(e^{(\f\oplus\g - \C / \epsilon} - 1\big)}$ converges to the constraint $\f\oplus\g\leq\C$ which is classical in OT duality~\cite[Section 4]{liero2015optimal}.
In this case, the dual problem for $\epsilon=0$ thus reads
\begin{align*}
\UOT(\al,\be) &= \sup_{\f(x)+\g(y)\leq\C(x,y)}
-\int_\X\phi_1^*(-\f(x))\dd\al(x) -\int_\X\phi_2^*(-\g(y))\dd\be(y).
\end{align*}
When $\epsilon>0$, an important optimality property is that once optimal potentials $(\f^\star,\g^\star)$ are computed, the optimal primal plan $\pi^\star$ is retrieved as
\begin{align}\label{eq-implicit-plan}
	\frac{\dd\pi^\star}{\dd\al\dd\be}(x,y) = \exp\big(\tfrac{\f^\star(x) + \g^\star(y) - \C(x,y)}{\epsilon}\big).
\end{align}

\myparagraph{Unbalanced Sinkhorn operators.}
The simplest algorithm to solve this regularized UOT problem applies an alternate coordinate ascent on the dual problem, alternately maximizing on $f$ and on $g$. 
We refer to this class of methods as Sinkhorn's algorithm, which was historically derived in the balanced OT case~\cite{sinkhorn1964relationship}.
Let us underline that this algorithm does not converge when $\epsilon=0$ (because of the constraint in the dual), however it does converge for the regularized case $\epsilon>0$. 
This algorithm appeared in a variety of fields, 
and we refer to~\cite[Remark 4.4]{peyre2017computational} for a historical account. 

To write the iteration of this algorithm in a compact way, and also ease the mathematical analysis, we write the dual optimality conditions of $\OTb$ using two operators called the \emph{Softmin} operator and the \emph{anisotropic proximity} operator~\cite{sejourne2019sinkhorn}.

\begin{definition}\label{def:softmin}
For any $\al \in \Mmpp(\X)$ and $\epsilon >0$, the Softmin operator $\Smin{\al}{\epsilon}$ defined for any $\f\in\Cc(\X)$ as
\begin{align}
	\Smin{\al}{\epsilon}(\f) \eqdef -  \epsilon \log\bigg(\int_X e^{-\f(x)/\epsilon}\dd\al(x)\bigg)\in\RR.
\end{align}
\end{definition}

We define two maps $\Ss_\al:\Cc(\X)\rightarrow\Cc(\X)$ and $\Ss_\be:\Cc(\X)\rightarrow\Cc(\X)$ derived from the Softmin, which play a key role in the definition of Sinkhorn algorithm. 
For any $(\f,\g)\in\Cc(\X)^2$ and $(x,y)\in\X^2$, the outputs $(\Ss_\al(\f), \Ss_\be(\g))$ read
\begin{gather}
\begin{aligned}\label{eq:defn-softmin-func}
	\Ss_\al(\f)(y)&\eqdef \Smin{\al}{\epsilon}(\C(\cdot,y) - \f),\\
	\Ss_\be(\g)(x)&\eqdef \Smin{\be}{\epsilon}(\C(x, \cdot) - \g).
\end{aligned}
\end{gather}
The \emph{balanced} Sinkhorn algorithm (which solves balanced OT) simply performs the updates $\f\leftarrow\Ss_\be(\g)$ and $\g\leftarrow\Ss_\al(\f)$.
To derive the unbalanced updates we need another operator defined below.
\begin{definition}\label{def:aprox}
    The anisotropic proximity operator involves a convex function $h : \RR \rightarrow \RR$ and $\epsilon > 0$.
It is defined for any $p\in\RR$ as
\begin{align}\label{eq-def-aprox}
\aprox{h}(p) \eqdef \arg\min_{q\in\RR} \epsilon \exp(\tfrac{p - q}{\epsilon}) + h(q)\in \text{dom}(h)\cup\{+\infty\}.
\end{align}
In what follows, we take $h=\phi^*$.
\end{definition}

As detailed in \cite{combettes2013moreau}, a generalized Moreau decomposition connects $\aprox{\phi^*}$ with a $\KL$ (Bregman) proximity operator that reads
\begin{gather}\label{eq-proxdiv}
    \begin{aligned}
    \aprox{\phi^*}(p) &=  p -  \epsilon\log \proxdiv{\phi}(p),\\
    \quad\text{where}\quad \proxdiv{\phi}(p)&\eqdef \arg\inf_{q\in\RR_+} \phi(q) + \KL(q, \exp(\tfrac{p}{\epsilon}))\,.
    \end{aligned}
\end{gather}

The above $\proxdiv{\phi}$ operator is used in~\cite{chizat2016scaling} to define the Sinkhorn algorithm.
Expressing the algorithm's iteration using $\aprox{\phi^*}$ is interesting, because this operator is  non-expansive, which is important to show the convergence of the method~\cite{sejourne2019sinkhorn}. 

\myparagraph{Sinkhorn iterates.}
Equipped with these definitions, we can conveniently write the optimality conditions for the dual UOT problem as two fixed point equations, which in turn are used alternatingly in the Sinkhorn's algorithm.
\begin{proposition}[Optimality conditions for the dual problem]\label{prop-optimality-prox}
The first order optimality condition of $\Dd_\epsilon$~\eqref{eq:def-uot-dual} reads
\begin{gather}\label{eq-optim-cond}
	\begin{aligned}
	\f(x) &= -\aprox{\phi^*}\big(-\Ss_\be(\g)(x)\big),\quad \al-\text{a.e.} \\
	\g(y) &= -\aprox{\phi^*}\big(-\Ss_\al(\f)(y)\big), \quad \be-\text{a.e.},
	\end{aligned}
\end{gather}
where $\aprox{\phi^*}$ is applied pointwise, and $(\Ss_\al,\Ss_\be)$ are defined Equation~\eqref{eq:defn-softmin-func}.
For the sake of brevity, we define operators $(\Aa\Ss_\al,\Aa\Ss_\be)$ outputting functions in $\Cc(\X)$ to write Equations~\eqref{eq-optim-cond} as $\f =  \Aa\Ss_\be(\g)$ and $\g=\Aa\Ss_\al(\f)$.
\end{proposition}
\begin{definition}[Sinkhorn algorithm]
\label{def:sinkhorn}
Starting from some $\g_0 \in \Cc(\X)$, the iterations of Sinkhorn read
\begin{gather}
\begin{aligned}\label{eq-sinkhorn-iter-1}
\f_{t+1} &\eqdef \arg\max_{\f}\Dd_\epsilon(\f,\g_t)=\Aa\Ss_\be(\g_t),\\
\g_{t+1} &\eqdef  \arg\max_{\g}\Dd_\epsilon(\f_{t+1},\g)=\Aa\Ss_\al(\f_{t+1}),
\end{aligned}
\end{gather}
where $(\Aa\Ss_\al,\Aa\Ss_\be)$ are defined in Proposition~\ref{prop-optimality-prox}.
\end{definition}

\myparagraph{Discrete setting.}
To help the reader motivated by a computational implementation, we detail the formulas when the inputs are discrete measures.
We write discrete measures as $\al = \sum_{i=1}^N \al_i \de_{x_i}$ and $\be=\sum_{j=1}^M \be_j \de_{y_j}$, where $(\al_i)_i,(\be_j)_j \in\RR_+^N\times\RR_+^M$ are vectors of non-negative masses and $(x_i)_{i},(y_j)_{j} \in \X^N\times\X^M$ are two sets of points. 
Potentials $(\f_i) = (\f(x_i))$ and $(\g_j) = (\g(y_j))$ become two vectors of $\RR^N$ and $\RR^M$.
The cost $\C_{ij}=\C(x_i,y_j)$ and the transport plan $\pi_{ij}=\pi(x_i,y_j)$ become matrices of $\RR^{N\times M}$.
The latter can be computed with Equation~\eqref{eq-implicit-plan} which becomes $\pi_{ij} =  \exp(\tfrac{1}{\epsilon}[\f_i + \g_j - \C_{ij}])\al_i \be_j$.
The dual functional reads
\begin{align*}
    \Dd_\epsilon(\f,\g) = &-\sum_{i=1}^N \phi_1^*(-\f_i)\al_i -\sum_{j=1}^M \phi_2^*(-\g_j)\be_j\\
&- \epsilon\sum_{i,j}\big(e^{(\f_i+\g_j - \C_{ij}) / \epsilon} \,-\, 1\big)\al_i\be_j.
\end{align*}
To reformulate Sinkhorn updates of Definition~\ref{def:sinkhorn} in a discrete setting, recall the operator $\aprox{\phi^*}$ is applied pointwise (see next Section~\ref{sec:exmpl-phidiv} for settings with closed forms).
The Softmin is reformulated as
\begin{align*}
    \Ss_\be(\f) = -\epsilon\log\bigg( \sum_{j=1}^{M} e^{\log\be_j + (g_j - \C_{ij})/\epsilon} \bigg).
\end{align*}
The Softmin is from a computational perspective a log-sum-exp reduction, which can be stabilized numerically for any vector $(u_k)\in\RR^N$ as
\begin{align}\label{eq:lse-trick}
\log\bigg( \sum_{k=1}^N e^{u_k} \bigg) = \max_p u_p + \log\bigg( \sum_{k=1}^\N e^{u_k - \max_p u_p} \bigg).
\end{align}
For any $k$, one has $u_k -\max_p u_p\leq 0$, thus any term $e^{u_k - \max_p u_p}\leq 1$, which prevents numerical overflows of the exponential function.

\rev{The pseudo-code to implement Sinkhorn algorithm is detailed in Algorithm~\ref{alg:sinkhorn}. The log-sum-exp reduction is abbreviated $\LSE$. Closed forms of $\aprox{\phi^*}$ are detailed in Section~\ref{sec:exmpl-phidiv}.}

\begin{algorithm} 
	\caption{~~~~\, {Sinkhorn($(\al_i)_i$,$(\be_j)_j$, $(\C_{ij})_{ij}$)} \label{alg:sinkhorn}}
	\textbf{Input~~~~\,:}~~  cost matrix $(\C_{ij})=(\C(x_i,y_j))\in\RR^{\N\times\M}$, with source $\alpha = \sum_{i=1}^\N \al_i\delta_{x_i}$ and target~ $\beta = \sum_{j=1}^\M \be_j\delta_{y_j}$, where $(\al_i)\in\RR^N$, $(\be_j)\in\RR^M$, $(x_i)_i\in\RR^{\N\times D}$, $(y_j)_j\in\RR^{\M\times D}$\\
	\textbf{Parameters~:}~~ entropy $\phi$, regularization $\epsilon > 0$ \\
	\textbf{Output~~\,:}~~ vectors $(\f_i)_i$ and $(\g_j)_j$, equal to the optimal potentials of $\UOT(\al,\be)$ (see Equation~\eqref{eq:def-uot-dual})
	\begin{algorithmic}[1]
		\STATE $\f_i\gets \text{zeros}(\M)$~~;~~$\g_j\gets \text{zeros}(\N)$ \COMMENT{Vectors of size $\M$ and $\N$}
		\vspace{.1cm}\WHILE{updates $>$ tol}\vspace{.1cm}
		\STATE $\g_j \gets - \,\epsilon \LSE_{i=1}^\N \big[ \log(\al_i) + (\f_i - \C_{ij})\,/\,\epsilon\,\big]$
		\STATE $\g_j \gets -\aprox{\phi^*}(-\g_j)$
		\label{alg:sinkhorn:line_a}
		\STATE $\f_i \gets - \,\epsilon \LSE_{j=1}^\M \big[ \log(\be_j) + (\g_j - \C_{ij}s)\,/\,\epsilon\,\big]$
		\STATE $\f_i \gets -\aprox{\phi^*}(-\f_i)$
		\label{alg:sinkhorn:line_b}
		\ENDWHILE
		\STATE \RETURN{~~$(\f_i)_i,~~(\g_j)_j$}
	\end{algorithmic}
\end{algorithm}

\myparagraph{A faster (but unstable) matrix-vector product formulation.}
The Sinkhorn updates can be formulated as matrix-vector products for computations between discrete measures, hence the speed-up on computations with GPUs (see~\cite{chizat2016scaling}).
Setting $K \triangleq (e^{-\C(x_i,y_i) /\epsilon})_{ij}$, $u \triangleq (e^{\f(x_i) / \epsilon})_i$ and $v \triangleq (e^{\g(y_j) / \epsilon})_j$.
It reads in the case of balanced OT
\begin{align*}
	u\leftarrow \frac{\proxdiv{\phi}(-\Ss_\be(\g))}{K(v\odot\be)},\qquad v\leftarrow \frac{\proxdiv{\phi}(-\Ss_\al(\f))}{K^\top(u\odot\al)},
\end{align*}
where $\odot$ represents the coordinate-wise product between vectors, and such that using Equation~\eqref{eq-proxdiv} one has
\begin{align*}
	\proxdiv{\phi}(-\Ss_\be(\g))&=\arg\inf_{q\in\RR_+} \phi(q) + \KL(q, [K(v\odot\be)]),\\
	\proxdiv{\phi}(-\Ss_\al(\f))&=\arg\inf_{q\in\RR_+} \phi(q) + \KL(q, [K^\top(u\odot\al)]),
\end{align*}
in other terms the numerators are rephrased to take as inputs $(K,u,v)$.
Furthermore the optimal plan reads $\pi=\diag(u\odot\al)K\diag(v\odot\be)$.
This change of variable yields a significant acceleration on GPUs, because the terms $K(v\odot\be)$ and $K^\top(u\odot\al)$ are matrix-vector products efficiently parallelized on such devices.
However, they are numerically sensitive.
For small values of $\epsilon$, the actual numerical computation of the kernel $K$ might contain too many zeros and underflow, thus breaking the implementation.
Those updates are frequently used for large $\epsilon$, due to the significant computational benefits.
Nevertheless, we consider the updates of Definition~\ref{def:sinkhorn} (called \hyphen{log-stable}) which can be implemented in a numerically stable way for any value of $\epsilon$ (see previous paragraph).
Those updates are more expensive since the log-sum-exp trick (Equation~\eqref{eq:lse-trick}) requires the computation of a maximum, but are more robust for small values of $\epsilon$.

\myparagraph{Extension - Spatially varying $\phi$-divergences.}
For some applications, it is desirable to vary across the domain the way mass variation are penalized.
Recall Csiszàr divergences integrate pointwise penalties on $\tfrac{\dd\al}{\dd\be}$.
It is thus possible to generalize $\D_\phi$ as
\begin{align}\label{eq:csiszar-div-varying}
\D_\phi(\al|\be) \eqdef \int_\X \phi\Big(\frac{\dd\al}{\dd\be}(x),x\Big) \dd\be +  \int_\X \phi^\prime_\infty(x) \dd\al^\bot(x),
\end{align}
where $\phi(\cdot,x)$ is an entropy function for each location $x \in \X$ with associated recession value  $\phi^\prime_\infty(x)$.
A typical example of such divergence consists in using a spatially varying parameter $\rho(x)$, such that for e.g. $\KL$ penalties one takes $\phi(p,x)=\rho(x) ( p \log p -p +1)$.
It allows modulating the strength of the conservation of mass constraint over the spatial domain $\X$.
Such a situation appears e.g. in biology where the frequency of cell duplications $\rho(x)$ depends on the cell functionality $x$.

Some regularity is however required to avoid measurability issues and be able to apply Legendre duality.
It is well-defined when the function (defined on $\mathbb{R}_+^2 \times \X$) $\Phi : (r,s,x) \mapsto \phi(r/s,x) s$
(properly extended when $s=0$ using $\phi^\prime_\infty(x)$) is a so-called normal-integrant, see~\cite[chap.14]{rockafellar2009variational}.
For instance, this is ensured if $\Phi$ is lower-semi-continuous.

Sinkhorn algorithm extends to this case, using a spatially varying map $\aprox{\phi^*(\cdot,x)}$ at each $x\in\X$.
Note that when $\phi(p,x)=\rho(x)\phi(p)$ one has $\phi^*(p,x) = \rho(x)\phi^*(p / \rho(x))$.
The full Sinkhorn update outputs the function $$x\mapsto -\aprox{\phi^*(\cdot,x)}(-\Ss_\al(\f)(x)),$$ and for $\KL$ penalties $\phi(p,x)=\rho(x) ( p \log p -p +1)$, it reads $\aprox{\phi^*(\cdot,x)}(q) = (1 + \tfrac{\epsilon}{\rho(x)})^{-1}q$.

\subsection{Popular settings, closed forms and numerical illustrations}
\label{sec:exmpl-phidiv}
We detail in this section examples of $\phi$-divergences popularly used in the literature. For each setting we provide the entropy $\phi$, its Legendre transform $\phi^*$ and its associated $\aprox{\phi^*}$.

\myparagraph{Balanced OT} ($\D_\phi=\iota_{(=)}$) corresponds to using $\phi = \iota_{\{1\}}$, the convex indicator function ($\phi(1)=0$ and $+\infty$ otherwise) which encodes the marginal constraints, i.e. $\frac{\dd\pi_1}{\dd\al} = 1$ and $\frac{\dd\pi_2}{\dd\be} = 1$. In this case we get $\phi^*(q) = q$ and $\aprox{\phi^*}(p) = p$.

\myparagraph{Kullback-Leibler} ($\D_\phi=\rho\KL$) corresponds to $\phi(p) = \rho ( p \log p -p +1)$, to $\phi^*(q) = \rho(e^{q / \rho} -1)$ and $\aprox{\phi^*}(p) = (1+\tfrac{\epsilon}{\rho})^{-1} p$.
As discussed in \cite{liero2015optimal}, when $\epsilon=0$ and $d_\X$ is a distance, unbalanced OT defines the Hellinger-Kantorovich  and the Gaussian-Hellinger distances on $\Mmp(\X)$ (respectively for $\C(x,y)=-2\log\cos(\min(d_\X(x,y),\tfrac{\pi}{2}))$ and $\C(x,y)=d_\X(x,y)^2$).

\myparagraph{Total Variation} ($\D_\phi=\rho TV$) corresponds to $\phi(p) = \rho|p-1|$ and for $q\leq \rho$, $\phi^*(q) = \max(-\rho, q)$ with $\text{dom}(\phi^*) = (-\infty, \rho]$. The anisotropic operator reads
\begin{align*}
\aprox{\phi^*}(p) =
\text{Clamp}_{[-\rho, +\rho]}(p)=
\begin{cases}
-\rho & \quad\text{if } p < -\rho \\
p & \quad\text{if } p \in [-\rho, \rho]\\
\rho & \quad\text{if } p > \rho.
\end{cases}
\end{align*}
In this case, unbalanced OT (i.e. when $\epsilon=0$) is a Lagrangian version of partial optimal transport~\cite{figalli2010optimal}, where only some fraction of the total mass is transported. 
When $\C$ is a distance, it is also equivalent to the flat norm (the dual norm of bounded Lipschitz functions)~\cite{hanin1999extension,hanin1992kantorovich,schmitzer2019framework}.

\myparagraph{Range} ($\D_\phi=RG_{[a,b]}$) is defined for $0\leq a \leq 1 \leq b$ with $\phi = \iota_{[a,b]}$ and $\phi^*(q) = \max(a q, b q)$. The proximal operator is
\begin{align*}
\aprox{\phi^*}(p) =
\text{Soft-Thresh}_{\epsilon \log a}^{\epsilon \log b}(p)=
\begin{cases}
p - \epsilon\log a & \quad\text{if } p - \epsilon\log a < 0,\\
p - \epsilon\log b & \quad\text{if } p - \epsilon\log b > 0,\\
0 & \quad\text{otherwise.}
\end{cases}
\end{align*}
Note that in this setting the problem can be infeasible, i.e. $\OTb(\al,\be)=+\infty$. We have $\OTb(\al,\be)<\infty$ if and only if $[m(\al)a,m(\al)b]\cap[m(\be)a,m(\be)b]\neq\emptyset$.

\myparagraph{Power entropies} divergences are parameterized by $s\in\RR\setminus\{0,1\}$ and $r \eqdef s / (s-1)$. When $s<1$ it corresponds to
\begin{align*}
\phi(p) &= \frac{\rho}{s(s-1)}\big(p^s -s(p-1) -1 \big),\\
\phi^*(q) &= \rho\frac{r-1}{r}\left[\big(1 + \frac{q}{\rho(r-1)}\big)^r -1\right].
\end{align*}
Special cases include \emph{Hellinger} with $s=1/2$, and \emph{Berg entropy} as the limit case $s=0$, defined by $\phi(p) = \rho(p - 1 - \log p)$ and $\phi^*(q) = - \rho\log( 1 - q / \rho)$ with $\text{dom}(\phi^*) = (-\infty, \rho)$. 
Kullback-Leibler is the limit $s=1$. 
We refer to~\cite{liero2015optimal} for more details, and to~\cite{de2019metric} for a proof that it defines a distance for a subset of power exponents ($s>1$).
Furthermore, when $r<1$, $\phi^*$ is strictly convex and the proximal operator satisfies
\begin{align*}
\aprox{\phi^*}(p) = \rho(1-r) - \epsilon (1-r) W\left(\frac{\rho}{\epsilon} \exp\big(\frac{-p + \rho(1-r)}{\epsilon(1-r)}\big)\right),
\end{align*}
where $W$ is the Lambert function, which satisfies for any $p\in\RR_+$ $W(p)e^{W(p)} = p$, see~\cite{corless1996lambertw}.
It is a non-expansive operator, and it is a contraction on compact sets.

\begin{figure*}[h]
	\centering
	\begin{tabular}{c@{}c@{}c}
		{\includegraphics[width=0.4\linewidth]{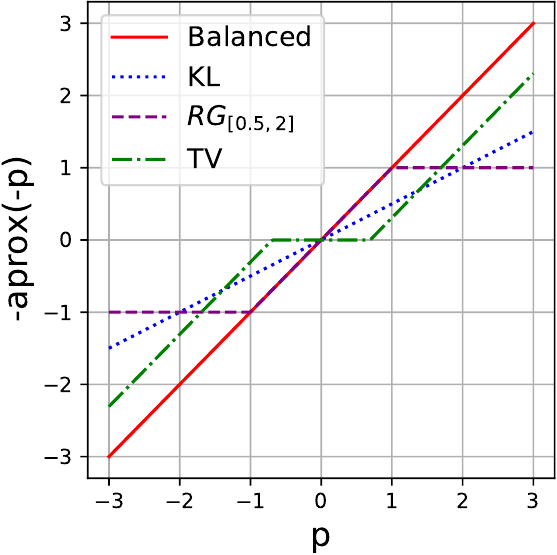}} & $\quad$ &
		{\includegraphics[width=0.4\linewidth]{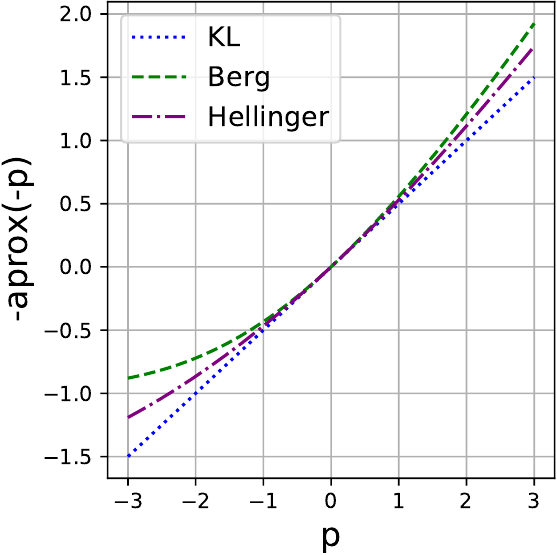}}
	\end{tabular}
	\caption{\textit{
	Display of the 1-Lipschitz operator $p \mapsto -\aprox{\phi^*}(-p)$ in six explicit settings, using $\epsilon=1$ and $\rho=1$.
		}
	}
	\label{fig-aprox}
	\vspace*{1.0em}
\end{figure*}

\begin{figure}[p]
	\centering
	
	\newcommand{\myfigE}[1]{\includegraphics[height=.14\linewidth]{figures/compare_entropy/comparison_entropy_#1}}

	\centering
	\begin{tabular}{c@{\hspace{1mm}}c@{\hspace{1mm}}c@{\hspace{1mm}}}%
		\myfigE{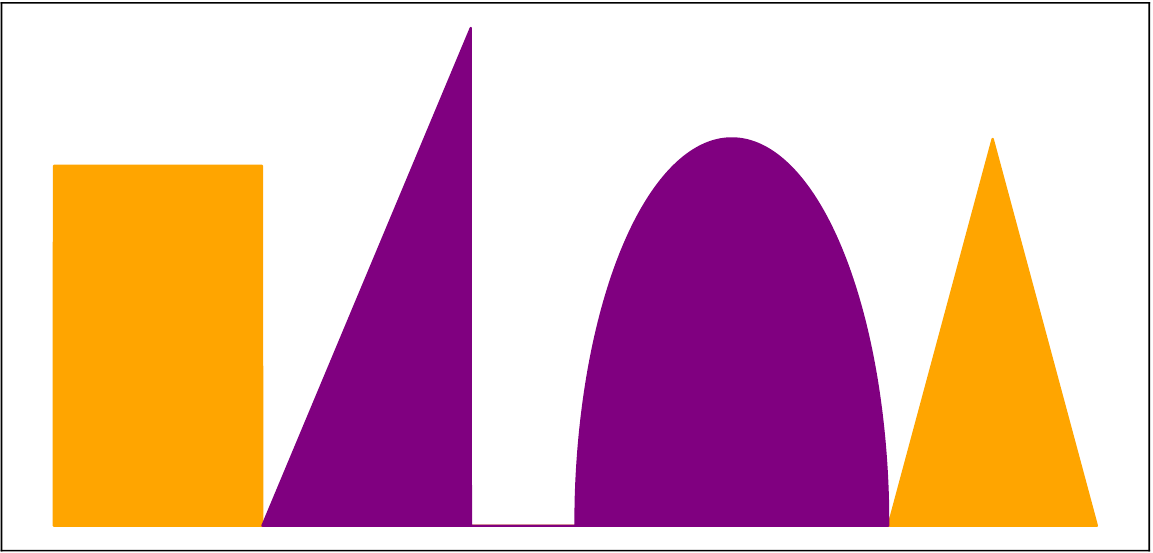} &
		\myfigE{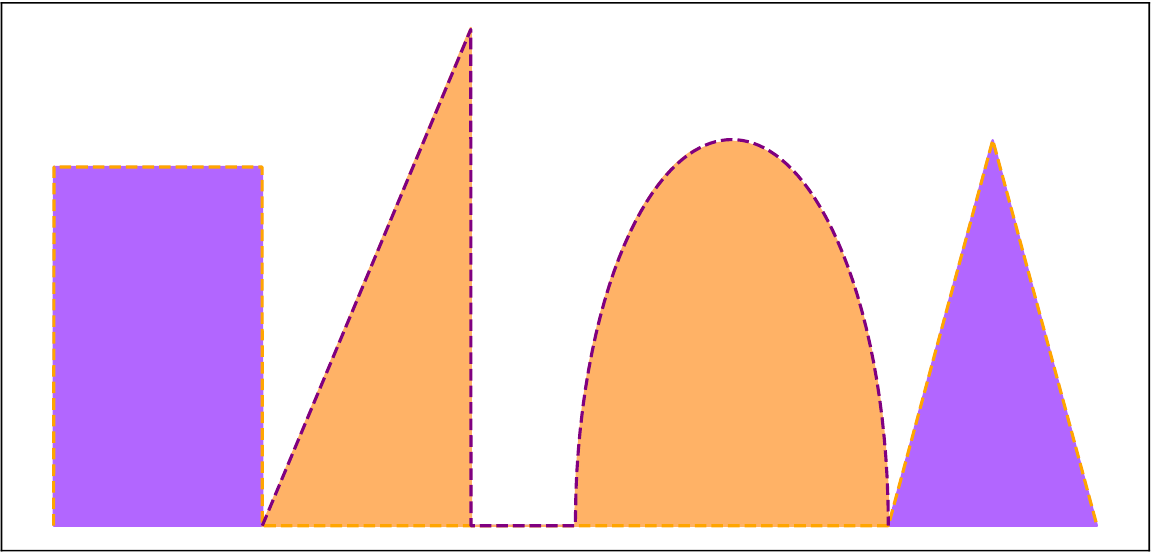} &
		\myfigE{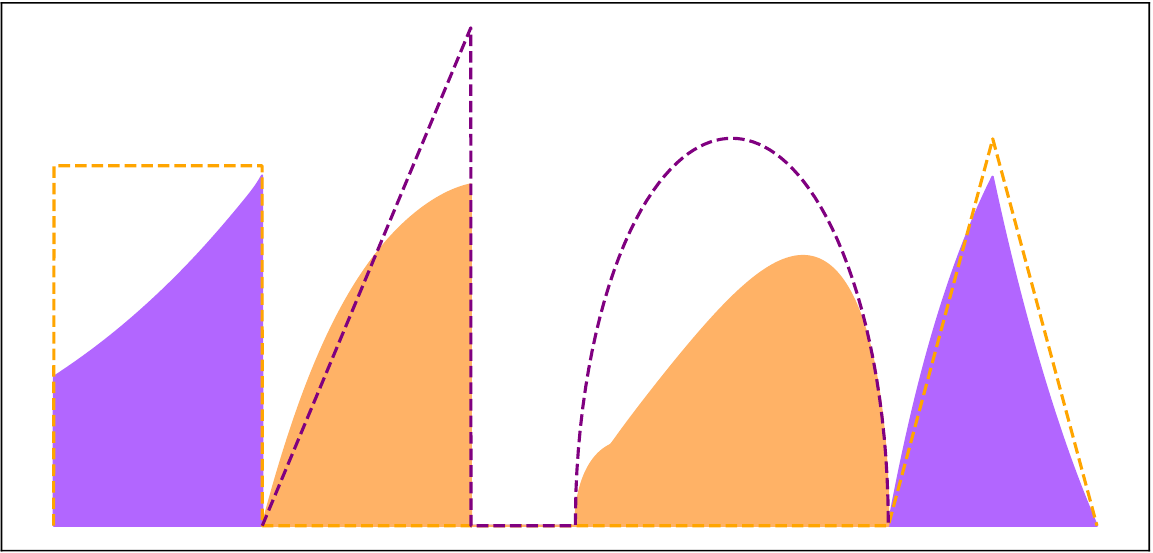} \\[-1mm]
		Inputs $(\al,\be)$ & $\iota_{(=)}$ & $10^{-1} * \KL$ \\[1mm]
		\myfigE{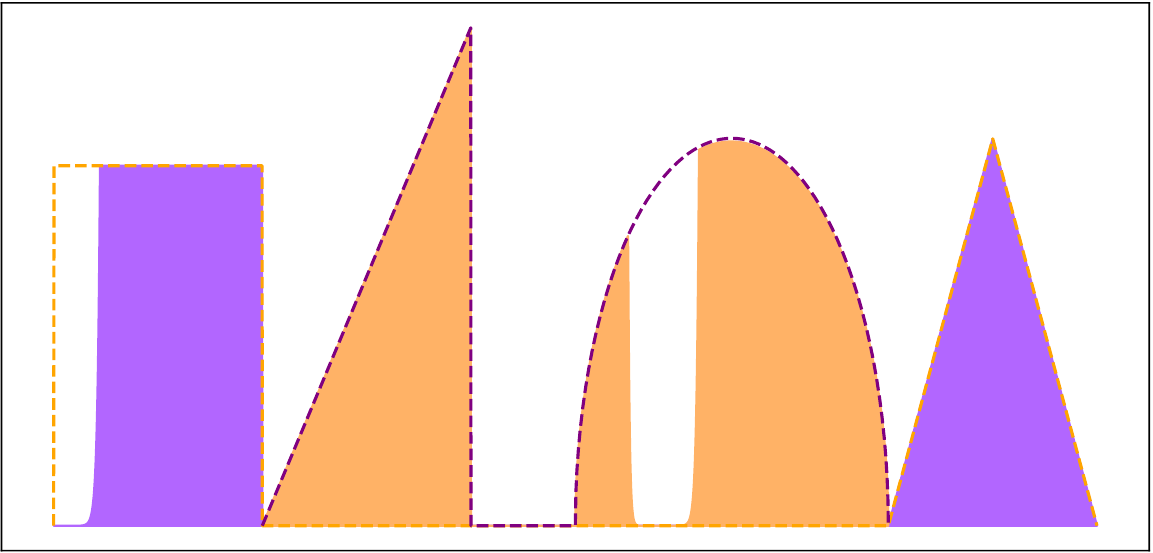} &
		\myfigE{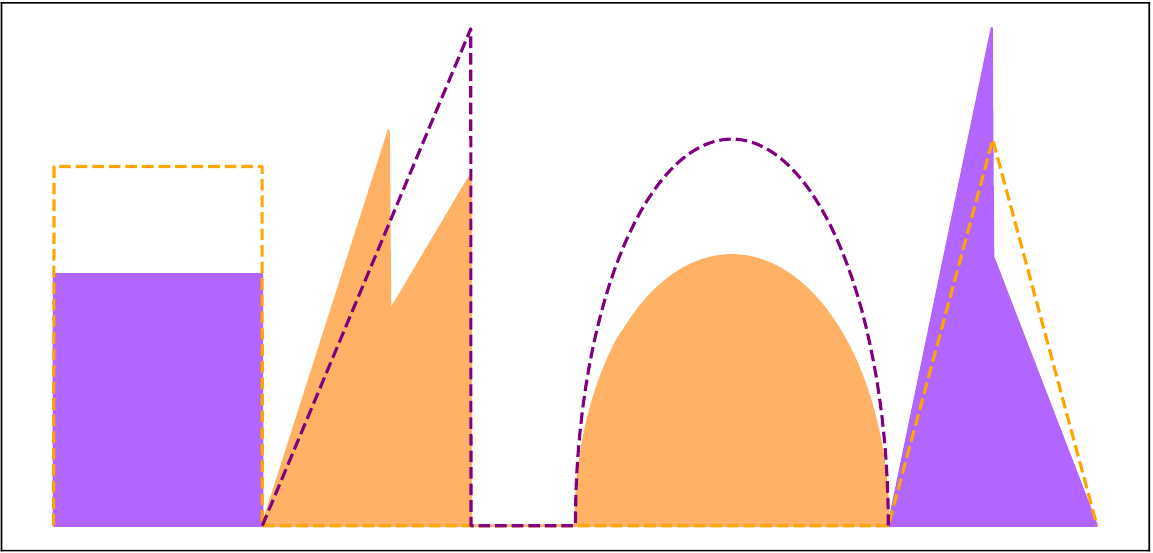} &
		\myfigE{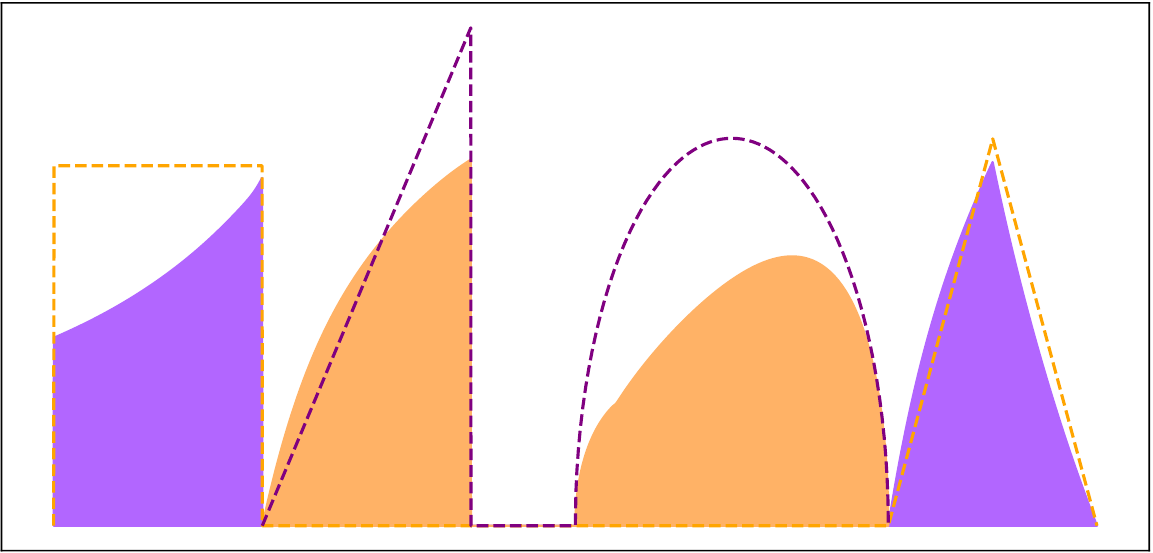} \\[-1mm]
		$10^{-1} * \TV$ & $\RG_{[0.7, 1.3]}$ & $10^{-1} * \text{Berg}$
	\end{tabular}
	\caption{\textit{
		Display of optimal marginals $(\textcolor{myo}{\pi_1},\textcolor{myp}{\pi_2})$ depending on $\phi$. 
		The inputs $(\textcolor{myp}{\al},\textcolor{myo}{\be})$ are 1D.
		Measures $(\textcolor{myp}{\al},\textcolor{myo}{\be})$ and $(\textcolor{myo}{\pi_1},\textcolor{myp}{\pi_2})$ are respectively plotted as dashed lines and filled colorings. 
		We use a regularization $\sqrt{\epsilon}=\sqrt{10^{-3}}$ on $[0,1]$.}
	}
	\label{fig-comp-ent}
	\vspace*{.7cm}
	
	\newcommand{\myfig}[1]{\includegraphics[width=.23\linewidth]{figures/compare_reach/comparison_#1}}
	\centering
	\begin{tabular}{@{}c@{\hspace{1mm}}c@{\hspace{1mm}}c@{\hspace{1mm}}c@{\hspace{1mm}}c@{}}
		\rotatebox{90}{\small $\D_\phi=\rho\KL$} & \myfig{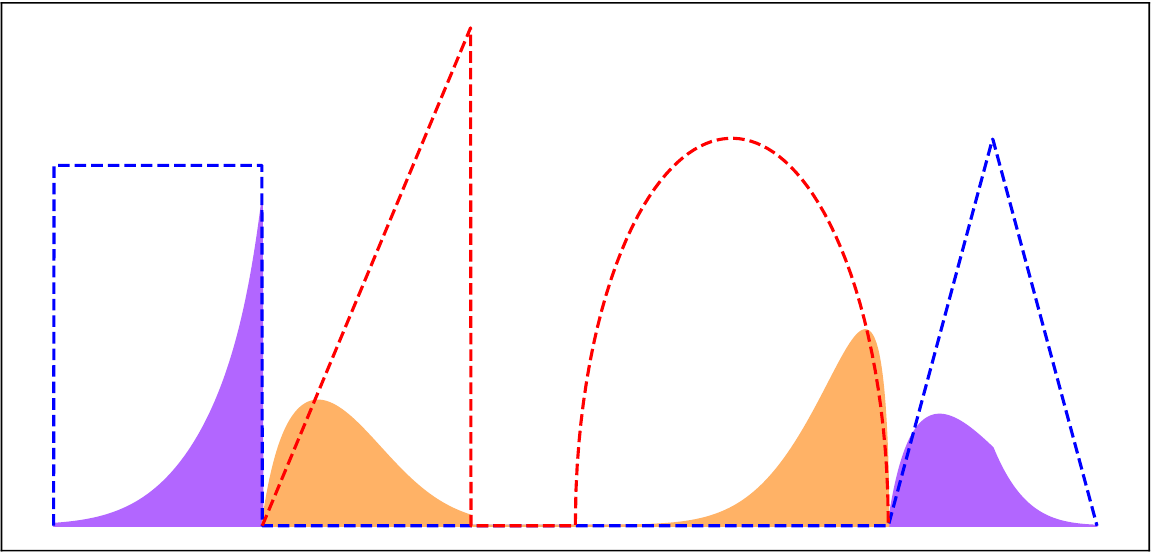} & \myfig{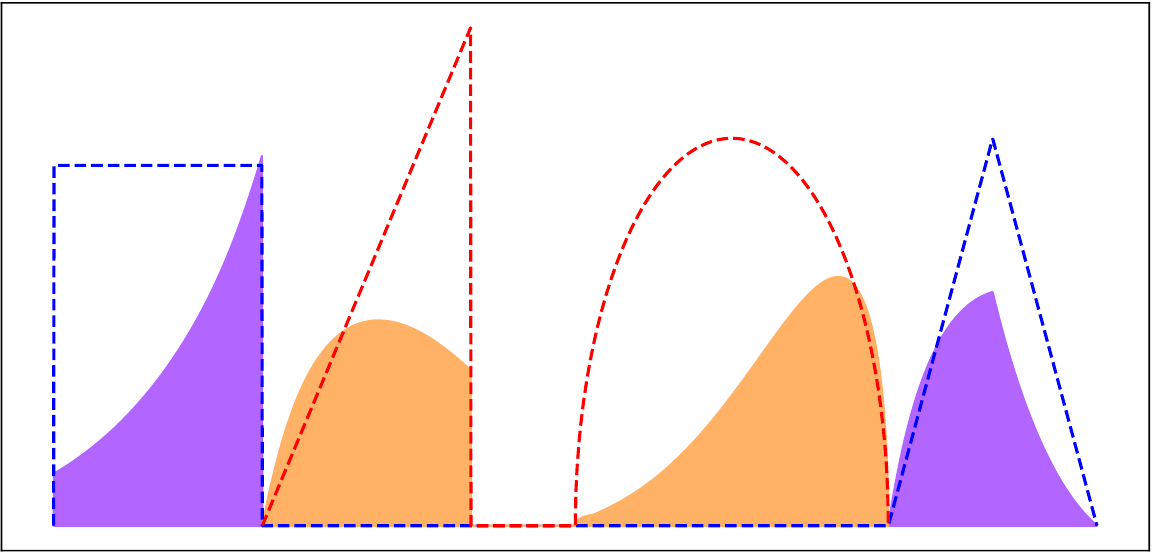} & \myfig{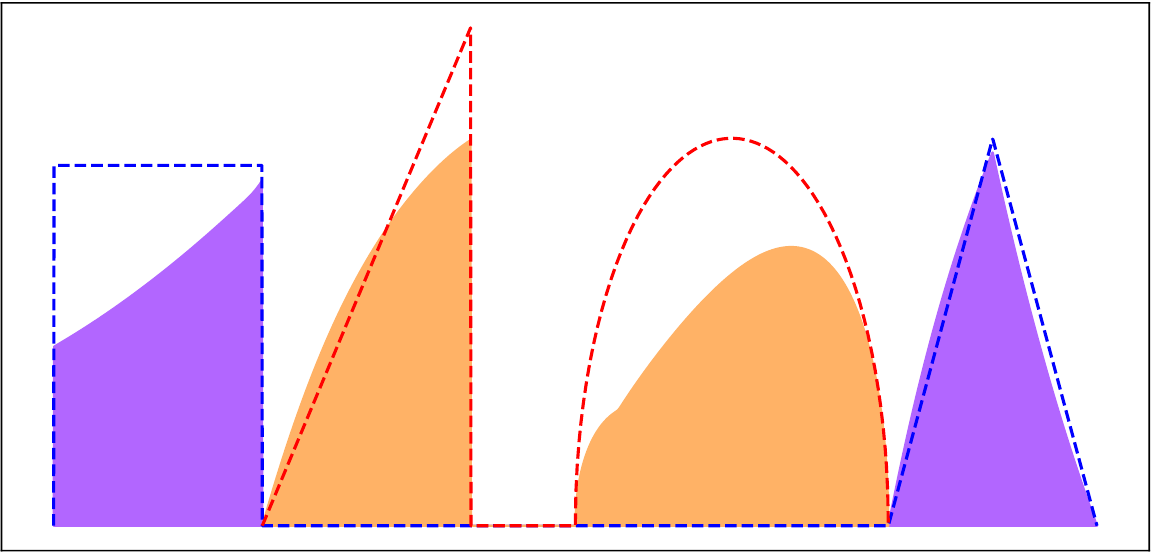} & \myfig{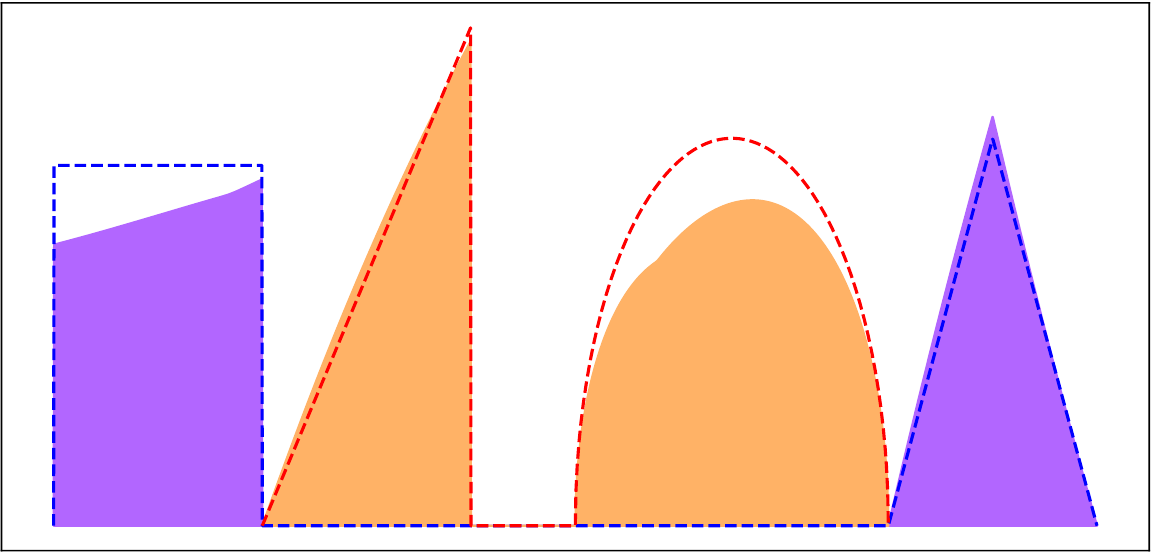} \\
		\rotatebox{90}{\small $\D_\phi=\rho\TV$\;\;} & \myfig{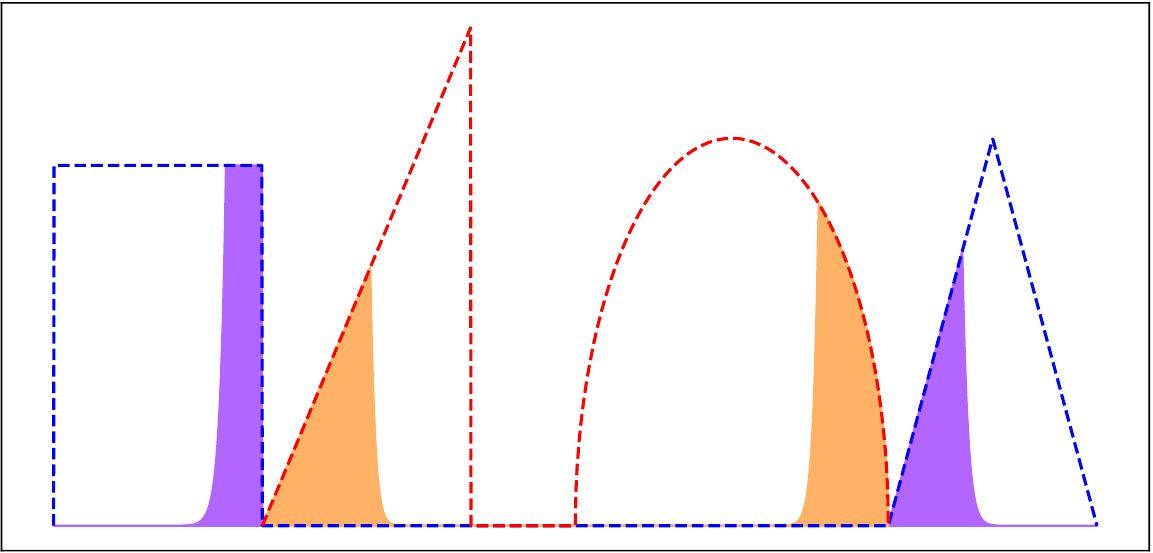} & \myfig{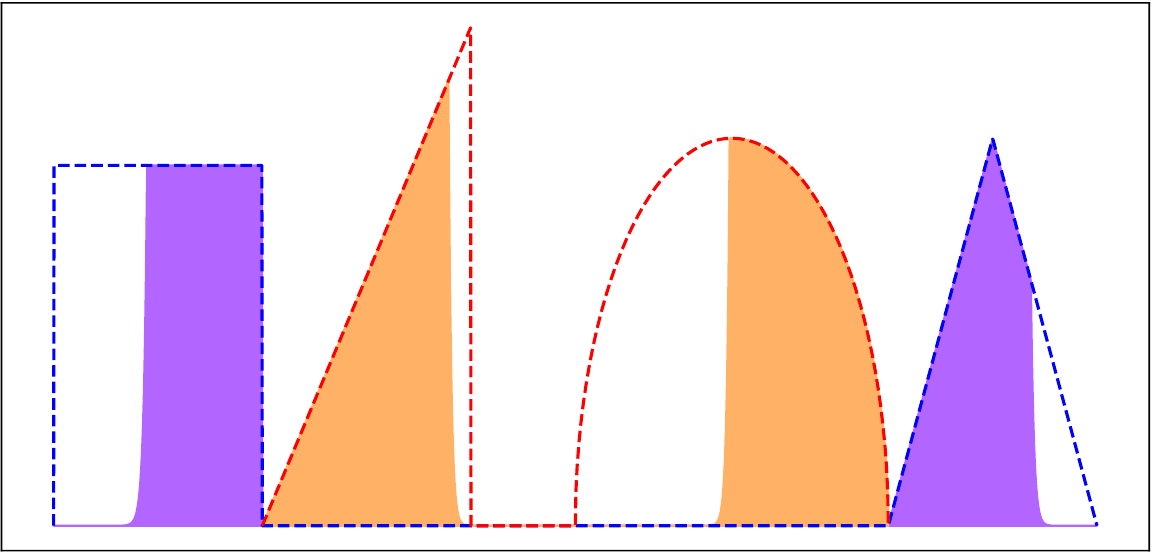} & \myfig{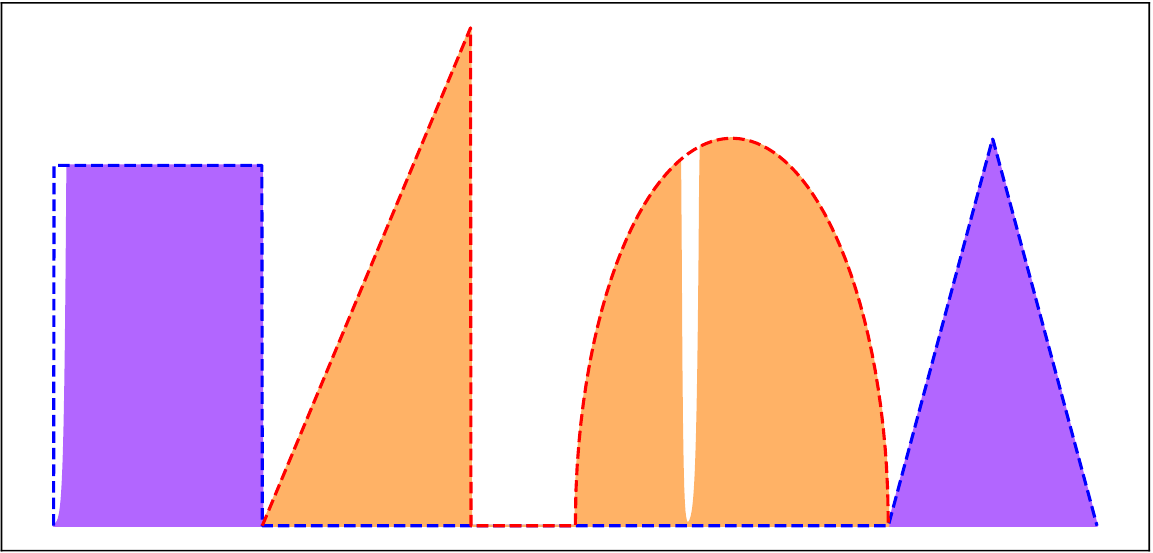} & \myfig{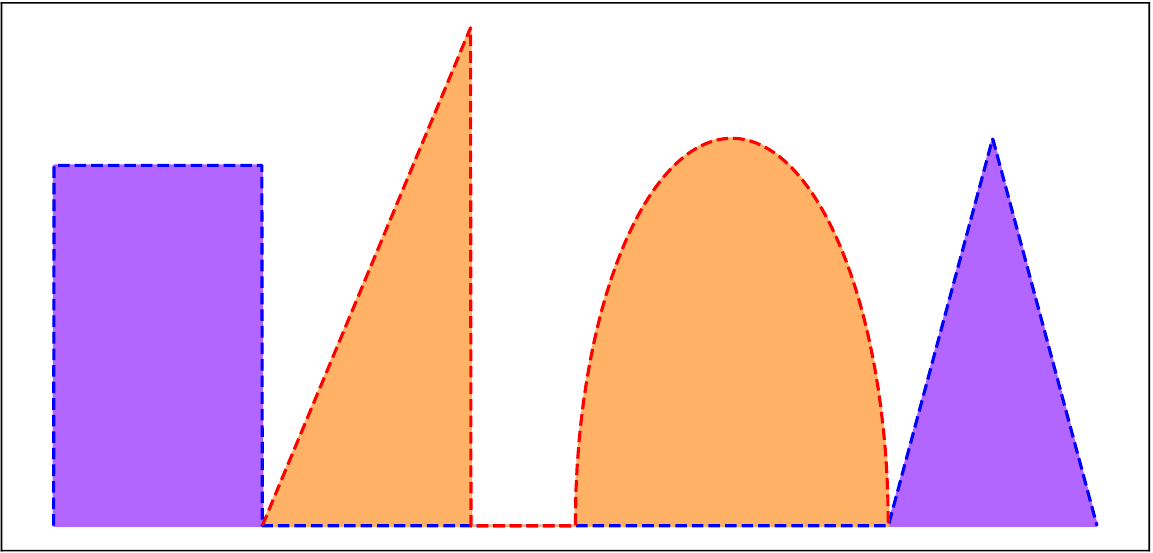} \\
		& $\rho=0.01$ & $\rho=0.03$ & $\rho=0.13$ & $\rho=0.5$
	\end{tabular}
	\caption{\textit{
		Display of marginals $(\textcolor{myo}{\pi_1},\textcolor{myp}{\pi_2})$ depending on parameter $\rho$.
		We use the same inputs $(\textcolor{myp}{\al},\textcolor{myp}{\be})$ from Figure~\ref{fig-comp-ent}. 
		First line corresponds to $\rho\KL$ and the second to $\rho\TV$.}
	}
	\label{fig-impact-reach}
\end{figure}

\myparagraph{Numerical illustration - Interpretation of Sinkhorn iterates.}
	We give an informal interpretation of Sinkhorn iterations for different divergences based on Proposition~\ref{prop-optimality-prox}, to illustrate the role of $\aprox{\phi^*}$.
	Optimality conditions have a compositional structure.
	Operators $(\Ss_\al,\Ss_\be)$ characterize optimal \emph{balanced} potentials as fixed points $\f=\Ss_\be(\g)$ and $\g=\Ss_\al(\f)$.
	The operator $\aprox{\phi^*}$ updates such fixed point by \emph{saturating} ($\D_\phi=\TV$) or \emph{dampening} ($\D_\phi=\KL$) dual potentials, see Figure~\ref{fig-aprox}.
	It indirectly impacts the plan via Equation~\eqref{eq-implicit-plan} by blocking or reducing transportation, see Figure~\ref{fig-impact-reach}.
	
Figure~\ref{fig-comp-ent} displays the impact of $\phi$ on the optimal plan $\pi$.
Here marginals $(\pi_1, \pi_2)$ are compared to the input marginals $(\al,\be)$.
\rev{Recall that $\pi_1, \pi_2$ represent the marginals of the transported mass (i.e. which is not teleported/created-deleted).}
Informally speaking, $\TV$ has \hyphen{sharp} marginals, i.e. it either transports s.t. $\pi_1(x)=\al(x)$ or destroys mass s.t. $\pi_1(x)=0$.
Marginals with $\KL$ are \hyphen{smooth} in the sense that it progressively transitions between transportation and destruction as $\C(x,y)$ increases.
Marginals for $\RG_{[a,b]}$ are less interpretable due to the box constraint, but we see that $\tfrac{\dd\pi_1}{\dd\al}\in\{a,b\}$.
We observe that the result of Berg entropy is similar to $\KL$. 

Figure~\ref{fig-impact-reach} shows when $\D_\phi=\rho\KL$ or $\D_\phi=\rho\TV$ the impact of the parameter $\rho$ on $(\pi_1,\pi_2)$. 
It illustrates that $\rho$ acts as a characteristic radius beyond which it is preferable to destroy mass than transport it. 
This phenomenon is sharp in the case of $\TV$ (it is known when $\epsilon=0$ that $\spt(\pi)\subset\{(x,y), \C(x,y)\leq 2\rho\}$) while there is a smooth dampening as $\C$ increases for $\KL$.

\subsection{Translation invariant Sinkhorn}
\label{sec:ti-sinkhorn}

In sharp contrast with balanced OT, UOT Sinkhorn algorithm might converge slowly, even if $\epsilon$ is large.
For instance when marginals are penalized with the Kullback-Leibler divergence $\D_\phi=\rho\KL$, Sinkhorn converges linearly at a rate $(1+\tfrac{\epsilon}{\rho})^{-1}$ for the sup-norm $\norm{\cdot}_\infty$~\cite{chizat2016scaling}.
This rate is close to $1$ and thus the convergence is slow when $\epsilon\ll\rho$.
This issue motivates in~\cite{sejourne2022faster} the introduction of a \emph{translation invariant} formulation of the dual problem with associated Sinkhorn-like iterations enjoying better convergence rates.

\myparagraph{Translation invariant dual functionals.}
This invariance is motivated by a difference between balanced and unbalanced OT.
The balanced OT problem corresponds to using $\phi_1(x)=\phi_2(x)=\iota_{\{1\}}(x)$ (i.e. $\phi(1)=0$ and $\phi(x)=+\infty$ otherwise), such that $\phi^*(x)=x$.
In this case, the dual functional $\Dd_\epsilon$ reads
\begin{align*}
\Dd_\epsilon(\f,\g) &= \int_{\X} \f(x)\dd\al(x) +  \int_{\X} \g(y)\dd\be(y)\\
&\quad- \epsilon\int_{\X^2}\big(e^{(\f(x) + \g(y) - \C(x,y)) / \epsilon} - 1 \big)\dd\al(x)\dd\be(y).
\end{align*}
A key property is that for any constant translation $\la\in\RR$, $\Dd_\epsilon(\f+\la,\g-\la)=\Dd_\epsilon(\f,\g)$ (because $\phi^*$ is linear), while this does not hold in general for UOT.
In particular, optimal $(\f^\star,\g^\star)$ for balanced OT are only unique up to such translation $(\f^\star + \la,\g^\star - \la)$, while for $\UOT$ with strictly convex $\phi^*$ such as $\KL$, the dual problem has a unique pair of maximizers $(\f^\star,\g^\star)$. 

The following example emphasizes the impact of this lack of translation invariance (TI) on the convergence. 
Let $(\f^\star,\g^\star)=\arg\max\Dd_\epsilon$ for some input measures $(\al,\be)$, with $\D_\phi=\rho\KL$.
If one initializes Sinkhorn algorithm with $\f_0 = \f^\star + \tau$ for some $\tau\in\RR$, $\UOT$ Sinkhorn iterates with $\D_\phi=\rho\KL$ read $\f_t=\f^\star + (\tfrac{\rho}{\epsilon + \rho})^{2t}\tau$.
Thus iterates are sensitive to translations, and the error $(\tfrac{\rho}{\epsilon + \rho})^{2t}\tau$ decays slowly when $\epsilon\ll\rho$.

We solve this issue by explicitly enforcing translation invariance using a new dual functional defined below.
\begin{definition}[Translation-invariant functionals]\label{def:trans-inv}
The TI dual function is defined as
\begin{align}
\Hh_\epsilon(\Bf,\Bg)&\triangleq\sup_{\la\in\RR}\Dd_\epsilon(\Bf + \la, \Bg - \la).
\end{align}
We also define the optimal translation as
\begin{align}\la^\star(\Bf,\Bg) \triangleq \underset{\la\in\RR}{\argmax}
\Dd_\epsilon(\Bf + \la, \Bg - \la).
\end{align}
We assume $\phi^*$ strictly convex, so that $\la^*$ is unique.
\end{definition}
By construction, one has $\Hh_\epsilon(\Bf+\la,\Bg-\la)=\Hh_\epsilon(\Bf,\Bg)$, making the functional \emph{translation invariant}.
To emphasize this invariance we denote by $(\Bf,\Bg)$ potentials defined up to such translation.
Note that maximizing $\Dd_\epsilon(\f,\g)$ or $\Hh_\epsilon(\Bf,\Bg)$ yields the same value $\UOT(\al,\be)$ and one can switch between the maximizers using 
\begin{align*}
(f,g) = (\Bf + \la^\star(\Bf,\Bg),\Bg - \la^\star(\Bf,\Bg))\,.
\end{align*}

Note that any term of the form $\Bf\oplus\Bg$ is left unchanged by the translation $(\Bf+\la,\Bg-\la)$.
This invariance is also preserved when $\epsilon=0$, in which case we replace $- \epsilon\dotp{\al\otimes\be}{\big(e^{\frac{\Bf\oplus\Bg - \C}{\epsilon}} - 1\big)}$ by the constraint $\Bf\oplus\Bg\leq\C$.
This invariance is exploited in~\cite{sejourne2022faster} to derive a Frank-Wolfe algorithm on the translation-invariant dual formulation.

In general there is a priori no closed form for $\la^\star(\Bf,\Bg)$.
However, the case $\phi_i(x)=\rho_i(x\log x -x+1)$ and $\phi_i^*(x)=\rho_i(e^{-x/\rho_i} - 1)$ (corresponding to $\D_{\phi_i}=\rho_i\KL$) enjoys simple closed form expressions.
It is useful to rewrite $\Hh_\epsilon$ explicitly, as stated by the following result.
\begin{proposition}
	\label{prop-kl-opt-trans}
	For the $\KL$ divergence relaxing the marginal constraint, one has 
	\eql{\label{eq-opt-trans-kl}
		\la^\star(\Bf,\Bg) = \tfrac{\rho_1\rho_2}{\rho_1 + \rho_2} \log\Big[\frac{\dotp{\al}{e^{-\Bf / \rho_1}}}{ \dotp{\be}{e^{-\Bg / \rho_2}}}\Big].
	}
	Furthermore, setting $\tau_1=\tfrac{\rho_1}{\rho_1 + \rho_2}$ and $\tau_2=\tfrac{\rho_2}{\rho_1 + \rho_2}$, one has
	\begin{align}
	\Hh_\epsilon&(\Bf,\Bg) =  \rho_1 m(\al) + \rho_2 m(\be) -\epsilon\dotp{\al\otimes\be}{e^{\tfrac{\Bf\oplus\Bg - \C}{\epsilon}} - 1}\nonumber\\
	&- (\rho_1 + \rho_2)\Big(\dotp{\al}{ e^{-\Bf / \rho_1} }\Big)^{\tau_1} \Big(\dotp{\be}{ e^{-\Bg / \rho_2} }\Big)^{\tau_2}.\label{eq-dual-form-hell}
	\end{align}
	In particular when $\rho_1 = \rho_2=\rho$ and $\epsilon=0$,
	\begin{align*}
	\Hh_0(\Bf,\Bg) = \rho \Big[m(\al) +  m(\be)- 2\sqrt{   \dotp{\al}{ e^{-\Bf/\rho} }  \dotp{\be}{ e^{-\Bg/\rho} }   }\Big].
	\end{align*}
\end{proposition}
Note that Equation~\eqref{eq-opt-trans-kl} can be computed in $O(N)$ time, and stabilized via a logsumexp reduction (see Equation~\eqref{eq:lse-trick}).

\myparagraph{Translation-Invariant Sinkhorn.}
We detail below an alternating dual ascent procedure on $\Hh_\epsilon$ to optimize it, which is called \emph{TI-Sinkhorn} in~\cite{sejourne2022faster}.
This procedure converges to invariant potentials (defined up to translation).
It admits closed forms in the setting $\D_\phi=\rho\KL$ (Proposition~\ref{prop:ti-sink-kl}), and it is proved in this special case to converge faster than standard Sinkhorn (Theorem~\ref{thm:cv_rate_ti_sinkhorn}).
To formalize the TI-Sinkhorn algorithm, we introduce the following operators
\begin{gather}\label{eq:ti-sink-operators}
	\begin{aligned}
		&\Psi_1: \Bf\mapsto \argmax_{\Bg} \Hh_\epsilon(\Bf,\Bg),\\
		&\Psi_2: \Bg\mapsto \argmax_{\Bf} \Hh_\epsilon(\Bf,\Bg),\\
		&\Phi: (\Bf,\Bg)\mapsto (\Bf + \la^\star(\Bf,\Bg), \Bg - \la^\star(\Bf,\Bg)).
	\end{aligned}
\end{gather}
Operators $(\Psi_1,\Psi_2)$ characterize TI-Sinkhorn updates, while $\Phi$ is the translation operator which converts optimal invariant potentials $(\Bf^\star,\Bg^\star)$ optimal for $\Hh_\epsilon$ into potentials $(\f^\star,\g^\star)$ optimal for $\Dd_\epsilon$.
The operators $(\Psi_1,\Psi_2)$ are consistent with the translation invariance property of $\Hh_\epsilon$, i.e. one has $\Psi_1(\Bf + \la) = \Psi_1(\Bf) - \la$.

When $\D_{\phi_i}$ is smooth, one can reformulate $(\Psi_1,\Psi_2)$ using operators $(\Ss_\al,\Ss_\be)$ and $\aprox{\phi^*}$ as stated in the following result.
\begin{proposition}\label{prop-h-sink-general}
	For fixed $(\Bf,\Bg)$, write $\hf\triangleq\Ss_\al(\Bf)$ and $\hg\triangleq\Ss_\be(\Bg)$ (defined Equation~\eqref{eq:defn-softmin-func}).
	Then $(\psi_1 \eqdef \Psi_1(\Bg), \psi_2 \eqdef \Psi_2(\Bf))$ satisfy the fixed points
	\begin{gather}\label{eq:ti-sink-general}
	    \begin{aligned}
	\psi_1 = &-\aprox{\phi^*_2}\big(- \hg + \la^\star(\psi_1, \Bg)\big) - \la^\star(\psi_1, \Bg),\\
	\psi_2 = &-\aprox{\phi^*_1}\big(- \hf - \la^\star(\Bf, \psi_2)\big) +\la^\star(\Bf, \psi_2).
	\end{aligned}
	\end{gather}
\end{proposition}

An implementation of TI-Sinkhorn updates is detailed in~\cite{sejourne2022faster} for general (but smooth) $\D_\phi$.
The computation of $\Psi_1(\Bf)$ relies on sub-iterations which estimates $\psi_1$ and $\la^\star$ iteratively until convergence.
The translation $\la^\star(\Bf,\Bg)$ is estimated using Newton methods since the dual functional is smooth.
All in all, the complexity of Sinkhorn iterations remains $O(N^2)$ (to compute $\hf$ and $\hg$), similar to unbalanced Sinkhorn, because the sub-iterations have a complexity of $O(N)$.
However, when $\D_{\phi_i}=\rho_i\KL$, closed forms can be derived, which avoid the need of sub-iterations to compute $\Psi_1(\Bf)$.
We give the result when $\rho_1=\rho_2$, and refer to~\cite{sejourne2022faster} for the case $\rho_1\neq\rho_2$.

\begin{proposition}\label{prop:ti-sink-kl}
	For fixed $(\Bf,\Bg)$, assuming $\rho_1=\rho_2=\rho$ and $\D_\phi=\rho\KL$, denoting $\xi \triangleq \tfrac{\epsilon}{\epsilon+2\rho}$, one has
	\begin{align*}
		\Psi_1(\Bf) = \hg + \xi \Smin{\be}{\rho}(\hg), \: \Psi_2(\Bg) = \hf + \xi  \Smin{\al}{\rho}(\hf) \\
		\text{where}\:
		\choice{
			\hg \triangleq \tfrac{\rho}{\rho+\epsilon}\Smin{\al}{\epsilon}(\C-\Bf)-\tfrac{1}{2}\tfrac{\epsilon}{\rho+\epsilon} \Smin{\al}{\rho}(\Bf),\\
			\hf \triangleq \tfrac{\rho}{\rho+\epsilon}\Smin{\be}{\epsilon}(\C-\Bg)-\tfrac{1}{2}\tfrac{\epsilon}{\rho+\epsilon} \Smin{\be}{\rho}(\Bg).		
		}
	\end{align*}
\end{proposition}

\myparagraph{Convergence rate of Sinkhorn and TI-Sinkhorn.}
The quantitative convergence analysis of Sinkhorn is quantified using two norms.
The first one is the sup-norm $\norm{\f}_\infty \eqdef \max |\f|$, and the second one is the Hilbert pseudo-norm $\norm{\f}_\star \eqdef \inf_{\la\in\RR}\norm{\f + \la}_\infty = \tfrac{1}{2}(\max \f - \min\f)$, which compares functions up to constant translations.
The Hilbert norm is useful to analyze the convergence of balanced Sinkhorn algorithm, as well as TI-Sinkhorn, since iterates share the same invariance to translations.
By contrast, the convergence of unbalanced Sinkorn is quantified for the sup-norm since it is not invariant.

To make the comparison between iterates of Sinkhorn and TI-Sinkhorn informative, one first computes iterates $(\Bf_t,\Bg_t)$ with $(\Psi_1,\Psi_2)$, then compares their translation $\Phi(\Bf_t,\Bg_t)$ to the optimal potentials $(\f^\star,\g^\star)$ of $\Dd_\epsilon$.
By definition of $\Hh_\epsilon$ and $\Phi$ (Definition~\ref{def:trans-inv} and Equation~\eqref{eq:ti-sink-operators}), if $(\Bf_t,\Bg_t)$ are optimal for $\Hh_\eps$, then $\Phi(\Bf_t,\Bg_t)$ are optimal for $\Dd_\epsilon$.
It yields the following convergence rates when $\D_{\phi_i}=\rho_i\KL$. It combines results from~\cite{chizat2016scaling, sejourne2022faster}.

\begin{theorem}\label{thm:cv_rate_ti_sinkhorn}
	Assume $\D_{\phi_i}=\rho_i\KL$.
	Write $(\f^\star,\g^\star)=\argmax\Dd_\epsilon$. 
	Take $\Bf_{t}$ obtained by $t$ iterations of the TI-Sinkhorn map $\Psi_2\circ\Psi_1$, starting from the function $\Bf_0$. One has
	\begin{align*}
	\norm{\Phi((\Bf_t, \Psi_1(\Bf_t))) - (\f^\star,\g^\star)}_\infty 
	\leq 2\bar{\kappa}^{t}\norm{\Bf_0 - \Bf^\star}_\star,
	\end{align*}
	where $\bar{\kappa}\triangleq (1 + \tfrac{\epsilon}{\rho_1})^{-1}\kappa_\epsilon(\al)(1 + \tfrac{\epsilon}{\rho_2})^{-1}\kappa_\epsilon(\be)$, $\norm{(\Bf,\Bg)}_\infty = \norm{\Bf}_\infty + \norm{\Bg}_\infty$, and where $\kappa_\epsilon(\al)<1$ is the contractance rate of $\Ss_\al$ for $\norm{\f}_\star$, i.e. such that
	\begin{align*}
	\norm{\Ss_\al(\Bf) - \Ss_\al(\Bg)}_\star \leq \kappa_\epsilon(\al)\norm{\Bf - \Bg}_\star.
    \end{align*}

By comparison, starting from $\f_0$, denote $\f_{t}$ obtained by $t$ iterations of the Sinkhorn map $\Aa\Ss_\be\circ\Aa\Ss_\al$ (Definition~\ref{def:sinkhorn}). One has
\begin{align*}
    \norm{(\f_t,\Aa\Ss_\al(\f_t)) - (\f^\star,\g^\star)}_\infty\leq 2\kappa^t \norm{\f_0 - \f^\star}_\infty,
\end{align*}
where $\kappa\eqdef (1 + \tfrac{\epsilon}{\rho_1})^{-1}(1 + \tfrac{\epsilon}{\rho_2})^{-1}$.
\end{theorem}

Note that the rate of TI-Sinkhorn is improved compared to Sinkhorn by a factor $\kappa_\epsilon(\al)\kappa_\epsilon(\be)$, which is a factor independent of $\rho$.
Being independent of $\rho$ means this rate is induced solely by the entropic regularization, not by the relaxation of marginal constraints.
The absence of such a term in the convergence rate of unbalanced Sinkhorn means that it does not benefit from entropic regularization for large $\rho$, whatever the value of $\epsilon$.

Local and global estimation of the contractance rate $\kappa_\epsilon(\al)$ of $\Ss_\al$ are detailed respectively in~\cite{knight2008sinkhorn} and~\cite{birkhoff1957extensions, franklin1989scaling}.
When $(\al,\be)$ are discrete measures and $\C=(\C_{ij})$ is a matrix, the latter estimate reads $\kappa_\epsilon(\al)\leq 1 - \tfrac{2}{1 + \eta}$, where $\eta=\exp(-\tfrac{1}{2\epsilon}\max_{i,j,k,l}(\C_{j,k} + \C_{i,l} - \C_{j,l} - \C_{i,k}))$.
Note that $\eta$ depends on $\al$ via its support.

\subsection{Sinkhorn divergences}
\label{sec:sinkdiv}

\myparagraph{The price of regularization.}
Since entropic regularization leads to efficient numerical schemes and is statistically more appealing (see next Section~\ref{sec:sample-comp}), it is tempting to use $\OTb$ \hyphen{off-the-shelf} as a loss function for ML or imaging sciences applications.
For instance many learning tasks such as generative learning~\cite{arjovsky2017wasserstein} or VAE~\cite{kingma2013auto} seek to optimize a parameterized model $\al_\thh$ with parameter $\thh\in\RR^d$ such that it approximates well a data distribution $\be_N$.
To train the model one minimizes
\begin{align*}
	\min_{\thh\in\RR^d} \Ll(\al_\thh,\be_N) + \Rr(\al_\thh),
\end{align*}
where $\Ll$ is a loss quantifying the discrepancy between the model and the dataset, and $\Rr$ is a regularization of the model.
Here one could take $\Ll=\OTb$ and expect that $\al_\thh\simeq\be_N$ after minimization of $\Ll(\cdot,\be_N)$.

\begin{figure}[h]
	\centering
	\begin{tikzpicture}[scale=0.5]
	\node[inner sep=0pt] (waffle) at (2, 0)
	{\includegraphics[scale=0.13]{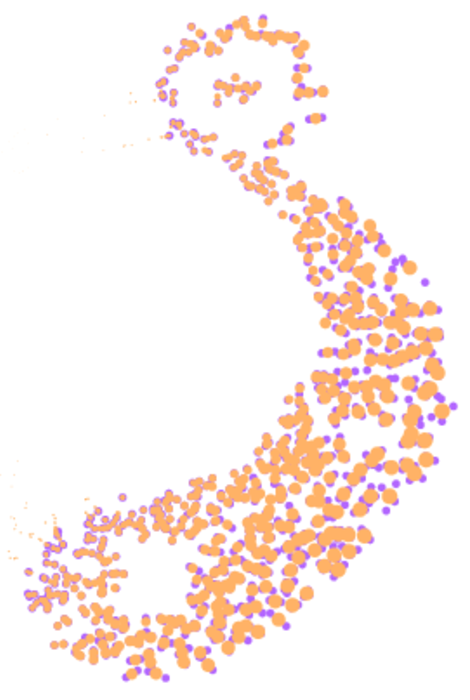}};
	\node[inner sep=0pt] (waffle) at (8, 0)
	{\includegraphics[scale=0.13]{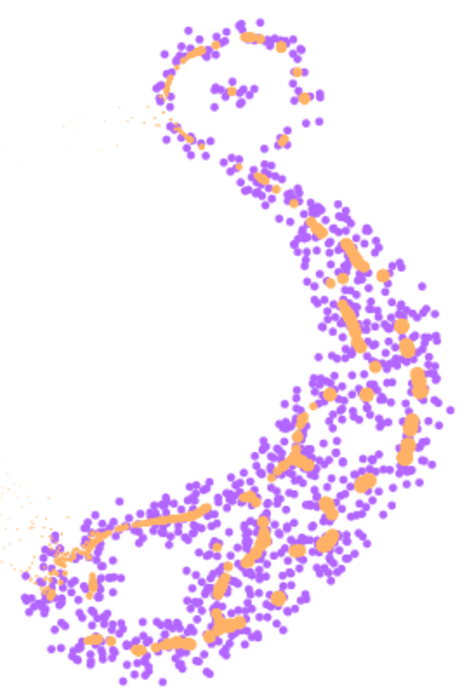}};
	\node[inner sep=0pt] (waffle) at (14, 0)
	{\includegraphics[scale=0.13]{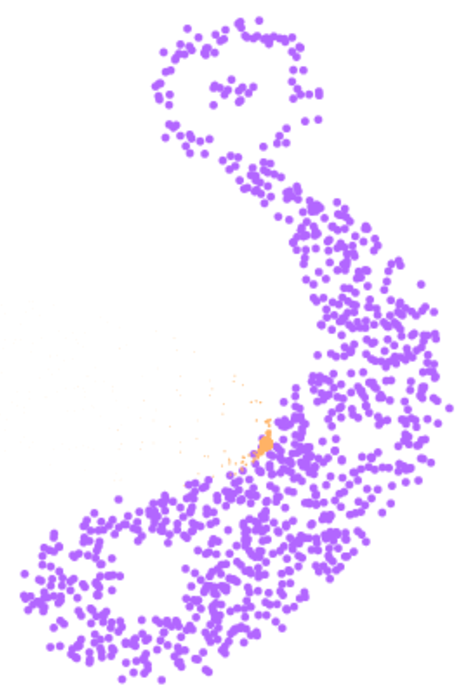}};
	\draw [->, thick] (-2,-3.5) -- (18,-3.5);
	\node at (18,-2.5) {$\bm{\epsilon}$} ;
	\end{tikzpicture}
	\caption{\textit{
		Display of $\arg\min\OTb(\cdot,\be)$ as $\epsilon$ increases, where $\be$ (in blue) is a point cloud discretization of a crescent. The model $\al$ (red) is also a points cloud whose support is optimized via a gradient flow (see Section~\ref{sec:mmd}, Figure~\ref{fig:grad_flow_kernel} for a detailed protocol).}
	}
	\label{fig1:illust-entropic-bias}
\end{figure}

However, those computational and statistical gains are paid at the price of losing metrization of weak* topology.
In fact, for any $(\al,\be)\in\Mmp(\X)$, it does not satisfy $\OTb(\al,\al)=0$ and even worse, one has in general $\al \notin\arg\min\OTb(\cdot,\al)$.
Consequently, training via gradient descent does not produce a model such that $\al_\thh\simeq\be$.
More precisely, it is shown in~\cite{sejourne2019sinkhorn} that if $\epsilon \rightarrow +\infty$, then $\al^\star=\arg\min_\al\OTb(\al,\be)$ degenerates to a Dirac mass $\de_{x^\star}$, where $$x^\star = \arg\min_{z\in\X}\int\C(z, y)\dd\be(y).$$ 
See also Figure~\ref{fig1:illust-entropic-bias} for an illustration of this phenomenon.

This impact of entropic regularization, which we call \emph{entropic bias}, is problematic in applications.
It forces a trade-off between taking a small $\epsilon$, which yields a well-posed approximation of OT at an expensive computational cost, and taking a large $\epsilon$, in which case the approximation is fast to obtain but differs radically from unregularized OT.

\myparagraph{Correcting the bias.}
The above issue motivated a variant of $\OTb$ which might correct the above shortcomings.
It is called the \emph{Sinkhorn divergence} and is defined as follows.

\begin{definition}[Sinkhorn divergence]\label{def:sink-div}
Consider $\OTb$ (see Definition~\ref{def:uot-entropic}) with $\D_{\phi_1}=\D_{\phi_2}$.
The unbalanced Sinkhorn divergence is defined as
\begin{align*}
\Sb(\al,\be)\triangleq \OTb(\al,\be) - \tfrac{1}{2}\OTb(\al,\al) - \tfrac{1}{2}\OTb(\be,\be) + \tfrac{\epsilon}{2}(m(\al) - m(\be))^2,
\end{align*}
where $m(\al)\eqdef \int_\X\dd\al$ denotes the total mass of the measure.
\end{definition}
Note that the term $\tfrac{\epsilon}{2}(m(\al) - m(\be))^2$ can be removed in the balanced setting, because if $m(\al)\neq m(\be)$ then $\Sb(\al,\be)=+\infty$, and it is zero if $m(\al)= m(\be)$.

This divergence satisfies $\Sb(\al,\al)=0$, but it is not immediately clear whether it is actually a positive quantity.
The quantity 
\begin{align*}
\OTb(\al,\be) - \tfrac{1}{2}\OTb(\al,\al) - \tfrac{1}{2}\OTb(\be,\be)
\end{align*}
was introduced in the balanced OT setting in~\cite{ramdas2017wasserstein} for statistical testing, and in~\cite{genevay2018learning, salimans2018improving} for generative learning.
However in those works it was only presented as a heuristic, and observed empirically to yield improved performance in training.
This quantity is proved in~\cite{genevay2018learning} to converge as $\epsilon\rightarrow\infty$ to a kernel norm $\tfrac{1}{2}\norm{\al - \be}^2_{\Hh^*}$ with kernel $k=-\C$. This somehow explains that using Sinkhorn divergences is a way to interpolate between OT and kernel methods.
One of the main contributions of~\cite{feydy2018interpolating} and~\cite{sejourne2019sinkhorn} is to prove that under conditions on $\C$, the Sinkhorn divergence recovers properties of unregularized OT such as positivity, definiteness, and metrization of the weak* convergence.

\begin{theorem}[Properties of the Sinkhorn divergence] \label{thm:sink-unb}
  Assume the cost $\C$ is symmetric, Lipschitz, and that $k_\epsilon = e^{- \C / \epsilon}$ is a positive universal kernel. 
  For any $\epsilon >0$, for any entropy $\phi$, the Sinkhorn divergence $\Sb(\al,\be)$ is positive, definite and convex in $\al$ and $\be$ (but not jointly).
  
  When $\D_\phi = \rho\KL$, $\D_\phi=\rho\TV$ or $\D_\phi=\iota_{(=)}$, the Sinkhorn divergence $\Sb$ metrizes the convergence in law.
  For any sequence $(\al_n)_n$ in $\Mmpp(\X)$ (or in $\Mmpo(\X)$ for balanced OT), we have $\al_n\rightharpoonup\al\Longleftrightarrow\Sb(\al_n,\al)\rightarrow 0$.
\end{theorem}

The take-home message of Theorem~\ref{thm:sink-unb} is that using $\Ll=\Sb$ is preferable, because it retains the computational advantages of $\OTb$ and the key mathematical properties of $\OT$ w.r.t. the weak* topology, which are both desirable to train models in practice.
Indeed, as discussed in~\cite{feydy2018interpolating, knight2014symmetry}, the additional terms $\OTb(\al,\al)$ and $\OTb(\be,\be)$ require even less computation time  than $\OTb(\al,\be)$.
Thus computing $\Sb(\al,\be)$ takes almost as much time as $\OT_\epsilon(\al,\be)$.
Furthermore, the properties of the Sinkhorn divergence hold for any $\epsilon >0$, which allows to consider $\Sb$ as an interesting object in itself, not only as a proxy for unregularized OT when $\epsilon$ is small.

\rev{This result holds for $\C(x,y)=\norm{\h(x) - \h(y)}_2^p$ where $1\leq p\leq 2$ and $\h:\X\rightarrow\RR^d$ is some feature map, which is a typical setup in machine learning applications.
However, it does not hold for the cost $\C(x,y)=-\log\cos^2(\min(\tfrac{\norm{x-y}_2}{\rho},\tfrac{\pi}{2}))$. When $\epsilon=2$, this corresponds to the kernel $\cos(\min(\tfrac{\norm{x-y}_2}{\rho},\tfrac{\pi}{2}))$ which is not positive definite as soon as the cut-off value $\tfrac{\pi}{2}$ is attained. Nevertheless, when the cut-off is not reached, the kernel $\cos(\norm{x-y}_2 / \rho))$ is positive definite (as a limit of positive definite kernels thanks to the Taylor expansion). Thus for the WFR setting, the Sinkhorn divergence is ``locally'' positive in $\Mmp(\X)$.}

The proof of Theorem~\ref{thm:sink-unb} (especially the positivity and definiteness) involves the following result, which is interesting because it highlights the connection between regularized optimal transport and kernel norms.
\begin{proposition}[The Sinkhorn divergence is bounded from below by a kernel norm]\label{prop:Seps-ineq-norm}
For any entropy $\phi$, write $(\f_\al,\g_\be)$ optimal symmetric potentials satisfying $\OTb(\al,\al)=\Dd_\epsilon(\f_\al,\f_\al)$ and $\OTb(\be,\be)=\Dd_\epsilon(\g_\be,\g_\be)$. 
Then, one has
\begin{align}
\Sb(\al,\be) \geq \tfrac{\epsilon}{2} \Vert  \al e^{\frac{\f_{\al}}{\epsilon}} - \be e^{\frac{\g_{\be}}{\epsilon}} \Vert^2_{k_\epsilon}.
\end{align}
\end{proposition}

\myparagraph{Extension -- asymmetric Sinkhorn divergence.}
In general one frequently takes $\D_{\phi_1}=\D_{\phi_2}$, but it is also possible to consider penalties $\D_{\phi_1}(\pi_1|\al)$ and $\D_{\phi_2}(\pi_2|\be)$ with $\phi_1\neq\phi_2$.
For instance, take $\D_{\phi_1}=\iota_{(=)}$ and $\D_{\phi_2}=\rho\KL$ (as in~\cite{vincent2021semi}), or $\D_{\phi_i}=\rho_i\KL$ with $\rho_1\neq\rho_2$ (as in~\cite{sejourne2021unbalanced}).
It is relevant in domain adaptation where $\al$ is a source dataset on which a predictor was trained, and $\be$ is a similar but shifted dataset on which we want to transfer the learned predictor~\cite{courty2014domain}.

In this setting it is possible to define a Sinkhorn divergence which is positive, but no longer symmetric.
It reads
\begin{align*}
	\Sb^{(\phi_1,\phi_2)}(\al,\be) &= \OTb^{(\phi_1,\phi_2)}(\al,\be) - \OTb^{(\phi_1,\phi_1)}(\al,\al) - \OTb^{(\phi_2,\phi_2)}(\be,\be) \\
	&+ \tfrac{\epsilon}{2}(m(\al) - m(\be))^2,
\end{align*}
where $\OTb^{(\phi_1,\phi_2)}$ is the regularized OT program penalized with $(\D_{\phi_1},\D_{\phi_2})$.
Proposition~\ref{prop:Seps-ineq-norm} also holds for the above formula, hence the positivity of $\Sb^{(\phi_1,\phi_2)}$ when $k_\epsilon = e^{- \C / \epsilon}$ is a positive kernel.
Recall the computation of $\OTb^{(\phi_1,\phi_2)}$ can be performed via the iterations of Definition~\ref{def:sinkhorn}.

\myparagraph{Numerical experiments -- 2D Gradient flows.}
We present numerical experiments on gradient flows from~\cite{sejourne2019sinkhorn}.
Given a set of particles $\theta = \{(x_i, r_i)_i \}$ with coordinates $x_i \in\RR^d$ and masses $r_i \in\RR_+$, one wishes to study their trajectories to minimize some potential $\theta\mapsto F(\theta)$.
The particles initialized at $t=0$ by $\theta_0$ undergo the dynamic $\partial_t\theta(t) = - \nabla F(\theta(t))$.
These dynamical systems are ubiquitous to model phenomena such as physical systems or crowd motions. 
They also gained attention in machine learning to model the dynamics of neural networks training~\cite{chizat2018global}.

We consider the same setting as~\cite{chizat2021sparse}.
We consider a target measure $\be\in\Mmp(\RR^d)$ and the potential $\al\mapsto\Sb(\al,\be)$, as one would do in e.g. generative learning in imaging~\cite{arjovsky2017wasserstein}.
The measure $\al$ represents the model we train, parameterized as $\al_\theta = \sum_i^n  r_i^2 \de_{x_i}$ with parameter $\theta = ((x_i, r_i))_i\in(\RR^2\times\RR_+)^n$.
Minimizing $\Sb(\cdot,\be)$ amounts \rev{to run a ``standard'' gradient update w.r.t. $x_i$, and a mirror gradient update w.r.t.} $r_i$., which reads
\begin{align*}
	x_i^{(t+1)} & = x_i^{(t)} - \eta_x \nabla_{x_i} \Sb(\al_\theta^{(t)}, \be),                  \\
	r_i^{(t+1)} & = r_i^{(t)}.\exp\big( - 2\eta_r \nabla_{r_i} \Sb(\al_\theta^{(t)}, \be) \big),
\end{align*}
where $(\eta_x, \eta_r)>0$ are two learning rates.
We retrieve a continuous time gradient flow when $(\eta_x, \eta_r)\rightarrow 0$.
Using such a model $\al_\theta$ and such updates is proved in~\cite{chizat2021sparse} to be equivalent to an unbalanced gradient flow in the space $\Mmp(\X)$, in contrast with classical (balanced) Wasserstein flows optimizing over $\Mmp_1(\X)$.
The update on $r_i$ is called a mirror descent step, and is used to enforce that $r_i\geq 0$.
The parameterization of $\al_\thh$ using $r_i^2$ and not $r_i$ improves the numerical stability, because for the latter parameterization, the mirror update reads
\begin{align*}
	r_i^{(t+1)} & = r_i^{(t)}.\exp\Big( - 2\tfrac{\eta_r}{r_i^{(t)}} \nabla_{r_i} \Sb(\al_\theta^{(t)}, \be) \Big),
\end{align*}
and the extra $1/r_i$ term might make the exponential overflow for small $r_i$.

We run the experiments in several settings.
We always take the Euclidean distance $\C(x,y)=\norm{x-y}^2_2$ on the unit square $[0,1]^2$, constant learning rates $(\eta_x, \eta_r) = (60, 0.3)$, $\D_\phi=\rho\KL$, a radius $\sqrt{\rho}=\sqrt{10^{-1}}$, and a default blur radius of $\sqrt{\epsilon}=\sqrt{10^{-3}}$.
In each timeframe we display iterations $[5, 10, 20, 50, 300]$ of the gradient descent steps.
Each dot represents a particle, and the diameter represents its mass.

\begin{figure}
\centering
\begin{subfigure}{\textwidth}
    \includegraphics[width=\textwidth]{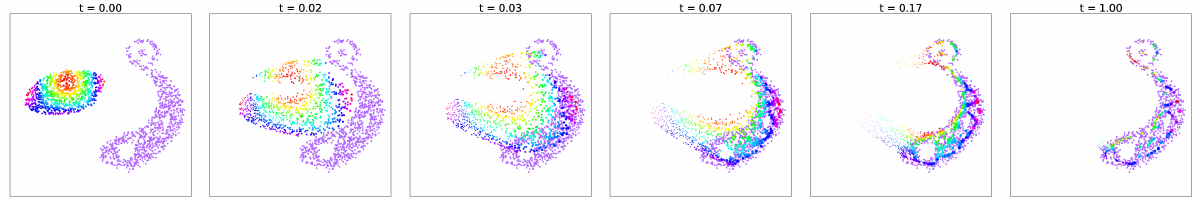}
    \caption{$\Ll=\OTb$ with $\epsilon=10^{-3}$.}
    \label{fig:first}
\end{subfigure}
\hfill
\vspace*{0.2em}
\begin{subfigure}{\textwidth}
    \includegraphics[width=\textwidth]{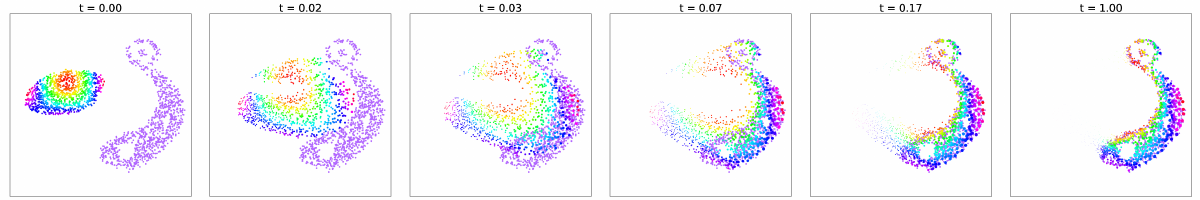}
    \caption{$\Ll=\Sb$ with $\epsilon=10^{-3}$.}
    \label{fig:second}
\end{subfigure}
\hfill
\vspace*{0.2em}
\begin{subfigure}{\textwidth}
    \includegraphics[width=\textwidth]{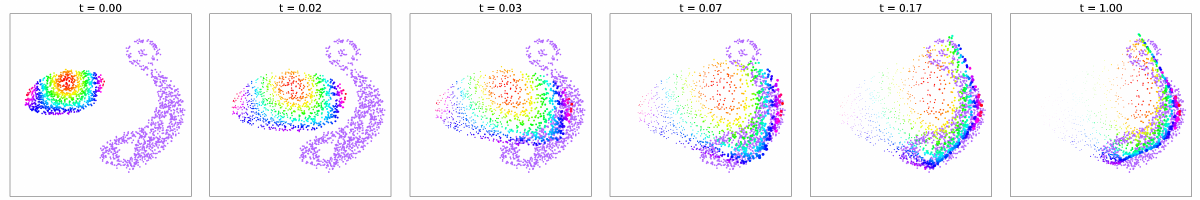}
    \caption{$\Ll=\Sb$ with $\epsilon=10^{-1}$.}
    \label{fig:third}
\end{subfigure}
        
\caption{Display of the gradient flow minimizing $\theta\mapsto\Ll(\al_\theta,\be)$ for different OT losses and entropy strength $\epsilon$. When $\epsilon$ is large and $\C(x,y)=\norm{x-y}^2_2$, $\Sb$ approaches a degenerate kernel norm which is the squared norm between the measure's means, hence the result of Figure~\ref{fig:third}.}
\label{fig:flow-reg}
\end{figure}

Figures~\ref{fig:flow-reg} (rows 1 and 2) show the difference between using $\OTb$ and the (debiased) Sinkhorn divergence $\Sb$.
Note that for $\OTb$ (row 1) the model $\al_\theta$ concentrates (i.e. suffers the \emph{entropic bias}) while for $\Sb$ it approaches $\be$ up to details of size $\sqrt{\epsilon}$.
One the same figure, comparing rows 2 and 3 shows the influence of $\epsilon$, which operates a low pass smoothing.
If $\epsilon$ is chosen too large then $\al_\theta$ discards finer details.

\subsection{Sample complexity}
\label{sec:sample-comp}
In statistics and learning one usually has access to $N$ samples $(x_1,\ldots,x_N)$ and $(y_1,\ldots,y_N)$ drawn from inaccessible probabilities $\al$ and $\be$.
A problem is to design finite estimators $\al_N(x_1,\ldots,x_N)$ and $\be_N(y_1,\ldots,y_N)$ of $(\al,\be)$ such that $\OT(\al_N,\be_N)$ is a good approximation of $\OT(\al,\be)$.
An issue is that OT suffers from a curse of dimensionality with the naive plugin estimator~\cite{dudley1969speed}, i.e. one has $\EE[|\OT(\al_N,\be_N) - \OT(\al,\be)|]=O(N^{-1/d})$, where $\al_N=\tfrac{1}{N}\sum_{i=1}^N \de_{x_i}$ and $\be_N=\tfrac{1}{N}\sum_{i=1}^N \de_{y_i}$.
The rate on $N$ depends on the dimension, so that the number of samples needed to reach a given tolerance grows exponentially fast with the dimension.
Furthermore, it is proved in~\cite{singh2018minimax} that without any additional assumption, the above empirical estimators are optimal in the sense that there exists no other estimator of the form $\al_N(x_1,\ldots,x_N)$ and $\be_N(y_1,\ldots,y_N)$ yielding a better rate than $O(N^{-1/d})$.

To improve the rate, additional assumptions on the measures $(\al,\be)$ are necessary.
The rate can be refined by replacing the dimension $d$ of the ambient space (e.g. $\X=\RR^d$) by the intrinsic dimension (such as the Hausdorff dimension) of the input measures~\cite{weed2019sharp}, and by the lowest intrinsic dimension of the two considered inputs~\cite{hundrieser2022empirical}.
Considering densities smooth to the order $p$ reduces the complexity to $O(N^{-p/d})$~\cite{weed2019estimation}, but the estimators $(\al_N,\be_N)$ rely on wavelet decomposition or kernel estimation, which complicates the computation of $\OT(\al_N,\be_N)$.
Another approach consists in estimating $\OT$ using an approximation of the optimization program itself.
Other approximations are possible such as an SDP relaxation which leverages the smoothness of inputs~\cite{vacher2021dimension}.
For unbalanced OT in~\cite{vacher2022stability}, similar statistical complexities than OT can be obtained. 
Note also that \hyphen{breaking the curse} means a rate $O(N^{-r})$ where $r$ does not depend on $d$, but the constants might depend exponentially on $d$, as exhibited in~\cite{vacher2021dimension}, as well as for regularized OT (see next paragraph).

The problem of approximating $\OTb(\al,\be)$ with $\OTb(\al_N,\be_N)$, where we take $(\al_N,\be_N)$ to be the naive plugin estimators is studied in~\cite{genevay2018sample, mena2019statistical} for the balanced setting, and in~\cite{sejourne2019sinkhorn} for unbalanced OT.
When $\X=\RR^d$, those results break the curse of dimensionality by transferring the exponential dependency on the dimension $d$ from the number of samples $N$ to the constants.
Assuming $(\al,\be)$ are sub-Gaussian distributions and $\C(x,y)=\norm{x-y}^2_2$ yields a rate $\EE[|\OTb(\al_N,\be_N)-\OTb(\al,\be)|]=O((\epsilon^{-\lceil 5d/4\rceil + 3} + 1)N^{-1/2})$ in~\cite{mena2019statistical}.
The works~\cite{genevay2018learning,sejourne2019sinkhorn} prove a similar rate under the assumption that $(\al,\be)$ are compactly supported measures, for general smooth cost $\C$.
The result of~\cite{sejourne2019sinkhorn} is an extension of~\cite{genevay2018learning} to unbalanced OT.
We summarize their results in the following theorem.

\begin{theorem}\label{thm:sample-comp-rate}
Assume $\phi^*$ and $\C$ are $\Cc^\infty$, and that $(\al,\be)$ are supported on a compact set. Then one has
\begin{align*}
\EE[|\OTb(\al_N,\be_N)-\OTb(\al,\be)|]=O((\epsilon^{-\lfloor d/2\rfloor} + 1)N^{-1/2}).
\end{align*}
\end{theorem}

Note that the use of the Sinkhorn divergence (Definition~\ref{def:sink-div}) is also interesting from a statistical perspective.
It is proved in~\cite{chizat2020faster} that for balanced OT, estimating $\OT(\al,\be)$ (not $\OTb$) using $\Sb(\al_N,\be_N)$ when $\C(x,y)=\norm{x-y}_2^2$ and $\al\neq\be$ yields an estimation error scaling as $O(N^{-2/d})$. This matches the rates of the (unregularized) estimator, but leads to a faster computational scheme thanks to Sinkhorn's iterations.

\section{Gromov-Wasserstein distances}
\label{sec:gw}

As detailed in the previous sections, OT allows one to compute assignments (or permutations) accounting for a metric similarity $\C$ between samples.
However such metric prior is not straightforwardly available in some tasks, for instance if the measures are defined on two different spaces $\X$ and $\Y$ and no cost $\C$ is defined on $\X\times\Y$.
Such a setting arises with graph data~\cite{hu2020open}.
They embed objects such as social networks~\cite{tabassum2018social}, molecules in biochemistry~\cite{gaudelet2021utilizing} or shapes~\cite{memoli2011gromov}.
One example where we seek an assignment between two graphs is the graph isomorphism problem, which is known to be NP-hard~\cite{fortin1996graph}.
Another practical setting where no cost is available happens when we seek an assignment between measures defined on Euclidean spaces of different dimensions, as is the case in biology when measurements are performed under distinct modalities~\cite{demetci2020gromov}.

\subsection{Comparing balanced metric measure spaces}

\begin{figure*}[h]
	\centering
	\begin{tabular}{c@{}c@{}c@{}c@{}c}
		{\includegraphics[width=0.28\linewidth]{figures/density_a.png}} & $\quad\quad$ &
		{\includegraphics[width=0.25\linewidth]{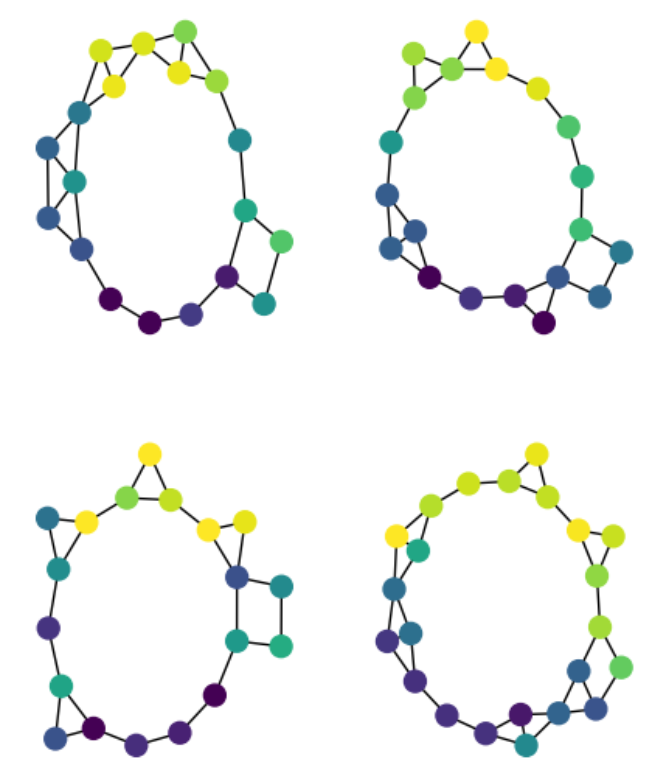}} & $\quad\quad$ &
		{\includegraphics[width=0.27\linewidth]{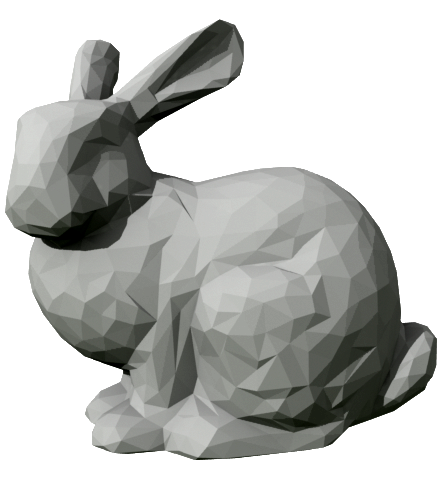}}
	\end{tabular}
	\caption{\textit{\rev{Examples of mm-spaces in ML.
			(Left) An Euclidean points cloud.
			(Center) A graph endowed with its shortest path distance and uniform distribution (from~\protect\cite{vayer2020fused}).
			(Right) A shape endowed with its geodesic distance and uniform distribution over the grid (from~\protect\cite{frwiki:184491466}).}
		}
	}
	\label{fig:exmpl-mm-space}
	\vspace*{1.0em}
\end{figure*}

\myparagraph{The metric measure space model.}
To compare measures defined on different spaces, one approach is to consider \emph{metric measure spaces} (abbreviated mm-spaces) which take both the metric and the measure into account together.
More precisely, it is a triple $\Xx=(\X, d_\X,\al)$ where $\X$ is a space endowed with a metric $d_\X$, and $\al\in\Mmp(\X)$.
It is a richer model where each measure is associated to a metric on its space.
A large variety of spaces fall under this framework: shapes or graphs with their geodesic distance~\cite{memoli2011gromov}, Riemannian manifolds~\cite{cheeger1997structure} (in particular Euclidean spaces), Finsler spaces~\cite{shen2001lectures}, finite dimensional Alexandrov spaces~\cite{burago1992ad}, groups equipped with the Haar measure~\cite{woess2000random} and fractals~\cite{kigami2001analysis} to name a few examples.
From the ML perspective, we provide in Figure~\ref{fig:exmpl-mm-space} illustrations of mm-spaces that are used in practice.
A first example is Euclidean point clouds in $\X=\RR^d$, endowed with the Euclidean distance $d_\X = \norm{\cdot-\cdot}_2$ and the uniform distribution over the samples $\al=\tfrac{1}{N}\sum_i \de_{x_i}$.
A second one is graphs or shapes where $\X$ is the set of nodes or a discretization of the shape, $d_\X$ is the shortest path/geodesic distance, and $\al$ is the uniform probability over the graph/shape.

\myparagraph{Comparing mm-spaces.}
For the same reasons as for measures, one needs a way of comparing two mm-spaces to quantify their similarity.
Sturm's distance~\cite{sturm2006geometry} combines ideas of the Gromov-Hausdorff distance~\cite{gromov1999metric, memoli2008gromov, memoli2011gromov} with OT.
It consists in mapping two mm-spaces $\Xx=(\X, d_\X,\al)$ and $\Yy=(\Y, d_\Y,\be)$ isometrically into a third space $(\Z,d_\Z)$, and performing OT on that third space.
It reads
\begin{align*}
	d_{Sturm}(\Xx,\Yy)\triangleq \inf_{(\Z,d_\Z), \psi_\X,\psi_\Y} \OT_{d_\Z}((\psi_\X)_\sharp\al, (\psi_\Y)_\sharp\be),
\end{align*}
where $d_\Z$ is the ground OT cost on $\Z$, $\psi_\X:\X\rightarrow\Z$ and $\psi_\Y:\Y\rightarrow\Z$ are isometric embeddings into $\Z$.
By comparison the Gromov-Hausdorff distance between two metric spaces reads
\begin{align*}
	\GH((\X,d_\X),(\Y,d_\Y)) \triangleq \inf_{(\Z,d_\Z), \psi_\X,\psi_\Y} \Ha_{d_\Z}(\psi_\X(\X),\psi_\Y(\Y)),
\end{align*}
where $(\Z,d_\Z), \psi_\X,\psi_\Y$ satisfy the same constraints as for $d_{Sturm}$, and $\Ha_{d_\Z}(\X,\Y)$ is the Hausdorff distance between two sets $\X,\Y\subset\Z$ and reads
\begin{align*}
	\Ha_{d_\Z}(\X,\Y)\triangleq\max\bigg\{ \sup_{x\in\X}\inf_{y\in\Y} d_\Z(x,y),\, \sup_{y\in\Y}\inf_{x\in\X} d_\Z(x,y) \bigg\}.
\end{align*}
A fundamental difference between Sturm's distance and the Gromov-Hausdorff distance is the extra information contained in the measure.
It allows defining integrals on $\X$ and $\Y$, which are smoother quantities than the suprema/infima used for the Gromov-Hausdorff distance.

Sturm's distance was motivated by theoretical considerations, namely comparing the curvature of spaces.
Unfortunately, it is difficult to implement exactly and efficiently,
since a triangle inequality constraint on $d_\Z$ as well as an isometry constraint on the maps $(\psi_\X,\psi_\Y)$ must be imposed.
Furthermore, OT is concave w.r.t. to the input cost $d_\Z$, thus the minimization in $d_\Z$ yields a non-convex problem.
It is the reason why we consider another distance between mm-spaces which is more amenable to computations.

\myparagraph{The Gromov-Wasserstein distance.}
A second approach to compare mm-spaces is the Gromov-Wasserstein distance~\cite{memoli2011gromov}.
This formulation consists in changing OT so that it becomes a quadratic assignment problem (QAP)~\cite{burkard1998quadratic, burkard2009assignment} instead of a linear one.

\begin{definition}\label{def:gw}
    The Gromov-Wasserstein distance between mm-spaces $(\Xx,\Yy)$ endowed with probabilities is defined as
\begin{align*}
    \GW(\Xx,\Yy) \eqdef \min_{\pi\in\Uu(\al,\be)} \Gg(\pi),
\end{align*}
where $\Uu(\al,\be)$ is the same constraint set as Definition~\ref{def:ot}, and
\begin{align*}
	\Gg(\pi) \eqdef \int_{\X^2\times\Y^2} \Om(|d_\X(x,x') - d_\Y(y,y')|)\dd\pi(x,y)\dd\pi(x',y'),
\end{align*}
where $\Om:\RR_+\rightarrow\RR_+$ is the same renormalization as Section~\ref{sec:conic-setups}.
\end{definition}
The similarity with OT stems from the use of the constraint set $\Uu(\al,\be)$.
The key difference is the quadratic dependence in the plan $\pi\otimes\pi$.
This quadratic dependence is what characterizes it as a QAP, and makes it an NP-hard problem to solve in general~\cite{burkard1998quadratic}.
It can be interpreted similarly to an OT program which maps pairs $(x,x')$ to $(y,y')$ such that $d_\X(x,x')$ is close to $d_\Y(y,y')$, under the constraint that $\pi$ is a transport plan between $(\al,\be)$.
The choice of $\Om$ is crucial for the metric properties of GW.
It is proved in~\cite{memoli2011gromov} that $\GW^{1/q}$ is a distance when $\Om(t)=t^q$ ($q\geq 1$).

The Gromov-Wasserstein distance and Sturm's distance both satisfy metric properties, up to an invariance to isometry.
One defines the equivalence relation $\Xx\sim\Yy$ between mm-spaces if there exists an isometry $\psi:\X\rightarrow\Y$ such that $d_\X(x,x')=d_\Y(\psi(x), \psi(x'))$ and $\psi_\sharp\al=\be$.
Note that $\psi$ preserves both the metric isometry and the measures, and that $(\X,d_\X,\al)\sim(\spt(\al),d_\X,\al)$.
Thus mm-spaces are only compared on the measures' supports (not necessarily on the entire spaces $\X,\Y$).
Both distances are zero if and only if $\Xx\sim\Yy$, assuming $(\Xx,\Yy)$ are endowed with a complete metric.
Thus Sturm's and GW distances are able to distinguish mm-spaces up to this equivalence relation.
However, the two distances~\cite{memoli2011gromov, sturm2012space} are not equivalent.
For instance, there are differences in topological properties.
Sturm's distance is complete over the set of mm-spaces endowed with a complete metric, while $\GW$ is not.
Gromov-Wasserstein (and Sturm's) distance is a complete metric if one considers instead the set of mm-spaces endowed with a pseudo-metric (i.e. $d_\X$ or $d_\Y$ are not necessarily definite, but still satisfy $d_\X(x,x)=0$ for any $x\in\X$)~\cite{sturm2012space}.

\myparagraph{Implementations.}
From an applications standpoint, the main difference between the GW distance and Sturm's distance is that the former is more amenable to computations.
Note however that $\GW$ is also a non-convex problem, thus the computations only guarantee an output $\pi$ which is a stationary point of the functional.
There are several ways to obtain such stationary points, and all consist in updating iteratively the plan $\pi$.
One approach is to linearize the quadratic program as in~\cite{memoli2011gromov}.
It yields a sequence of OT programs where one alternates between the following two updates until convergence to a stationary point
\begin{align*}
	\C_\pi(x,y)&\leftarrow\int_{\X\times\Y} \Om(|d_\X(x,x') - d_\Y(y,y')|)\dd\pi(x',y'),\\
	\pi&\leftarrow\argmin_{\pi\in\Uu(\al,\be)} \int_{\X\times\Y} \C_\pi \dd\pi.
\end{align*}
It is possible to approximate the transport plan using entropy and Sinkhorn algorithm~\cite{peyre2016gromov, solomon2015convolutional}.
It is the approach considered in~\cite{sejourne2021unbalanced}.

Note that representing mm-spaces has an $O(N^2)$ memory footprint, thus the distortion $|d_\X(x,x')-d_\Y(y,y')|$ is an $O(N^4)$ 4D-tensor.
As noted in~\cite{peyre2016gromov}, under some assumptions (satisfied by the squared distortion $|d_\X - d_\Y|^2$), computing the cost $\C_\pi$ takes $O(N^3)$ in time and $O(N^2)$ in space, instead of $O(N^4)$.
This might however remain prohibitive for large scale graphs, thus several works propose to perform approximations of the plan to reduce computations.
One can assume for instance a hierarchical~\cite{xu2019scalable} or a low-rank~\cite{scetbon2021linear} structure of the plan.
An approach motivated by QAPs but which applies to GW is to interpolate between the optimization of a convex and a concave objective~\cite{zaslavskiy2008path}.
The GW problem can also be reformulated as a semi-definite program where the plan $\pi\otimes\pi$ is a 4D-tensor satisfying a low rank constraint.
Removing this rank constraint yields a relaxation with polynomial complexity, and which satisfies the triangle inequality between mm-spaces of same cardinality equipped with a uniform measure~\cite{villar2016polynomial}.

There exists variants of GW which are motivated by enriching the distance to account for additional information.
For instance many graph datasets~\cite{hu2020open} contain vector features providing information on the nodes or the edges (such as atomic information for molecules represented as graphs).
The GW distance is not naturally able to handle these features, which motivated variants such as the Fused-GW distance~\cite{vayer2020fused}, which penalizes features with OT and the graph structure with GW at the same time.
It is also possible to learn the nodes or edges' features as in~\cite{xu2019gromov}.
By analogy with the low complexity of OT between univariate measures, there is a sliced version of Gromov-Wasserstein between Euclidean mm-spaces~\cite{vayer2019sliced}.
One can also focus on Euclidean spaces equipped with Gaussian distributions, but so far (in contrast with OT~\cite{janati2020entropic}) the optimal plan is not known to be Gaussian.
Thus, assuming the transport plan is Gaussian, it is possible to derive a closed form which is an upper bound of $\GW$~\cite{salmona2021gromov}.
In the case of shapes it is also possible to replace the geodesic distance by the heat diffusion kernel~\cite{memoli2009spectral}.

\myparagraph{Applications.}
The $\GW$ distance was first motivated in~\cite{memoli2011gromov} to solve shape matching problems.
It was applied successfully in natural language processing for unsupervised translation learning~\cite{grave2019unsupervised, alvarez2018gromov}, in generative learning for objects lying in spaces of different dimensions~\cite{bunne2019learning} and to build VAE for graphs~\cite{xu2020learning}. 
It has been adapted for domain adaptation over different spaces~\cite{titouan2020co}.
It is also a relevant distance to compute barycenters between graphs or shapes ~\cite{vayer2020fused,chowdhury2020gromov}.
When $(\Xx,\Yy)$ are Euclidean spaces, this distance compares distributions up to rigid isometries, and is closely related (but not equal) to metrics defined by Procrustes analysis~\cite{grave2019unsupervised, alvarez2019towards}.
It was successfully applied in biology to align multi-omics biology~\cite{demetci2020gromov, demetci2022scot}.

\myparagraph{Computational hardness and relaxations of GW.}
Recall that $\GW$ is NP hard to solve in general, as for QAPs.
In particular, even though the above algorithm converges to a stationary plan~\cite{tseng2001convergence}, it is not necessarily the optimal one, one cannot prove its optimality, and it depends on the initialization.
Thus in~\cite{memoli2011gromov} several more computationally efficient lower bounds are proposed.
As proved in~\cite{memoli2018distance}, these bounds are not tight, i.e. they might output a score strictly smaller than $\GW(\Xx,\Yy)$.
Some of them were also motivated for shape registration as computationally amenable ways of representing and comparing mm-spaces.
For instance an mm-space $\Xx=(\X,d_\X,\al)$ can be summarized using the global histogram of distances $(d_\X)_\sharp(\al\otimes\al)$~\cite{osada2002shape}, or a distribution of local distance histograms $x\mapsto(d_\X(x,\cdot)_\sharp\al)$~\cite{hamza2003geodesic}.
Those representations being distributions in $\Mmp(\RR_+)$ or $\Mmp(\Mmp(\RR_+))$, they can be compared using OT distances.
The optimal plan $\pi^\star$ obtained via those relaxations can be used as initializations for GW, or applied in place of GW in applications.
For instance the local histograms of distances representation was applied in~\cite{chowdhury2019gromov}.

\subsection{Unbalanced Gromov-Wasserstein}
\label{sec:ugw}

\myparagraph{Robustness issues of GW.}
A limitation with GW is the mass conservation encoded by the constraint $\pi\in\Uu(\al,\be)$.
Similar to OT, this constraint induces a sensitivity to noisy data and outliers, which is not desirable in applications.
Thus it seems natural to relax those constraints as is performed for unbalanced OT (see Section~\ref{sec:uot}).
Two extensions of GW satisfying this property are proposed in~\cite{sejourne2021unbalanced}.
One formulation is amenable to computations, the other defines a distance between mm-spaces equipped with positive measures.

The first formulation (called UGW) consists in relaxing marginal constraints of GW by using penalties $\D_\phi^\otimes(\al|\be)=\D_\phi(\al\otimes\al|\be\otimes\be)$.
Using $\D_\phi^\otimes$ (and not $\D_\phi$ as for $\UOT$) is fundamental to derive an algorithm, and to connect it with the second formulation (called CGW) detailed in Definition~\ref{def:cgw}.

\begin{definition}[Unbalanced Gromov-Wasserstein]\label{def:ugw}
    The Unbalanced Gromov-Wasserstein (UGW) cost between two mm-spaces endowed with arbitrary positive measures is defined as $\UGW(\Xx,\Yy)\eqdef\inf_{\pi\geq 0} \Aa(\pi)$, where
\begin{align*}
	\Aa(\pi)\eqdef  \Gg(\pi) + \D_\phi^\otimes(\pi_1|\al) + \D_\phi^\otimes(\pi_2|\be).
\end{align*}
Recall $\Gg$ is defined in Definition~\ref{def:gw}.
\end{definition}

It is connected to the settings of conic formulations introduced in Section~\ref{sec:uot-conic}, see Definition~\ref{def:cgw}.
In practice one takes $\Om(t)=t^2$ for computational purposes~\cite{peyre2016gromov}.
Theorem~\ref{thm:cgw-dist} below states that the formulation $\UGW$ is definite.
We then leverage entropic regularization to derive a Sinkhorn-like algorithm.
An important modification from~\cite{solomon2016entropic, peyre2016gromov} is that we use a regularization $\epsilon\KL^\otimes$ instead of $\epsilon\KL$.

\myparagraph{Metric properties.}
By analogy with the construction of $\COT$ from $\UOT$ in Section~\ref{sec:uot-conic}, one can derive a conic formulation $\CGW$ from $\UGW$.
The cost $L_{\Om(\Gamma)}$ is converted in an homogeneous function $H_{\Om(\Gamma)}$.
Since metrics $(d_\X,d_\Y)$ lie in $\RR_+$, the cone distance is defined on $\Co[\RR_+]$ instead of $\Co[\X]$ and $\Co[\Y]$.
More specifically, we use coordinates $([u,a],[v,b]) = ([d_\X(x,x'),rr'], [d_\Y(y,y'),ss'])$ in $\Co[\RR_+] \times \Co[\RR_+]$, and the cone cost reads
\begin{align*}
	d_{\Co[\RR_+]}([u,a],[v,b])^q 
	&= H_{\Om(|u-v|)}(a^p, b^p)\\
	&=H_{\Om(|d_\X(x,x')-d_\Y(y,y')|)}((rr')^p, (ss')^p),
\end{align*}
where $H_c$ is the function defined Equation~\ref{eq:persp_hc}.

While UOT plans depend on variables  $([x,r],[y,s])$ and $([x',r'],[y',s'])$ in $\Co[\X]\times\Co[\Y]$, the transportation cost involved in our conic formulation only makes use of the 2-D cone $\Co[\RR_+]$ over $\RR_+$ endowed with the distance $|u-v|$ (note that any other distance on $\RR$ could be used as well).
All in all, it yields the following definition of CGW.

\begin{definition}[Conic Gromov-Wassserstein]\label{def:cgw}
	We define
	\begin{align*}
	    \CGW(\Xx,\Yy) \eqdef \inf_{\eta\in\Uu^{\Co}(\al,\be)}\Hh(\eta),
	\end{align*}
	where
	\begin{equation}\label{eq:ugw-conic}
	\begin{aligned}
	\Hh(\eta)\eqdef\int d_{\Co[\RR_+]}(&[d_\X(x,x'), r r'], [d_\Y(y,y'), s s'])^q \\
	&\dd\eta([x,r], [y,s])\dd\eta([x',r'], [y',s']),
	\end{aligned}
	\end{equation}
	and $\Uu^{\Co}(\al,\be)$ is defined as the set of plans $\eta\in\Mm_+(\Co[\X]\times\Co[\Y])$ satisfying
	\begin{equation}\label{eq:const-conic}\left\{
	\begin{aligned}
	\int_{\RR_+} r^p \dd\eta_1([\cdot,r])=\al,\;
	\int_{\RR_+} s^p \dd\eta_2([\cdot,s])=\be
	\end{aligned}
	\right\}.
	\end{equation}
\end{definition}

Note that the constraint set $\Uu^{\Co}(\al,\be)$ is similar to the one of $\COT$ (see~\ref{sec:uot-conic}).
Note also that similarly to the $\GW$ formulation (Definition~\ref{def:gw}) -- and in sharp contrast with the conic formulation of UOT -- here the transport plans are defined on the cone $\Co[\X] \times \Co[\Y]$ but the cost $d_{\Co[\RR_+]}$ is a distance on $\Co[\RR_+]$.

The CGW formulation is convenient to prove metric properties.
Its derivation shares some similarities with the formulation $\COT$.
In particular, when the cone cost $d_{\Co[\RR_+]}$ is a distance then so is $\CGW$, and we refer to Section~\ref{sec:uot-conic} for a review of explicit setting where it is the case.

\begin{theorem}\label{thm:cgw-dist}
	(i) The divergence $\UGW$ is symmetric, positive and definite up to isometries.
	(ii) If $d_{\Co[\RR_+]}$ is a distance on $\Co[\RR_+]$, then $\CGW^{1/q}$ is a distance on the set of mm-spaces up to isometries.
	(iii) For any $(\D_\phi, \Om, p, q)$ with associated cost $d_{\Co[\RR_+]}$ on the cone, one has $\UGW\geq\CGW$.
\end{theorem}

\myparagraph{Numerical computation.}
In order to derive a simple numerical approximation scheme for $\UGW$, following~\cite{memoli2011gromov}, we consider a lower bound obtained by introducing two transportation plans. To further accelerate the method and enable GPU-friendly iterations, similarly to~\cite{gold1996softmax,solomon2016entropic}, we consider an entropic regularization. We define for any $\epsilon \geq 0$
\begin{align}
	\begin{aligned}
	\UGW_\epsilon(\Xx, \Yy) &\triangleq \inf_{\pi\in\Mmp(\X\times\Y)} \Aa_\epsilon(\pi),\\
	\text{where} \qquad \Aa_\epsilon(\pi) &\triangleq \Aa(\pi) +\epsilon\KL^\otimes(\pi|\al\otimes\be).
	\end{aligned}
\end{align}
We consider the following relaxation
\begin{align}\label{eq:lower-bound}
	\UGW_\epsilon(\Xx, \Yy) &\geq\uinf{\pi,\ga} \Bb(\pi,\ga) +\epsilon\KL(\pi\otimes\gamma|(\al\otimes\be)^{\otimes 2}),\\
		\qandq \Bb(\pi,\ga) &\eqdef \int_{X^2 \times \Y^2} \Om(|d_\X - d_\Y|)\dd\pi \otimes \ga\nonumber\\
	&\quad+ \D_\phi(\pi_1\otimes\ga_1|\al\otimes\al) + \D_\phi(\pi_2 \otimes \ga_2|\be \otimes \be),\nonumber
\end{align}
where $(\ga_1,\ga_2)$ denote the marginals of the plan $\ga$.
Most importantly, note that $\Aa(\pi) = \Bb(\pi,\pi)$.
In the sequel we write $\Bb_\epsilon = \Bb + \epsilon\KL^\otimes$.

The crux of these computational methods is to reformulate tensorized $\D_\phi^\otimes$ using standard divergences $\D_\phi$.
We provide below formulas for the KL, Berg and balanced setting.
A priori, such relations might not exist for all divergences.
\begin{proposition}\label{prop:decompose-kl}
	For any measures $(\mu,\nu, \al,\be)\in\Mm_+(\Xx)$, one has
	\begin{equation}
	\begin{aligned}
	\KL(\mu\otimes\nu|\al\otimes\be) &= m(\nu)\KL(\mu|\al) +  m(\mu)\KL(\nu|\be)\\
	&\qquad+ (m(\mu) - m(\al))(m(\nu) - m(\be)).
	\end{aligned}
	\end{equation}
	In particular, $\KL(\mu\otimes\mu|\nu\otimes\nu) = 2m(\mu)\KL(\mu|\nu) + (m(\mu) - m(\nu))^2$.
	\rev{The same formula holds for the Berg entropy $\text{Berg}(\mu|\al)=\KL(\al|\mu)$}.
	Furthermore, for balanced GW, it reads $\iota_{(=)}(\mu\otimes\nu|\al\otimes\be)=\iota_{(=)}(\mu|\al) +\iota_{(=)}(\nu|\be)$.
\end{proposition}
In the Balanced setting, with $(\mu,\nu)$ probabilities, the regularization reads 
\begin{align*}
	\KL^\otimes(\pi|\mu\otimes\nu) = 2\KL(\pi|\mu\otimes\nu).
\end{align*}
Thus (up to a factor 2) we retrieve as a particular case the setting of~\cite{peyre2016gromov}.

We now give an important result from~\cite{sejourne2021unbalanced}.
It states that minimizing the bi-convex relaxation~\eqref{eq:lower-bound} by doing an alternate descent on $(\pi,\ga)$ is equivalent to solving a sequence  of regularized UOT problems.
This result only holds when $\D_\phi=\rho\KL$, balanced OT and Berg entropy, and remains open in other settings.
However, it is a strong result since it allows us to use the Sinkhorn algorithm we detailed in Section~\ref{sec:sinkhorn} as a subroutine.
\rev{We provide the closed forms of this equivalence for KL and Balanced OT.}

\begin{proposition}\label{prop:alternate-simple}
	When $\D^\otimes_\phi=\rho\KL^\otimes$, for a fixed $\ga$, the optimal
	\begin{align*}
	    \pi\in\arg\umin{\pi} \Bb(\pi,\ga) +\epsilon\KL(\pi\otimes\gamma|(\al\otimes\be)^{\otimes 2})
	\end{align*}
	solves
	\begin{align*}
		\umin{\pi} \int c^\epsilon_\ga(x,y) \dd\pi(x,y) &+ \rho m(\ga) \KL(\pi_1|\al) + \rho m(\ga) \KL(\pi_2|\be)\\ &+ \epsilon m(\ga) \KL(\pi|\al\otimes\be),
	\end{align*}
	where $m(\ga) \eqdef \ga(\X \times \Y)$ is the mass of $\ga$, and
	where we define the cost associated to $\ga$ as
	\begin{align*}
		c^\epsilon_\ga(x,y) \!  &\eqdef \!  \int \! \Om(|d_\X(x,x') - d_\Y(y,y')|)\dd\ga(x',y') \! 
		+ \!  \rho\!\!  \int \! \log\Big(\frac{\dd\ga_1}{\dd\al}(x')\Big)\dd\ga_1(x') \! 
		\\&+ \! \rho \! \!  \int \! \log\Big(\frac{\dd\ga_2}{\dd\be}(y')\Big)\dd\ga_2(y') \! 
		 + \!  \epsilon \! \! \int \!   \log\Big(\frac{\dd\ga}{\dd\al\dd\be}(x',y')\Big)\dd\ga(x',y').
	\end{align*}
	
	In the setting of balanced OT, the minimization is equivalent to
	\begin{align*}
		\umin{\pi\in\Uu(\al,\be)} \int c^\epsilon_\ga(x,y) \dd\pi(x,y) + \epsilon m(\ga) \KL(\pi|\al\otimes\be),
	\end{align*}
	where the cost $c^\epsilon_\ga$ reads
	\begin{align*}
		c^\epsilon_\ga(x,y) \!  &\eqdef \!  \int \! \Om(|d_\X(x,x') - d_\Y(y,y')|)\dd\ga(x',y') \!
		 + \!  \epsilon \! \! \int \!   \log\Big(\frac{\dd\ga}{\dd\al\dd\be}\Big)\dd\ga.
	\end{align*}
\end{proposition}
While the balanced GW functional and its associated Sinkhorn algorithm are retrieved as the limit of the $\KL$ setting with $\rho\rightarrow\infty$, the cost $c^\epsilon_\ga$ for balanced GW is not retrieved with such limit, hence the need to specify the equivalence of formulations in such setting.

\rev{We detail the pseudo-code to implement the results of Proposition~\ref{prop:alternate-simple} in Algorithm~\ref{alg:gw}.
Note that it reuses Sinkhorn Algorithm~\ref{alg:sinkhorn}.
Since UGW is a non-convex problem, the limit of the algorithm depends on the initialization. An efficient initialization strategy is to use the solutions of lower bounds of GW as detailed in~\cite{memoli2011gromov}.

Practically speaking, the computation of UGW might suffer several instabilities. First, the non-convexity of the functional makes it highly dependent on the initialization, which impact the quality of the computed local minimum. 
Another possible issue is the scaling in $1/m(\pi)\epsilon$ of the update Line~\ref{eq:update-ugw-plan} in Algorithm~\ref{alg:gw}, which might be numerically sensitive and can lead to  numerical over/underflows when using poor initialization. 
}

\begin{algorithm} 
	\caption{~~~~\, {UGW($(\al_i)_i$, $(D^\X_{ij})_{ij}$, $(\be_j)_j$, $(D^\Y_{ij})_{ij}$} \label{alg:gw}}
	\textbf{Input~~~~\,:}~~  source $\alpha = \sum_{i=1}^\N \al_i\delta_{x_i}$ with metric $(D^\X_{ij}) = (D^\X(x_i,x_j))$,~target~ $\beta = \sum_{j=1}^\M \be_j\delta_{y_j}$ with metric $(D^\Y_{ij}) = (D^\Y(y_i,y_j))$, with $(\al_i)\in\RR^N$, $(\be_j)\in\RR^M$, $(x_i)_i\in\RR^{\N\times D}$, $(y_j)_j\in\RR^{\M\times D}$\\
	\textbf{Parameters~:}~~ entropy $\phi$, regularization $\epsilon > 0$ \\
	\textbf{Output~~\,:}~~ stationary plans $(\pi,\ga)$ of the functional $\Bb$
	\begin{algorithmic}[1]
		\STATE Initialize $\pi$ \COMMENT{Heuristically or take $\pi=\al\otimes\be$}
		\vspace{.1cm}\WHILE{$(\pi,\ga)$ have not converged}\vspace{.1cm}
		\STATE $\C_{ij} \gets c^\epsilon_\pi(x_i, y_j)$
		\STATE $\epsilon_t, \rho_t \gets m(\pi)\epsilon, m(\pi)\rho$ \COMMENT{KL setting}
		\STATE $(\f_i), (\g_j) \gets \text{Sinkhorn}(\al,\be, \C_{ij})$ \COMMENT{see Algorithm~\ref{alg:sinkhorn}}
		\STATE $\ga_{ij} \gets \exp[(\f_i + \g_j - \C_{ij}) / \epsilon_t]\al_i\be_j$
		\vspace*{0.5em} \label{eq:update-ugw-plan}
		\STATE $\C_{ij} \gets c^\epsilon_\ga(x_i, y_j)$
		\STATE $\epsilon_t, \rho_t \gets m(\ga)\epsilon, m(\ga)\rho$ \COMMENT{KL setting}
		\STATE $(\f_i), (\g_j) \gets \text{Sinkhorn}(\al,\be, \C_{ij})$ \COMMENT{see Algorithm~\ref{alg:sinkhorn}}
		\STATE $\pi_{ij} \gets \exp[(\f_i + \g_j - \C_{ij}) / \epsilon_t]\al_i\be_j$
		\ENDWHILE
		\RETURN{~~$(\pi,\ga)$}
	\end{algorithmic}
\end{algorithm}

\myparagraph{Towards debiasing \texorpdfstring{$\GW_\epsilon$}{}.}
\label{sec:debiasing}
We detailed above an approximation of $\GW$ using its entropic regularized version $\GW_{\epsilon}$.
By analogy with the debiasing performed in~\cite{feydy2018interpolating, sejourne2019sinkhorn} and detailed in Section~\ref{sec:debiasing}, it is a natural question to ask if there is an entropic bias (similar to $\OTb$ with $\Sb$, see Definition~\ref{def:sink-div}), and if it needs to be corrected.
Providing answers on this problem for $\GW_{\epsilon}$ is difficult due to the non-convexity of the formulation, and remains an open problem.
To provide a partial answer, we detail results from~\cite{sejourne2020generalized} on the asymptotic $\epsilon\rightarrow\infty$, for balanced $\GW$.
We now write the distortion $k(d_\X(x,x'), d_\Y(y, y'))$ instead of $|d_\X(x,x') - d_\Y(y, y')|$, where $k$ is a kernel defined on $\RR$, which is a more general expression than Definition~\ref{def:ugw}.
Typically, we take $k(x,y) = |x-y|^p$ where $1\leq p<2$.
We start with a result on the limit of $\GW_{\epsilon}$ when $\epsilon\rightarrow\infty$.

\begin{proposition}\label{prop:limit-gw-entropic}
	One has $\GW_{\epsilon}(\Xx,\Yy)\xrightarrow{\epsilon\rightarrow\infty}\GW_{\infty}(\Xx,\Yy)$, where
	\begin{align}\label{eq:limit-gw-entropic}
		\GW_{\infty}(\Xx,\Yy) \triangleq \int_{(\Xx\times\Yy)^2} k(d_\X(x,x'), d_\Y(y, y')) \dd\al(x)\dd\al(x')\dd\be(y)\dd\be(y').
	\end{align}
\end{proposition}

Similar to $\OTb$, a large $\epsilon$ makes the independent coupling $\al\otimes\be$ optimal.
This ``decorrelation'' allows a reformulation of $\GW_{\infty}$ as a kernel inner product.
Define the map $\Phi(\Xx)\triangleq d_{\X\sharp}(\al\otimes\al)\in\Mmp(\RR)$.
If $\al\in\Mmpo(\X)$ then $\Phi(\Xx)\in\Mmpo(\RR)$.
The kernel inner product reads
\begin{align}\label{eq:gw-kernel-norm}
	\GW_{\infty}(\Xx,\Yy) = \dotp{\Phi(\Xx)}{\Phi(\Yy)}_{\Hh^*} = \int_{\RR^2} k(x, y) \dd\Phi(\Xx)(x)\dd\Phi(\Yy)(y).
\end{align}
One can define a debiased Sinkhorn-GW divergence and its limit $\eps\rightarrow\infty$
\begin{gather*}
	\SGW_\epsilon(\Xx,\Yy) \triangleq \GW_\epsilon(\Xx,\Yy) - \tfrac{1}{2}\GW_\epsilon(\Xx,\Xx) -\tfrac{1}{2}\GW_\epsilon(\Yy,\Yy),\\
	\SGW_{\infty}(\Xx,\Yy)\triangleq \tfrac{1}{2}\norm{\Phi(\Xx) - \Phi(\Yy)}^2_{\Hh^*}.
\end{gather*}
Note that Proposition~\ref{prop:limit-gw-entropic} ensures that
$
\SGW_{\epsilon}(\Xx,\Yy)\xrightarrow{\epsilon\rightarrow\infty}\SGW_{\infty}(\Xx,\Yy).
$
Focusing once again on the asymptotic $\SGW_{\infty}$, the following proposition shows the positivity of the divergence.

\begin{proposition}\label{prop:pos-gw-div}
	Assume $-k$ is a conditionally positive kernel, and $(\Xx,\Yy)$ are mm-spaces endowed with measures of equal masses.
	Then $\SGW_{\infty}(\Xx,\Yy)\geq 0$.
\end{proposition}

By continuity, this result shows that positivity of the kernel $-k$ ensures that $\SGW_{\epsilon}$ is positive for $\epsilon$ large enough. It is an open problem to show that this also ensures  positivity for all $\epsilon$.
Note furthermore that when $\epsilon=+\infty$, even if $\SGW_{\infty}$ is positive, it fails to be definite, i.e. there exists non isometric mm-spaces such that $\SGW_{\infty}(\Xx,\Yy) = 0$, see~\cite[Figure 8]{memoli2011gromov}.
We refer to~\cite{memoli2018distance} for the study of counter-examples and other maps $\Phi$.

\begin{figure}[h]
	\begin{subfigure}{.45\textwidth}
		\centering
		\includegraphics[width=\linewidth]{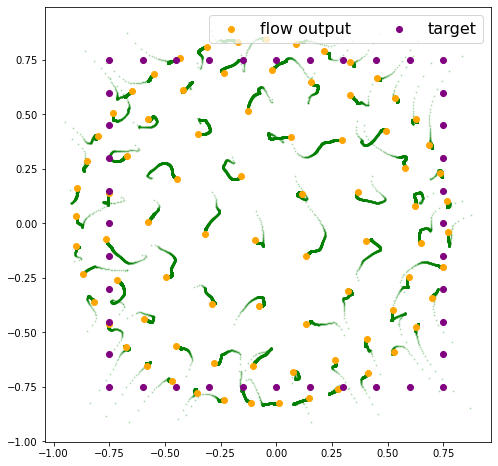}
		\caption{$\GW_{\infty}(\Xx(x),\Yy)$}
	\end{subfigure}
	\hfill
	\begin{subfigure}{.45\textwidth}
		\centering
		\includegraphics[width=\linewidth]{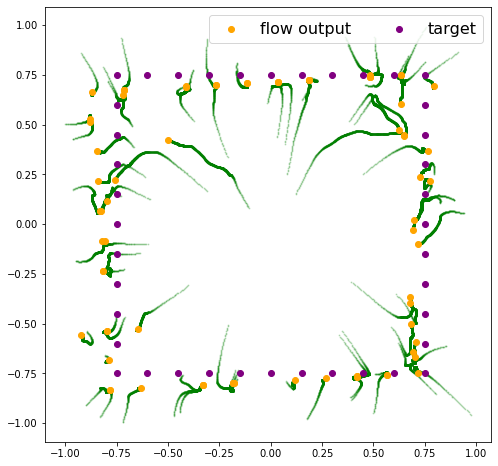}
		\caption{$\SGW_{\infty}(\Xx(x),\Yy)$}
	\end{subfigure}
	\caption{\textit{Output of the gradient flow to minimize $x\mapsto\GW_{\infty}(\Xx(x),\Yy)$ and $x\mapsto\SGW_{\infty}(\Xx(x),\Yy)$.
		The space $\Yy$ is displayed in blue, the output $\Xx(x)$ in red, and the trajectories of the particles in green.
		Extracted from~\protect\cite[Chapter 5]{sejourne2020generalized}.
		}
	}
	\label{fig:flow_debiased_gw}
\end{figure}

Figure~\ref{fig:flow_debiased_gw} showcases numerically the interest of this debiased quantity $\SGW_\epsilon$ in order to reduce the bias induced by entropic regularization. 
We consider the particular case of $k(x,y)=|x-y|^p$.
It defines a conditionally negative kernel (for $1\leq p<2$) and is smooth (for $p>1$).
Thus in that setting one has
\begin{align*}
	\GW_{\infty}(\Xx,\Yy) &= \dotp{\Phi(\Xx)}{\Phi(\Yy)}_{\Hh^*}\\
	&=\int |d_\X(x,x') - d_\Y(y,y')|^p\dd\al(x)\dd\al(x')\dd\be(y)\dd\be(y'),
\end{align*}
and
\begin{align*}
\SGW_{\infty}(\Xx,\Yy) &= \dotp{\Phi(\Xx)}{\Phi(\Yy)}_{\Hh^*}\\
&-\tfrac{1}{2} \dotp{\Phi(\Xx)}{\Phi(\Xx)}_{\Hh^*} -\tfrac{1}{2} \dotp{\Phi(\Yy)}{\Phi(\Yy)}_{\Hh^*}.
\end{align*}
Note that taking $p=2$ induces a degenerate kernel norm $\norm{\al - \be}^2_{\Hh^*} = |\EE[\al] - \EE[\be]|^2$ which only compares the means. We used $p=1.5$ in Figure~\ref{fig:flow_debiased_gw}.
We take $\Yy$ to be a regular discretization of a rotated square in $\RR^2$, endowed with the Euclidean distance and uniform measure.
We consider a discrete mm-space parameterized by its support $x=(x_i)\in(\RR^2)^N$, i.e. which reads $\Xx(x)=(\{x_i\}, \norm{.}_2, \al(x))$ where $\al(x) = \tfrac{1}{N}\sum_i\de_{x_i}$.
We perform a gradient flow to minimize $x\mapsto\GW_{\infty}(\Xx(x),\Yy)$ and $x\mapsto\SGW_{\infty}(\Xx(x),\Yy)$ w.r.t to the positions of the samples.
It is similar to the gradient flow experiment of~\cite{sejourne2019sinkhorn} shown above (Figure~\ref{fig:flow-reg}), with the difference that the learning rate on the particles' masses is $0$ to keep them constant.
The support of $\Xx(x_0)$ is initialized at random on the unit square.
The results are displayed in Figure~\ref{fig:flow_debiased_gw}.
We see that the output of the flow for $\SGW_{\infty}$ seems close to be a square, up to an isometry of $\Yy$ (rotation and translation).
By comparison, the output of the gradient flow for $\GW_{\infty}$ converges to a distribution supported on a disk, and thus ignores the structure of $\Yy$.
This experiment highlights the bias of entropic regularization and the relevance of correcting it.
It motivates as a future research direction the study of $\SGW_\epsilon$ for $\epsilon>0$.

\section*{Conclusion}

This review paper has shown how to integrate several recent mathematical and algorithmic ideas to make OT noise-aware without sacrificing scalability. Sinkhorn's algorithm remains the workhorse of this class of generalized OT problems, leveraging GPU architectures and fighting the curse of dimensionality. The extension of this framework to  Gromov-Wasserstein opens the door to new challenges to import existing OT results to this non-convex setting.

\section*{Acknowledgements}

The work of G. Peyr\'e is supported by the European Research Council (ERC project NORIA) and by the French government under management of
Agence Nationale de la Recherche as part of the ``Investissements d’avenir'' program, reference ANR19-P3IA-0001 (PRAIRIE 3IA Institute).

\bibliographystyle{plain}
\bibliography{biblio}
\end{document}